%% file: main.tex
\documentclass[10pt,twocolumn,letterpaper]{article}

\usepackage{cvpr}              % To produce the CAMERA-READY version
\usepackage{graphicx}
\usepackage{amsmath}
\usepackage{amssymb}
\usepackage{booktabs}
\usepackage[pagebackref,breaklinks,colorlinks]{hyperref}

\usepackage[capitalize]{cleveref}
\crefname{section}{Sec.}{Secs.}
\Crefname{section}{Section}{Sections}
\Crefname{table}{Table}{Tables}
\crefname{table}{Tab.}{Tabs.}

\usepackage{multirow}
\usepackage{array}
\usepackage{graphicx}
\usepackage{booktabs} % For formal tables
\usepackage{enumerate}
\usepackage{times}
\usepackage{epsfig}
\usepackage{graphicx}
\usepackage{amsmath}
\usepackage{amssymb}
\usepackage{soul}
\usepackage{multirow}
\usepackage{booktabs}
\usepackage{wrapfig}
\usepackage{footnote}
\usepackage{subcaption}
\usepackage{algorithmicx}
\usepackage{algorithm}
\usepackage{algpseudocode}
\usepackage{enumitem}
\usepackage{xcolor}

\algrenewcommand\textproc{}
\algblock{Input}{EndInput}
\algnotext{EndInput}
\algblock{Output}{EndOutput}
\algnotext{EndOutput}
\newcommand{\Desc}[2]{\State \makebox[2em][l]{#1}#2}

\setlist[itemize]{leftmargin=*}
\setlist[enumerate]{leftmargin=*}

\def\cvprPaperID{9594} % *** Enter the CVPR Paper ID here
\def\confName{CVPR}
\def\confYear{2023}

\input{macros}

\definecolor{purple}{rgb}{0.6, 0.4, 0.8}
\definecolor{orange}{rgb}{1.0, 0.49, 0.0}

\begin{document}

\title {Unsupervised 3D Shape Reconstruction by Part Retrieval and Assembly}
\author{\normalsize Xianghao Xu\\
\normalsize Brown University\\
\normalsize USA\\
%{\tt\small xianghao\_xu@brown.edu}
\and
\normalsize Paul Guerrero\\
\normalsize Adobe Research\\
\normalsize UK\\
%{\tt\small guerrero@adobe.com}
\and
\normalsize Matthew Fisher\\
\normalsize Adobe Research\\
\normalsize USA\\
%{\tt\small matfishe@adobe.com}
\and
\normalsize Siddhartha Chaudhuri\\
\normalsize Adobe Research\\
\normalsize India\\
%{\tt\small siddhartha.chaudhuri@gmail.com}
\and
\normalsize Daniel Ritchie\\
\normalsize Brown University\\
\normalsize USA\\
%{\tt\small daniel\_ritchie@brown.edu}
}

\twocolumn[{%
\renewcommand\twocolumn[1]{#1}%
\maketitle
\begin{center}
    \centering
    \includegraphics[width=0.98\linewidth]{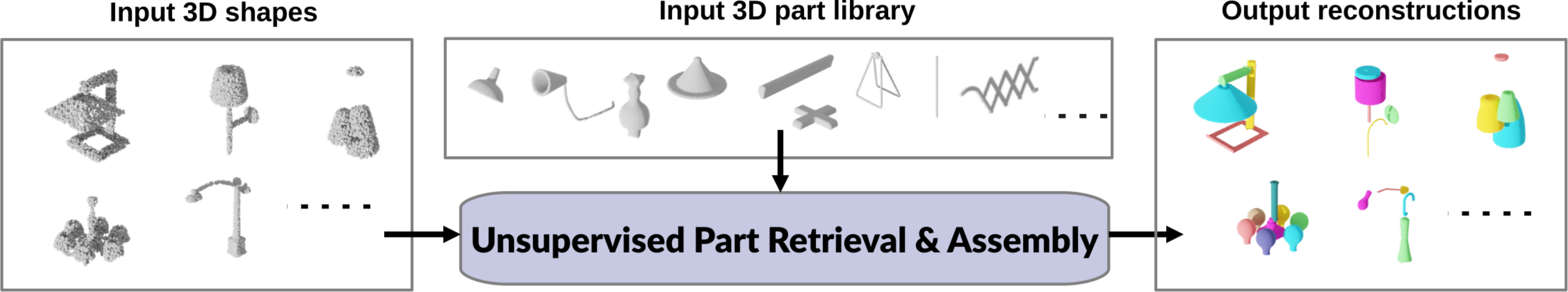}
    \captionof{figure}{Our system takes target 3D shapes together with a 3D part library as input and outputs a set of retrieved and transformed parts from the part library that recreates the input target shapes.
    }
    \label{fig:teaser}
\end{center}%
}]

\begin{abstract}
\input{00-abstract}
\end{abstract}

\input{01-intro}

\input{02-related}
\input{03-method}

\input{04-results}

\input{05-conclusion}

{\small
\bibliographystyle{ieee_fullname}
\bibliography{egbib}
}

\clearpage

\input{supplement_append}

\end{document}

%% file: macros.tex
\newcommand{\reals}{\ensuremath{\mathbb{R}}}

\newenvironment{packed_itemize}
{\begin{itemize}
    \vspace{-\topsep}
    \setlength{\itemsep}{1pt}
    \setlength{\parskip}{0pt}
    \setlength{\parsep}{0pt}
}{\end{itemize}}

%% file: 00-abstract.tex
Representing a 3D shape with a set of primitives can aid perception of structure, improve robotic object manipulation, and enable editing, stylization, and compression of 3D shapes. Existing methods either use simple parametric primitives or learn a generative shape space of parts. Both have limitations: parametric primitives lead to coarse approximations, while learned parts offer too little control over the decomposition. We instead propose to decompose shapes using a library of 3D parts provided by the user, giving full control over the choice of parts. The library can contain parts with high-quality geometry that are suitable for a given category, resulting in meaningful decompositions with clean geometry. The type of decomposition can also be controlled through the choice of parts in the library. Our method works via a self-supervised approach that iteratively retrieves parts from the library and refines their placements. We show that this approach gives higher reconstruction accuracy and more desirable decompositions than existing approaches. Additionally, we show how the decomposition can be controlled through the part library by using different part libraries to reconstruct the same shapes.

%Reconstructing 3D shapes using a set 3D geometries are important for populating virtual worlds and for synthetic data generation. Existing methods accomplish this by either retrieving and deforming entire 3D shapes or by assembling pre-defined primitives. In this project, we present a self-supervised approach for reconstructing 3D shapes by assembling arbitrary 3D parts with a wide range of geometry variations. Our approach accomplishes this by an iterative part retrieval process. We evaluate our method by reconstructing testing shapes from segmented parts of source shapes in PartNet Dataset, and we compare our method against two state-of-the-art shape reconstruction baselines.

%% file: 01-intro.tex
\section{Introduction}

The ability to compactly represent a 3D shape as a combination of primitive elements has applications in multiple domains.
In computer vision, the ability to automatically decompose a shape into parts can aid machine perception of the 3D structure of objects, which can in turn help autonomous agents plan how to manipulate such objects.
In computer graphics, a combination of primitives can be used as a compressed geometry representation, as a way to abstract, stylize, or edit a 3D shape   by allowing users to alter the underlying primitive library.
Ideally, a system that performs this kind of shape decomposition should be able to do so without supervision in the form of ground-truth decompositions, as such data is rarely available at scale.

Past research in vision and graphics has studied this unsupervised shape decomposition problem.
Initially, researchers sought methods for decomposing 3D shapes into sets of simple parametric primitives, such as cuboids or superquadric surfaces~\cite{abstractionTulsiani17,sun2019abstraction,yang2021unsupervised,Paschalidou2019CVPR}. These methods produce clean, parametric geometry as output, and the choice of primitive type allows a small degree of user control over the decomposition. However, parametric primitives produce only a coarse approximation of the input shape, which may not be desirable in all applications. Thus, more recent work has investigated unsupervised decomposition of shapes into arbitrarily-shaped primitives whose geometries are determined by a neural network~\cite{chen2019bae_net,NeuralStarDomain,Paschalidou2021CVPR}. These methods produce a set of ``neural primitives'' whose union closely approximates the input shape. However, the geometries of these primitives may contain artifacts (e.g. bumps, blobs). Further, these methods offer little to no control over the type of decomposition they produce -- the network outputs whatever primitives it thinks are best to reconstruct the input shape since it lacks access to a supervised part prior.
% \sid{I added the ``since\dots'' to make the point clearer but obvious question is: why not try to regularize the primitive shapes produced by using an additional part prior trained on our part library, e.g. see AdaCoSeg? We should address this point.}

Is it possible to obtain a decomposition of a 3D shape whose primitives exhibit clean geometry and closely reconstruct the input shape, while also providing more control over the type of decomposition produced?
% Neural primitives got better match to target shape, but sacrificed the clean geometry and decomposition control offered by parametric primitives. Is there a way get all? \pg{We could also try to argue that both parametric primitives and neural primitives offer too little control over the type of primitives used in the composition. We want to limit the primitives used to those suitable for a given category of shapes, like armrests for chairs, etc. We could also argue the following, but I am less sure about that since we don't have clear experiments backing this up: Apart from making sure that only suitable primitives are used to compose a given shape, using only semantically meaningful parts of a category also encourages a more semantically meaningful decomposition.}
This is possible if, rather than using simple parametric primitives or arbitrary neural primitives, one chooses a middle point between these two extremes: reconstruct an input shape by retrieving and assembling primitives from a \emph{library of pre-defined 3D parts}.
This retrieve-and-assemble approach has several advantages.
First, the parts in the library can be high-quality 3D meshes, guaranteeing clean geometry as output.
Second, a large part library can contain parts that are good geometric matches for different regions of various shapes, meaning that accurate reconstructions of input shapes are possible.
Finally, this approach offers a high degree of controllability, as the user can change the part library to produce different decompositions of the same input shape.
% - This approach subsumes the parametric primitive approaches: can create a big part library by sampling many different instantiations of parametric primitives, then use our method.

% \sid{Maybe we should mention \cite{PartBasedStructureRecovery} at this point, cite its limitations, and then go on to describing our method and its advantages?}
% \dr{I disagree. I think this complicates the narrative in the introduction. We do address our relationship to it in Related Work.}

In this paper, we present a method for unsupervised decomposition of 3D shapes using a user-defined library of parts.
Finding a subset of parts from a large part library which best reconstructs an input shape is a large-scale combinatorial search problem.
To make this problem tractable, we represent the library of parts on a continuous manifold by training a part autoencoder. This continuous representation of the part library allows us jointly optimize for the identities and poses of parts which reconstruct the input shape. To escape the many worst of local optimas in this optimization landscape, the algorithm periodically uses its current predicted set of parts to segment the input shape; these segments are then re-encoded into the part feature manifold to produce a new estimate of the parts that best reconstruct the input shape. This data-driven, discontinuous jump in the optimization state is similar to stages from other non-gradient-based algorithms for global optimization or latent variable estimation, including the mean shift step from the mean shift algorithm and the E-step from expectation maximization~\cite{MeanShiftEM}.
% [Draw some connection to k-means, or some other clustering algorithm?] \pg{maybe expectation maximization?}

Our algorithm can be run independently for any individual target shape, allowing it to work in a ``zero-shot'' setting. When a larger dataset of related shapes is available, we can also optimize for their part decomposition in advance (a ``training'' phase) and then perform fast decomposition of a new shape from that category by initializing its decomposition using its nearest neighbor from the ``training'' set.
% We show that this optimization-based approach performs better than a trained network on shape categories with lots of structural heterogeneity.

We evaluate our algorithm by using it to reconstruct shapes from point clouds, using parts from the PartNet dataset.
We compare to the recent Neural Parts unsupervised decomposition system~\cite{Paschalidou2021CVPR} and show that our algorithm produces qualitatively more desirable decompositions that also achieve higher reconstruction accuracy.
We demonstrate the control offered by our method by showing how it is possible to reconstruct shapes from one category using parts from another (e.g. make a chair out of lamp parts).
This also has application for 3D graphics content creation, which we demonstrate by reconstructing target shapes using parts from a modular 3D asset library.

In summary, our contribution is an unsupervised algorithm which retrieves and poses 3D parts to reconstruct input 3D shapes. 
We will release our code upon publication.

%% file: 02-related.tex
\section{Related Work}

Our contribution is related to prior work on unsupervised shape decomposition and on modeling by part retrieval and assembly.
We do not discuss the considerable body of work on \emph{supervised} decomposition/segmentation of 3D shapes.

\paragraph{Unsupervised shape decomposition:}

One class of unsupervised shape decomposition method reconstructs shapes using parametric primitives.
Several approaches approximate 3D shapes as collections of cuboids~\cite{abstractionTulsiani17,sun2019abstraction,yang2021unsupervised}.
One can obtain slightly more geometric flexibility by using a collection of superquadric surfaces instead of cuboids~\cite{Paschalidou2019CVPR}.
These approaches produce clean output geometry and offer some small degree of control over the type of decomposition produced, but their low-degree-of-freedom primitive representation results in a poor fit to the input shape.

Another class of approaches approximates the input shape with a collection of more general polyhedra.
Several methods focus on convex polyhedra, either decomposing individual shapes in isolation~\cite{kaick2014shape,asafi2013weak,lien2008approximate} or training a neural network to produce similar convex decompositions for similar shapes from a category~\cite{deng2020cvxnet}.
Another option is to decompose the input shape into pieces which can be represented as generalized cylinders~\cite{GeneralizedCylinderDecomp}.
These approaches produce clean geometry with a better fit to the input shape than paramatric primitives allow, but they offer no control over the type of decomposition produced.
They also typically need many primitives to fit the input shape well, making them non-compact and not well-suited for shape editing.

Recent research in this space has focused on decomposing shapes using neural primitives.
BAE-Net trains an implicit shape representation that uses multiple decoder ``heads,'' where each head tends to represent the same localized part across many shape instances~\cite{chen2019bae_net}.
Other approaches represent neural parts as star domains~\cite{NeuralStarDomain} or deformed sphere meshes~\cite{Paschalidou2021CVPR}.
These approaches produce decompositions that fit the input shape well using a small number of primitives.
However, their output geometry can exhibit undesirable artifacts, and they provide no control over the type of decomposition produced.
We compare our algorithm to one of these approaches later in the paper and show that ours achieves even better reconstruction accuracy while also producing qualitatively better decompositions.

\paragraph{Modeling by retrieval and assembly:}

A large body of work in computer graphics has considered computer-assisted or fully-automated 3D modeling via retrieving and assembling pre-existing 3D shapes.
Early work in this space focused on geometric heuristics to decide what object parts might be connected to one another~\cite{ModelingByExample,Shuffler}.
Later, researchers began exploring machine learning methods to suggest relevant parts to add to partial 3D object~\cite{SidVangelisAssembly} or to synthesize entire 3D objects by putting parts together~\cite{SidVangelisSynthesis}.
More recently, deep networks have been applied to the problems of suggesting parts~\cite{Sung:2017} and part-based shape synthesis~\cite{li2017grass,ShapePartSlotMachine}.
These methods focus on creating new 3D shapes, rather than approximating/reconstructing existing ones, so they tackle a different problem than we do.

There has also been some work on retrieval-based reconstruction of input 3D shapes.
One method learns to retrieve and deform entire shapes from a shape database to best match an input point cloud or image~\cite{uy-joint-cvpr21}.
If the input is structurally distinct from any of the shapes in the database, however, this approach will not perform well.
More closely related to our algorithm is the Structure Recovery by Part Assembly system~\cite{PartBasedStructureRecovery}, which takes a depth scan and a library of shapes as input and produces a reconstruction of the shape implied by the scan using parts from the shape library.
This approach relies heavily on having access to a library of complete shapes drawn from the same semantic category as the input scan; in contrast, our algorithm assumes only a library of isolated parts.
As a consequence, our algorithm also supports stylized reconstructions of shapes using e.g. parts from other semantic shape categories.

%% file: 03-method.tex
\section{Method}

\begin{figure*}[t!]
    \centering
    \includegraphics[width=0.95\linewidth]{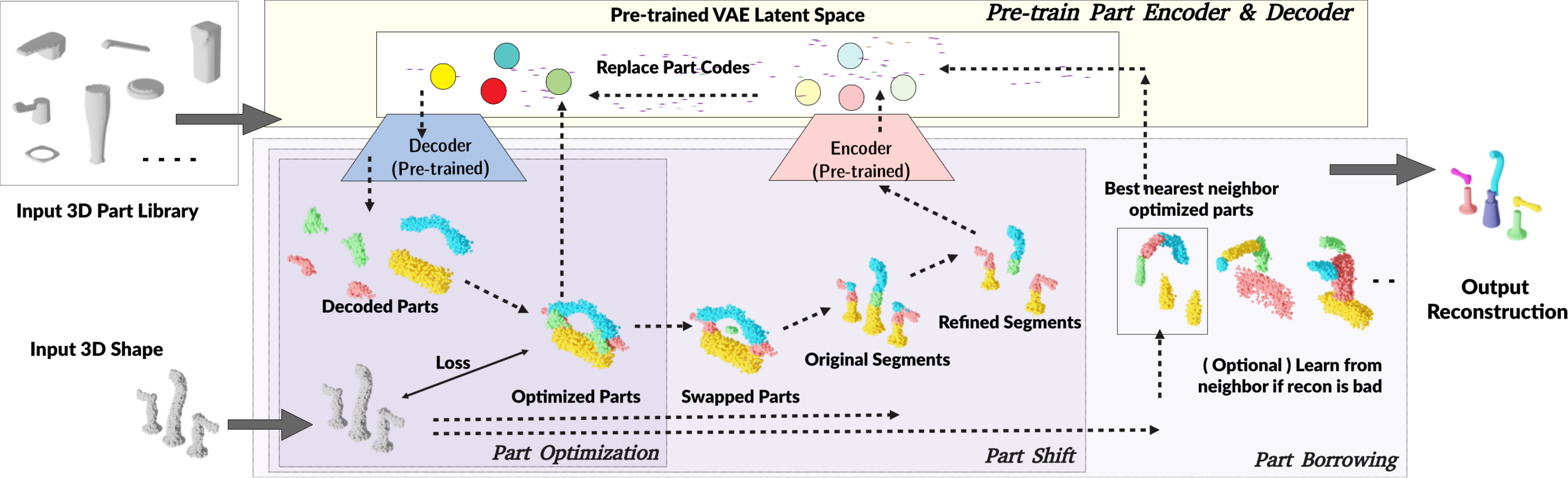}
    \caption{System Overview: Our system takes a target 3D volumetric point cloud $\mathbf{T}$ and a library $\mathcal{B}$ of parts as input and outputs a set of transformed parts $\mathcal{P}$ from the part library which approximate $\mathbf{T}$. It first pre-trains a variational autoencoder (VAE) to project all parts into a continuous latent space.
    This allows it to turn the combinatorial part retrieval problem into a continuous optimization problem which proceeds in three phases: \textbf{Phase I: Part Optimization}, \textbf{Phase II: Part Shift} and \textbf{Phase III: Part Borrowing}.
    Phase I directly optimizes part latent codes $\mathbf{e}$, translations $\mathbf{t}$, and rotations $r$ to reconstruct the target shape.
    Phase II segments the input target shape using the optimized parts from Phase I and re-projects them back to the latent space.
    Phase III is an optional phase which borrows good part decompositions from other well-reconstructed similar shapes.
    When the optimization converges, real parts from the part library are retrieved.
    Note: This figure shows the example with symmetry constraints, so each part has a reflected duplicate.
    }
    
    \label{fig:pipeline}
\end{figure*}

Our system takes as input either a single target 3D shape $\mathbf{T}$ to reconstruct, or a collection of target 3D shapes $\mathcal{T}$ from the same semantic category; as we will show, by using a shape collection our system can take advantage of within-category shape similarity.
We assume all 3D target shapes are represented as volumetric point clouds, which can be obtained from various sources (e.g. sampling the interior volume of a mesh or of a learned neural implicit shape representation~\cite{chen2019learning,Park_2019_CVPR}.
In addition to the target shapes, our system also takes as input a library of 3D parts $\mathcal{B}$ that it will use to reconstruct the target shape(s).
The output of our system is, for each target shape $\mathbf{T}$, a collection of transformed parts $\mathcal{P}$ from the part library that approximates the target shape. Figure~\ref{fig:pipeline} shows a schematic overview of our approach, and Algorithm~\ref{alg:framework} provides pseudocode.

Our method begins by pre-training a variational autoencoder on all the parts in the part library (Section~\ref{subsec:vae}).
This VAE's continuous latent space helps turn the discrete combinatorial search problem of part retrieval into a tractable continuous optimization problem.

The heart of our algorithm is an iterative optimization process, composed of three phases structured as nested loops.
In \textbf{Phase I: Part Optimization}, for multiple parts $\mathbf{P}_i$ the algorithm directly optimizes the VAE latent code $\mathbf{e}_i$, translation $\mathbf{t}_i$, and rotation around world up-axis $r_i$ such that the parts reconstruct the target shape well (Section~\ref{ssubsec:part_optimize}).
In \textbf{Phase II: Part Shift}, the algorithm segments the target shape $\mathbf{T}$ into regions using the optimized parts $\mathcal{P}$ from Phase I and then re-projects them back to the latent space using the pre-trained VAE encoder to form the new initial state for the next iteration of Phase I (Section~~\ref{ssubsec:part_shift}).
By doing this, Phase II helps Phase I to escape from spurious local optima; its operation is analogous to the mode-seeking before of the mean shift algorithm or the expectation maximization algorithm.
The algorithm also contains an optional \textbf{Phase III: Part Borrowing}, which it can use when a collection of target shapes $\mathcal{T}$ is available (Section~\ref{ssubsec:part_borrow}).
For a given target shape $\mathbf{T}$ that is currently not well-reconstructed by its optimized parts, this phase copies (or borrows) the optimization state $\{ (\mathbf{e}_i, \mathbf{t}_i, r_i) \}$ from some other target shape $\mathbf{T}' \in \mathcal{T}$ where $\mathbf{T}'$ is geometrically similar to $\mathbf{T}$.
This phase further helps Phase I to escape from local optima.

To determine the number of parts to use for each shape, our system runs this iterative optimization process with different numbers of parts $k \in \mathcal{K}$, returning the $k$ for which it achieved the best reconstruction (Section~\ref{subsec:number_of_parts}).
Finally, it retrieves parts from the part library which are the closest match to the continuous latent space parts produced by the optimization (Section~\ref{subsec:finalize_retrieval}).

The process described thus far can be run independently for each new shape presented to the system.
Alternatively, when given a collection of target shapes $\mathcal{T}$, the system can run the above optimization procedure on all of them as a preprocess and treat the output as a ``training set.''
Given a new target shape, it can retrieve the geometric nearest neighbor of this target in its training set, initialize optimization using that part decomposition, and execute a short optimization run to refine the decomposition to better fit the new target shape (Section~\ref{sec:inference}).
We also considered training a neural network on the ``training set'' to perform amortized inference of part decompositions, but we found that it did not perform as well as our nearest neighbor retrieval + optimization inference scheme (see supplemental for details).

\subsection{Part VAE}
\label{subsec:vae}

To turn the combinatorial search problem of part retrieval into a continuous optimization problem, we construct a continuous latent space of part geometries by training a variational autoencoder (VAE)~\cite{kingma2014auto} on all the parts in the input library.
Part meshes are first converted into volumetric point clouds by sampling their interior with 512 points.
We use a 4-layer PointNet~\cite{qi2017pointnet} as the encoder network, a 64-dimensional latent space, and a 3 layer MLP decoder which produces point clouds with 512 points.
% \xx{We use a PointNet~\cite{qi2017pointnet} as the encoder network which contains 4 convolutional layers, it takes in a $n_{in} \times 3$ volumetric point cloud and output a latent gaussian center vector $\boldsymbol{\mu} \in R^{64}$ together with a gaussian variance vector $\boldsymbol{\sigma} \in R^{64}$. The decoder net is a MLP which contains 3 fully connected linear layers, it takes in a sampled latent code $\textbf{e} \in R^{64}$ and output a $n_{out} \times 3$ volumetric point cloud.}
The VAE is trained using a combination of Chamfer distance reconstruction loss and the standard KL divergence latent space regularization loss. Once trained, the weights of the encoder and decoder are fixed for all the subsequent processes.

\subsection{Iterative Part Optimization}
\label{subsec:retrieval}

Given the part latent space, the system seeks to optimize for the latent codes $\mathbf{e}_i$ and poses $(\mathbf{t}_i, r_i)$ of a set of parts $\mathcal{P} = \{ \mathbf{P}_i | i \in \{1 \ldots k \} \}$ such that the latent codes, when decoded and posed, reconstruct the target shape.
This optimization proceeds in three stages, organized in a nested loop.

\begin{algorithm}
\small
\begin{algorithmic}[1]
\Input
\Desc{Target shape $\mathbf{T}$, Other target shapes $\mathcal{T}$, Part library $\mathcal{B}$} \Desc{Possible numbers of parts $\mathcal{K}$}
%\Desc{Nested loop counts $n_1, n_2, n_3$} 
\EndInput
\Output
\Desc{Retrieved and assembled parts $\mathcal{P}$ for target $\mathbf{T}$ }
\EndOutput
\Procedure {}{}
\State $\text{Encoder}, \text{Decoder} \gets \text{PreTrain}(\mathcal{B})$
\For {$k \in \mathcal{K}$} \Comment{executed in parallel}
\State $\mathbf{e}_i \gets \text{rand}() $ \Comment{part latent code, $i \in k$} 
\State $\mathbf{t}_i \gets \text{rand}() $ \Comment{part translation, $i \in k$} 
\State $r_i \gets \text{rand}()$ \Comment{part rotation, $i \in k$} 
\For {$i_3 \in n_3$} 
\For {$i_2 \in n_2$}
\For {$i_1 \in n_1$}
\State $\mathbf{e}_i, \mathbf{t}_i, r_i \gets \text{Optimize}(\mathbf{e}_i, \mathbf{t}_i, r_i, \mathbf{T})$
\EndFor
\State $\mathbf{e}_i, \mathbf{t}_i, r_i \gets \text{Shift}(\mathbf{e}_i, \mathbf{t}_i, r_i, \mathbf{T})$
\EndFor
\State $\mathbf{e}_i, \mathbf{t}_i, r_i \gets \text{Borrow}(\mathbf{e}_i, \mathbf{t}_i, r_i, \mathbf{T}, \mathcal{T})$
\EndFor
\State $\mathbf{D}_i^k \gets \text{Decode}(\mathbf{e}_i)$
\State $\mathbf{D}_i^k \gets \text{Pose}(\mathbf{D}_i^k, \mathbf{t}_i, r_i)$
\State $\mathbf{P}_i^k \gets \text{Retrieve}(\mathbf{D}_i^k, \mathcal{B})$
\State $\mathcal{P}^k \gets \bigcup_i \mathbf{P}_i^k$
\EndFor

\State \Return $\text{ChooseK}(\bigcup_{k \in \mathcal{K}} \mathcal{P}^k)$
\EndProcedure
\caption{
    Iterative Part Optimization
    }
\label{alg:framework}
\end{algorithmic}
\end{algorithm}

\subsubsection{Phase I: Part Optimization} \label{ssubsec:part_optimize}

In Phase I, the system directly optimizes the latent codes and poses of the parts via gradient-based optimization.
For each target shape $\textbf{T}$, this phase starts with $k$ randomly initialized latent codes $\textbf{e}_i \in \reals^d$,  $k$ rotation angles about the world up axis $r_i \in \reals$, and $k$ translation vectors $\textbf{t}_i \in \reals^3$.
We don't consider scaling in our system, as scaling can warp the geometry of a high-quality part from the part library; this may be unacceptable for some applications. During the optimization, each latent part code $\textbf{e}_i$ is decoded through the pre-trained decoder to produce a decoded volumetric point cloud $\textbf{D}_i$. This point cloud is then rotated and translated according to $r_i$ and $\textbf{t}_i$. The values of these variables are iteratively updated by the gradient acquired from an objective function $\mathcal{L}$ which consists a target reconstruction loss and a part non-overlap collision loss, i.e. $\mathcal{L} = \mathcal{L}_\text{recon} + \mathcal{L}_\text{overlap}$. 
% Please see supplement for more information.

\paragraph*{Target Reconstruction Loss:}
To approximate the input shape, a collection of decoded part volumetric point clouds $\mathcal{D} = \bigcup_i \mathbf{D}_i$ should match the volumetric point cloud of the target shape $\mathbf{T}$.
We use chamfer distance between these two volumetric point clouds to measure this matching:
$$
\mathcal{L}_\text{recon} = d_\text{chamfer}(\mathcal{D}, \mathbf{T})
$$

\paragraph*{Part Overlap Penalty Loss:} 
The optimized parts should not only cover the target shape; they also should not overlap with each other.
Since decoded parts can have complex geometry, it would be non-trivial and time consuming to compute a bounding proxy for them for differentiable collision checking.
We thus designed a collision penalty term based on the volumetric point cloud representation.
The collision penalty between two decoded parts $\mathbf{D}_a$ and $\mathbf{D}_b$ is computed by summing point-to-point pairwise distances between all points in those parts. If the distance $||\mathbf{p}_i - \mathbf{p}_j||_2$ between a pair of points $ \mathbf{p}_i \in \mathbf{D}_a, \mathbf{p}_j \in \mathbf{D}_b$ is below a threshold $\tau=0.1$, an overlap penalty is added as $\tau - ||\mathbf{p}_i - \mathbf{p}_j||_2$, otherwise the penalty is zero. The final overlap penalty is the average penalty across all point pairs.
%Figure~\ref{fig:collision} visually illustrates how this loss operates.
$$
\mathcal{L}_\text{overlap} = \frac{1}{|\mathbf{D}_a||\mathbf{D}_b|} \sum_{\mathbf{p}_i \in \mathbf{D}_a}\sum_{\mathbf{p}_j \in \mathbf{D}_b} \max(0, \tau - ||\mathbf{p}_i - \mathbf{p}_j||_2)
$$

% \begin{figure}[t!]
%     \centering
%     \includegraphics[width=1.0\linewidth]{figs/collision.pdf}
%     % \vspace{-1.5em}
%     \caption{Illustrating the part overlap penalty loss $\mathcal{L}_\text{penalty}$ The collision penalty is computed by summing point-to-point distances between two decoded part volumetric point clouds $\textbf{D}_a$ and $\textbf{D}_b$. If the distance between a pair of points from different point clouds is below a threshold $\tau$, an overlap penalty is added as $\tau - ||\textbf{p}_i - \textbf{p}_j||_2$. 
%     }
%     \label{fig:collision}
% \end{figure}

\subsubsection{Phase II: Part Shift} \label{ssubsec:part_shift}

\begin{figure*}[t!]
    \centering
    \includegraphics[width=1.0\linewidth]{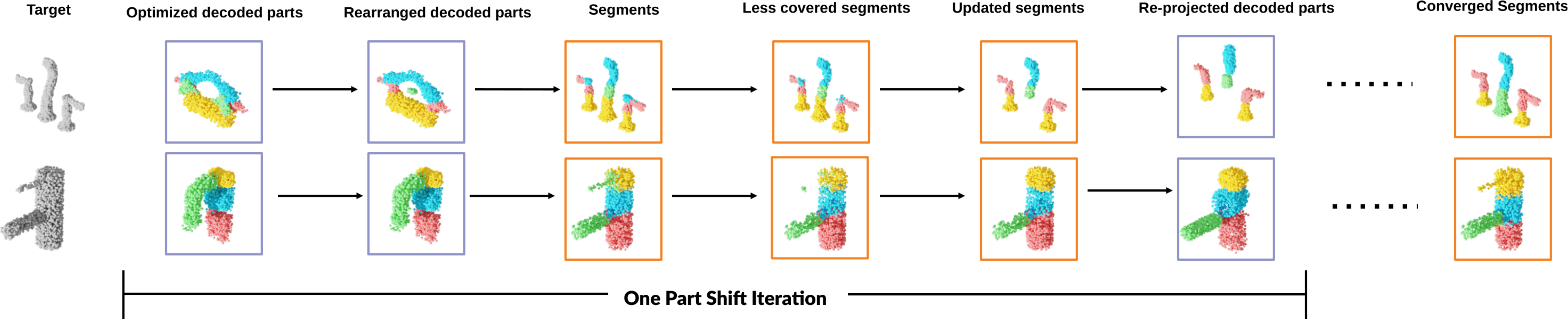}
    % \vspace{-1.5em}
    \caption{Phase II Part Shift.
    Given decoded and posed parts $\{ \mathbf{D}_i \}$ (2nd column), the algorithm segments the target point cloud (1st column) $\mathbf{T}$ based on which part each point is closest to (4th column).
    It then discards some of the points which were very close to their assigned part (5th column) and then retains only one connected component for each segment--whichever is farthest away from any other segment (6th column).
    These segments are finally re-encoded into the part latent space to produce the input for the next round of Phase I optimization (7th column shows what these parts look like when decoded).
    When the entire optimization procedure has converged, the output segments from this process will be the same as the original segments (last column).
    \textcolor{purple}{Purple} boxes denote decoded part point clouds; \textcolor{orange}{orange} boxes denote segmentations of the target point cloud using those decoded parts.
    }
    \label{fig:shift}
\end{figure*}

The optimization problem solved by Phase I is non-convex with many local minima.
In particular, it can be sensitive to different initializations of part latent codes and poses.
These local minima manifest as regions of the target shape which are uncovered by any part.
Phase II of the algorithm is designed to help the optimization escape such local optima.
It does so by shifting the optimized parts $\mathbf{D}_i$ from Phase I around the target to capture more regions in the target shape.

Given the optimized latent code $\mathbf{e}_i$ and pose $(\mathbf{t}_i, r_i)$ for each part, the algorithm produces the decoded part point cloud $\mathbf{D}_i$ via the pre-trained decoder and poses it according to $(\mathbf{t}_i, r_i)$.
As shown in Figure~\ref{fig:shift} 2nd column, these parts may have become stuck in a local optimum, missing large regions of the target shape.
Thus, instead of using the decoded parts directly, the algorithm instead uses them to segment the target point cloud $\mathbf{T}$ into a set of segments $\mathcal{S} = \bigcup_i \mathbf{S}_i$, where $\mathbf{S}_i$ consists of all points from $\mathbf{T}$ which are closer to  $\mathbf{D}_i$ than to any other decoded part (Figure~\ref{fig:shift}, 4th column).
This guarantees that each every point in the target point cloud is assigned to one of the optimized parts.

However, the shapes of these segments may be implausible; for example, the green segment in the Figure~\ref{fig:shift} bottom row example is divided into two connected components, which is implausible for a single physical part.
Ideally, we want each part to consist of a single connected component, and for those parts to cover the target point cloud.
Thus, for each segment, the algorithm discards all but one of its connected components (Figure~\ref{fig:shift} column 6).
Specifically, it chooses the connected component which is farthest from any of the other segments, so as to focus on the most distinct part of the target point cloud covered by this segment.
We have found that this connected component selection step performs most consistently when the segmented target point cloud $\mathcal{S}$ is first filtered to remove points which were very well-covered by one of the original decoded parts $\mathbf{D}_i$ (i.e. this helps the algorithm shift each segment to include the less-well-covered regions of the target shape).
Specifically, the algorithm computes the distance between each point in the target point cloud $\mathbf{T}$ and its closest point in the part $\mathbf{D}_i$ to which it is assigned, and discards the points with the smallest $p=30\%$ of these distances (Figure~\ref{fig:shift} column 5).

We now have one connected component $\mathbf{C}_i$ for each segment $\mathbf{S}_i$ of the segmented target point cloud $\mathcal{S}$.
The algorithm normalizes the pose of each connected component by translating its centroid to the origin and rotating it such that the axes of its minimum volume bounding box align with the world axes.
This normalized part is re-encoded to the part latent space to produce a new part latent code $\mathbf{e'}_{i}$.
This code, along with the inverse of the pose normalization transformation $(\mathbf{t'}_i, r'_i)$, become the optimization variables for the next round of Phase I (Figure~\ref{fig:shift} column 7).

Finally, we have found that it can be helpful to add an additional part rearrangement step to the beginning of Phase II (Figure~\ref{fig:shift} column 3).
This step directly replaces one of the optimized parts $\mathbf{D}_i$ with the segment $\mathbf{S}_i \in \mathcal{S}$ which is least covered by any of the parts (call this $\mathbf{S^-}$), before proceeding with the rest of Phase II.
Specifically, the algorithm iterates over all optimized parts $\mathbf{D}_i$ and tries to replace each part with the $\mathbf{S}^-$.
The algorithm accepts whichever replacement results in the greatest improvement in coverage of the target points by the resulting set of parts (if any does).

\subsubsection{Phase III: Part Borrowing} \label{ssubsec:part_borrow}
If the system is given a collection of target shapes $\mathcal{T}$ as input, this opens up the possibility of using similarities between target shapes to further improve part decompositions and escape from local optima.
In this optional phase, the system identifies all target shapes $\mathbf{T}$ that are currently not well-reconstructed and sets their optimization variables $\{(\mathbf{e_i}, \mathbf{t_i}, r_i)\}$ to the values of those copied from other geometrically-similar target shapes that are well-reconstructed.
The algorithm performs this update on the subset of target shapes $\mathcal{T}_\text{bad}$ whose reconstruction error is in the worst $60\%$ of all target shapes $\mathcal{T}$.
To quickly identify geometrically-similar targets, we pre-compute a target-to-target distance matrix $\mathbf{M}$, where $M_{ij}$ is the chamfer distance between target shapes $\mathbf{T}_i$ and $\mathbf{T}_j$.
% \dr{The details about how we optimize for translation, rotation and isotropic scaling to make these match better should be deferred to supplemental, I think.}
% \xx{We do not have scaling anymore. And for optimizing translation and rotation, it's same as phase I, we can describe more in details in supp for sure.}
For each $\mathbf{T} \in \mathcal{T}_\text{bad}$, the algorithm iterates over the $m=5$ nearest neighbors of $\mathbf{T}$ in $\mathcal{T} / \mathcal{T}_\text{bad}$ according to the distance matrix $\mathbf{M}$ and uses their current optimization variables $\{(\mathbf{e_i}, \mathbf{t_i}, r_i)\}$ to reconstruct $\mathbf{T}$.
If the resulting reconstruction error is within the best $10\%$ of reconstruction errors across $\mathcal{T}$, then these variable values become the new values for $\mathbf{T}$ for the next iteration of optimization Phase I.
Otherwise, $\mathbf{T}$'s variables are re-initialized to a random state.

\subsection{Final Part Retrieval} \label{subsec:finalize_retrieval}
\begin{figure}[t!]
    \centering
    \includegraphics[width=1.0\linewidth]{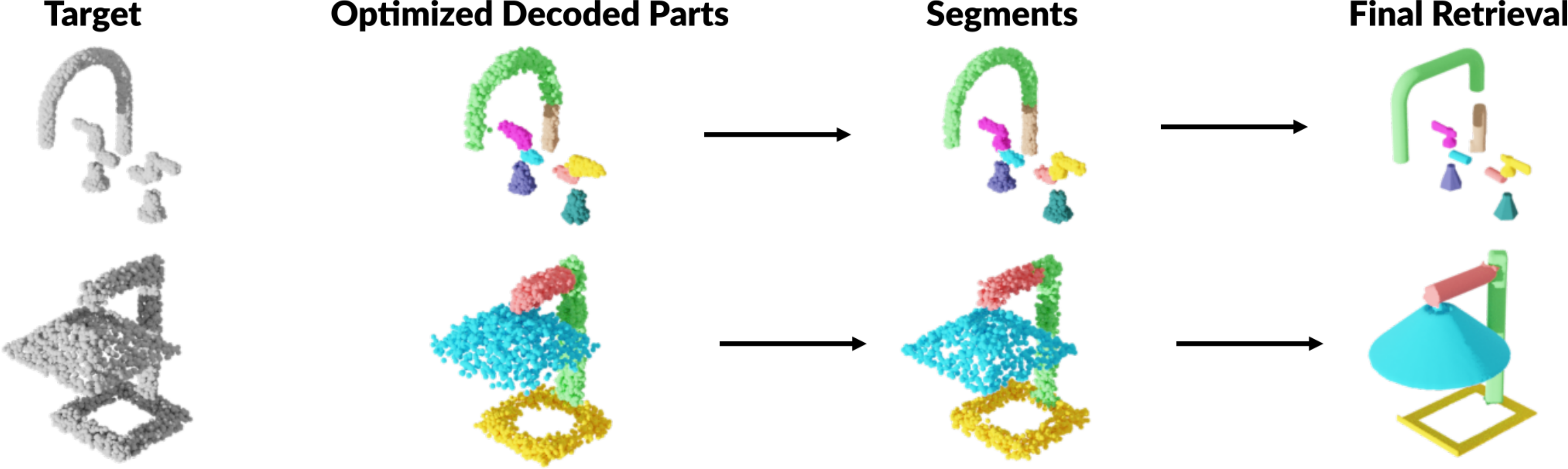}
    % \vspace{-1.5em}
    \caption{The final part retrieval phase. The target shape point cloud is segmented into regions based on which points are closest to which optimized part point cloud. Then, for each segment, the poses of $q$ parts from the part library are optimized to fit the segment. The part that matches the segment best is retrieved and used in the final part decomposition.}
    \label{fig:retrieval}
\end{figure}

The output from the above three phases is the decoded and posed part volumetric point clouds $\{ \mathbf{D}_i \}$.
To convert these point clouds into actual parts from our part library, the algorithm first segments the target shape $\mathbf{T}$ into segments $\mathcal{U} = \bigcup_i \mathbf{U}_i$, where $\mathbf{U}_i$ consists of all points from $\mathbf{T}$ which are closer to $\mathbf{D}_i$ than to any other decoded part (as in Phase II).
For each segment $\mathbf{U}_i$, it then considers the $q$ nearest neighbors of $\mathbf{U}_i$ in the part library $\mathcal{B}$ by Chamfer distance.
We could use our pre-trained part latent space to accelerate this nearest neighbor search using standard Euclidean approximate nearest neighbor methods.
However, for fairer comparison with methods which do not build such a latent space, in our experiments we set $q = |\mathcal{B}|$, i.e. a linear scan across the entire part library.
The algorithm then optimizes poses $(\mathbf{t}, r)$ for each of these $q$ candidate parts and takes the one which achieves minimal volumetric Chamfer distance between the part and the target shape segment as the final retrieved part.
Figure~\ref{fig:retrieval} visualizes this process.

\subsection{Leveraging Symmetry}

Many manufactured shapes exhibit symmetries; our algorithm takes advantage of this property to improve its reconstructions.
As a preprocess, we automatically detect the bilateral symmetry plane (if there is one) of each target shape.
If a target shape $\mathbf{T}$ has a bilateral symmetry, then whenever the algorithm decodes and poses a part point cloud $\mathbf{D}_i$, it also produces a symmetric copy of this part $\mathbf{D}_{i+k}$ by duplicating $\mathbf{D}_i$ and reflecting it about the symmetry plane.
We disable collision checking between pairs of symmetric parts in Phase I.
At the beginning of Phase II, if two symmetric parts are in contact, then the algorithm merges them together into one part.
The same merge step is also applied before performing final part retrieval.

\subsection{Choosing the Number of Parts}
\label{subsec:number_of_parts}
\begin{figure}[t!]
    \centering
    \small
    \setlength{\tabcolsep}{1pt}
    \begin{tabular}{rcccccc}
        \includegraphics[width=0.99\linewidth]{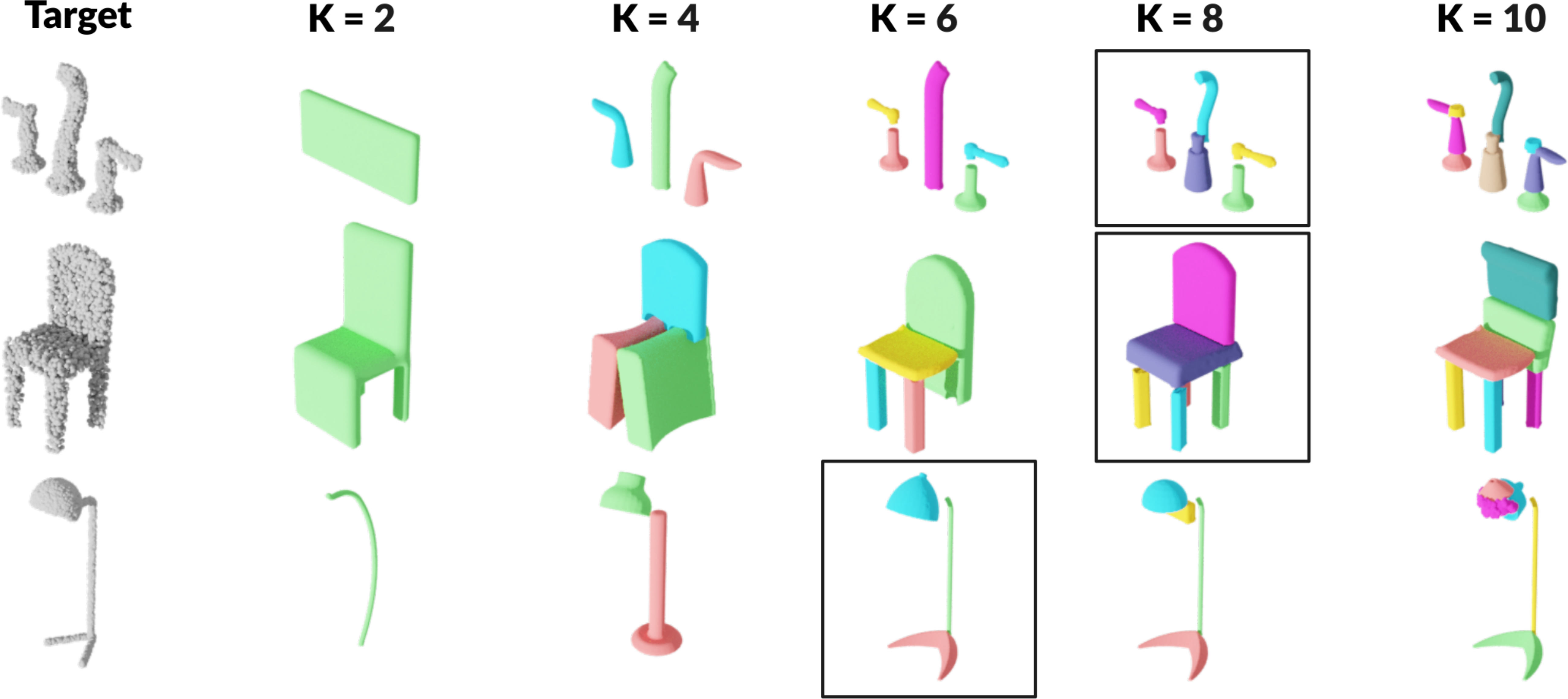}
    \end{tabular}
    \caption{Choosing the number of parts $k$. The value of $k$ which achieves the best balance between reconstruction quality and complexity (i.e. number of parts) is chosen. Note: some reconstructions for a particular value of $k$ may contain fewer than $k$ parts because symmetric pairs of parts are merged if they are in contact.}
    \label{fig:k}
\end{figure}

Due to the structural complexity of the input shapes, different numbers of parts $k$ can lead to the best reconstruction for different shapes.
Thus, our algorithm runs iterative multi-phase part optimization for different numbers of parts $k \in \mathcal{K} = [2, 4, 6, 8, 10]$.
Then for each target shape $\mathbf{T}$, it selects the $k$ whose optimized parts $\mathcal{P}^k$ give the minimum value of a penalty function $d_\text{chamfer}((\mathcal{P}^k), \mathbf{T}) + \alpha |(\mathcal{P}^k)|$.
This penalty function is a linear combination of a reconstruction error term and a complexity penalty term (which prefers fewer parts, all else being equal).
We set $\alpha = 1.5^{-4}$ in our experiments.
Figure~\ref{fig:k} shows examples of target shapes reconstructed with different numbers of parts.

\subsection{Amortized Inference}
\label{sec:inference}

If our system receives a set of target shapes $\mathcal{T}$ as input, then the optimized part decompositions for these shapes can facilitate amortized inference on new target shapes.
Given a never-before-seen target shape $\mathbf{T}$, the system first finds its nearest neighbor in $\mathcal{T}$, runs another round of optimization Phase I with $\mathbf{T}$'s variables initialized to those of the retrieved nearest neighbor, and finally retrieves parts for $\mathbf{T}$ from the part library.
This short optimization is considerably faster than optimizing for $\mathbf{T}$ from scratch and is also more likely to avoid spurious local optima.
% As you may notice, the logic behind inference is the same as the phase III(Part Borrow), they both rely on taking and optimizing the variable from a well-reconstructed neighbor shape.

%% file: 04-results.tex
\section{Results \& Evaluation}

We evaluate our method by using it to reconstruct target shapes from three categories from the PartNet dataset~\cite{PartNet}: Chair, Lamp and Faucet.
For each category, we split the shapes into source shapes (from which we construct part libraries), training shapes, and testing shapes.
For each category, the sizes of the source set / part library / training set / test set are: Chair (601/1766/1000/120), Lamp (601/1985/1000/120), Faucet (151/780/250/100).
For all three categories, 100 randomly selected shapes from the training set are used when evaluating training set performance.
The part libraries are built using the ground-truth part segmentations from PartNet.

On a machine with an Intel(R) i9-9900K CPU and a NVIDIA RTX 2080 Ti GPU, our algorithm takes around 12 to 20 seconds on average to optimize the part decomposition for one shape from scratch.
In amortized inference mode, it takes around 5 to 8 seconds.
For Phase I optimization, we use the Adam optimizer~\cite{Adam} with learning rate 0.008.

\paragraph*{Comparison to baselines:}
In this section, we compare our method against several baselines in terms of how well their part decompositions reconstruct the target shape.
We evaluate the reconstruction quality using two metrics: \textbf{Surface Chamfer Distance (SCD)} and \textbf{Volume Chamfer Distance (VCD)}, i.e. chamfer distance evaluated between surface point cloud and volume point cloud shape representations, respectively.
We compare against the following baselines:
\begin{packed_itemize}
    \item \textbf{Joint Learning of 3D Shape Retrieval and Deformation (JRD)}~\cite{uy-joint-cvpr21} learns to reconstruct shapes by retrieving and deforming 3D shapes from a database. It requires complete 3D shapes, whereas ours requires only parts. Since our method does not deform parts, we compare against a version of JRD with deformation disabled.
    \item \textbf{Neural Parts (NP)}~\cite{Paschalidou2021CVPR} learns to reconstruct shapes by deforming a set of sphere meshes. In their paper, the authors focus on reconstruction from images; we swap out their image encoder with a 3D encoder (a volumetric CNN) for fair comparison to our method, which receives 3D data as input. Also motivated by fair comparison to our method, we primarily evaluate NP by using its predicted meshes to retrieve parts from our part library (using the same logic from Section~\ref{subsec:finalize_retrieval}).
    \item \textbf{Brute Force (BF)} is a naive baseline which randomly samples $k$ parts from the part library and optimizes their poses to best reconstruct the target shape. For fair comparison, we let this process run iteratively for as long as our method took to run, and then take the best result.
\end{packed_itemize}

Table~\ref{tab:comparison} shows the results of comparing our method to JRD and NP.
Our method outperforms both across all metrics.
JRD can only retrieve whole shapes, which may not structurally match the target shape.
The gap between our method and NP is largest on more geometrically-complex categories such as Faucet and Lamp, where knowledge of the irregular potential part geometries via the part latent space helps our method.
Figure ~\ref{tab:np_vs_ours} show qualitative results for NP and Ours. 
See the supplemental material for JRD results and other comparison experiments.

\begin{figure}[t!]
    \centering
    \includegraphics[width=1.0\linewidth]{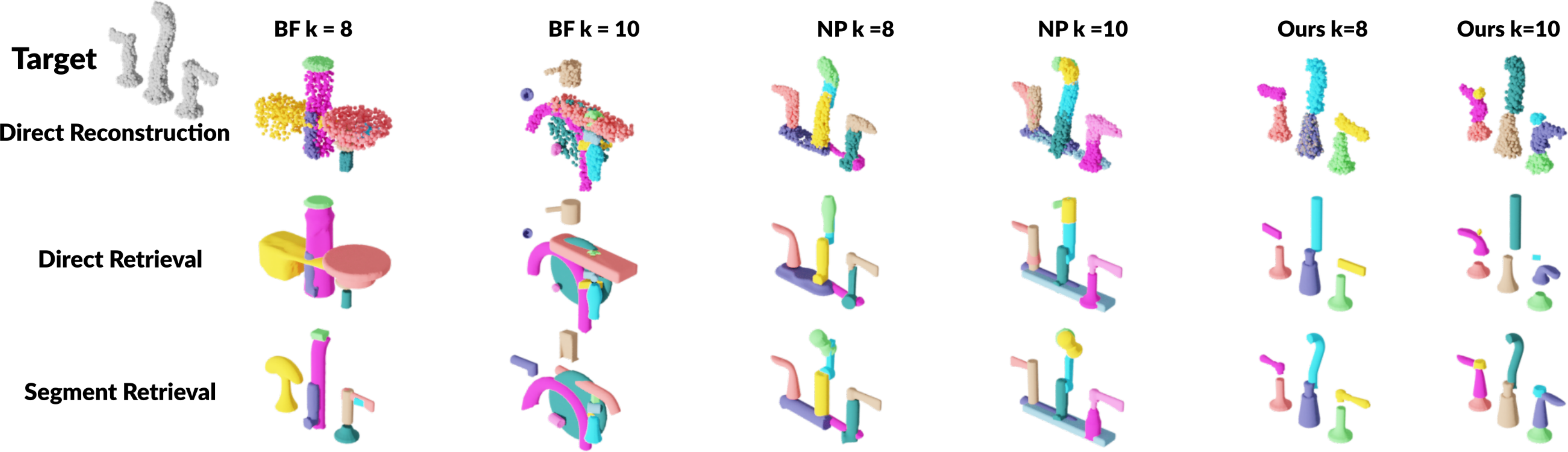}
    % \vspace{-1.5em}
    \caption{Comparing our method with Brute Force (BF) and Neural Parts (NP) with different final decomposition output formats.
    NP's direct reconstruction format is a mesh; here we visualize it as a point cloud via sampling.}
    \label{fig:output_comp}
\end{figure}

We also conducted an experiment evaluating the impact of part output representations on reconstruction quality, comparing our method to NP and BF.
The representations are: using the predicted geometry from the method directly (\textbf{Direct Recon}); retrieving parts which best match the predicted geometry (\textbf{Direct Retrieval}); retrieving parts which best match the segmentation of the target shape induced by the predicted geometry (\textbf{Segment Retrieval}).
For Ours, the predicted geometry is the part point clouds output by the pre-trained decoder; for NP, it is the part meshes output by their network; for BF, it is a point sampling of the randomly-retrieved parts.
Table~\ref{tab:bf_np_ours} shows quantitative results, and Figure~\ref{fig:output_comp} shows qualitative results.
For BF, Direct Recon and Direct Retrieval are quite bad, as BF is unlikely to find a good part decomposition by random sampling; it performs better for Segment Retrieval, as the segments can happen to correspond to an actual part in the part library.
NP and Ours give reliable results across all output formats, with ours consistently outperforming NP.

\begin{table}[t!]
    \centering
    \setlength{\tabcolsep}{2pt}
    \scriptsize
    \begin{tabular}{lccccc}
        \toprule
        \textbf{Category} & \textbf{Method} & \textbf{Train (SCD) $\downarrow$} & \textbf{Train (VCD) $\downarrow$} & \textbf{Test (SCD) $\downarrow$} & \textbf{Test (VCD) $\downarrow$}
        \\
        \midrule
        Lamp 
        & NP & 0.349  & 0.204 & 0.390  & 0.195
        \\
        & Ours & \textbf{0.307}  & \textbf{0.163} & \textbf{0.303}  & \textbf{0.163}
        \\
        \midrule
        Faucet
        & NP & 0.326 & 0.171 & 0.370  & 0.174
        \\
        & Ours & \textbf{0.256} & \textbf{0.135} & \textbf{0.288}  & \textbf{0.134}
        \\
        \midrule
        Chair 
        & JRD & 0.746 & 0.448 & 0.669 & 0.397 \\
        & NP & 0.495  & 0.233 & 0.547 & 0.240 \\
        & Ours & \textbf{0.470}  & \textbf{0.219} & \textbf{0.539 } & \textbf{0.234}
        \\
        \midrule
        Average
        & JRD & 0.746 & 0.448 & 0.669 & 0.397
        \\
        & NP & 0.390 & 0.203 & 0.436 & 0.203
        \\
        & Ours & \textbf{0.344} & \textbf{0.172} & \textbf{0.377} & \textbf{0.177}
        \\
        \bottomrule
    \end{tabular}
    \caption{Comparing our method to two baselines, JRD and NP, on reconstructing shapes from three PartNet categories. CD values are multiplied by 100. JRD is evaluated only on Chair because it does not provide data for the other categories.
    }
    \label{tab:comparison}
\end{table}

\begin{table}[t!]
    \centering
    \scriptsize

    \begin{tabular}{ccccc}
        \toprule
        \textbf{Method} & \textbf{Direct Recon$\downarrow$} & \textbf{Direct Retrieval$\downarrow$} & \textbf{Segment Retrieval$\downarrow$ }
        \\
        \midrule
        BF (VCD) & 0.680 & 0.750 & 0.182  
        \\
        NP (VCD) & 0.128 & 0.207  & 0.153  
        \\
        Ours (VCD) & \textbf{0.114} & \textbf{0.142} & \textbf{0.124}
        \\
        \bottomrule
    \end{tabular}
    \caption{Comparing our method with Brute Force (BF) and Neural Parts (NP) with different final decomposition output formats. CD values are multiplied by 100.
    }
    \label{tab:bf_np_ours}
\end{table}

\begin{figure}[t!]
    \centering
    \small
    \setlength{\tabcolsep}{0pt}
    \renewcommand{\arraystretch}{0}
    \newcommand{\resultimg}[1]{\includegraphics[trim={40pt 40pt 40pt 40pt},clip,width=0.15\linewidth]{#1}}
    \begin{tabular}{rcccccc}
        \raisebox{1.5em}{Targets} & 
        \resultimg{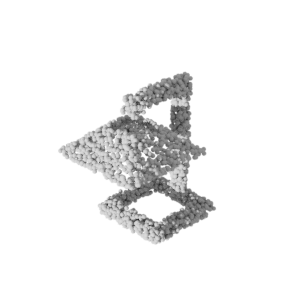} &
        \resultimg{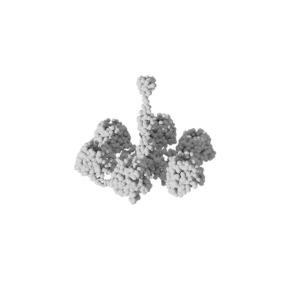} &
        \resultimg{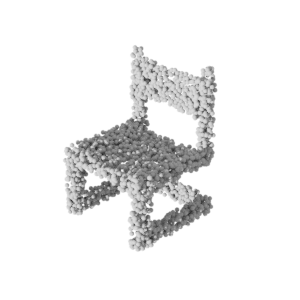} &
        \resultimg{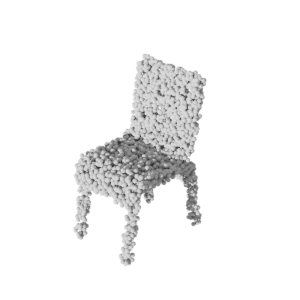} &
        \resultimg{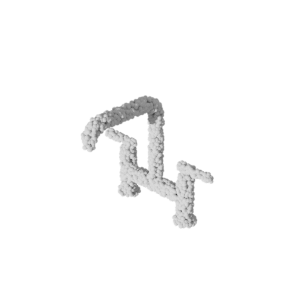} &
        \resultimg{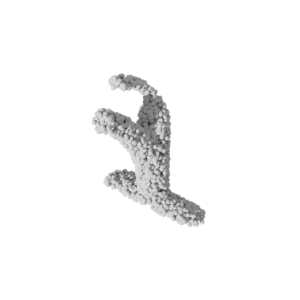} 
        \\
        \raisebox{1.5em}{NP} & 
        \resultimg{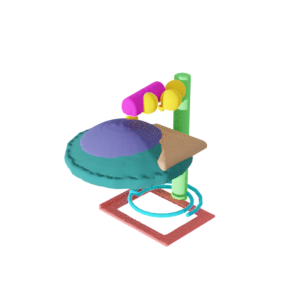} &
        \resultimg{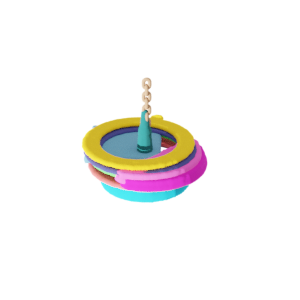} &
        \resultimg{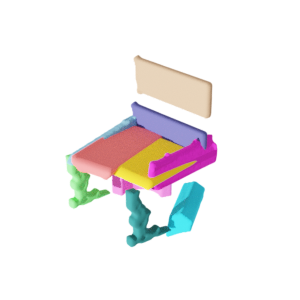} &
        \resultimg{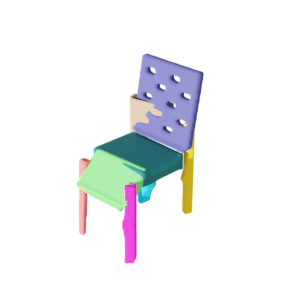} &
        \resultimg{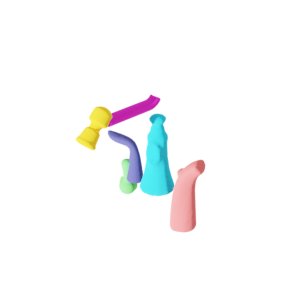} &
        \resultimg{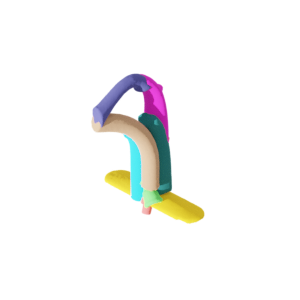} 
        \\
        \raisebox{1.5em}{Ours} & 
        \resultimg{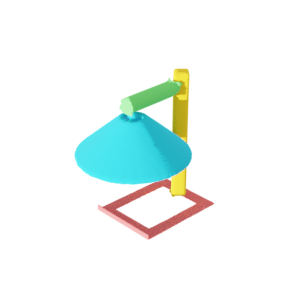} &
        \resultimg{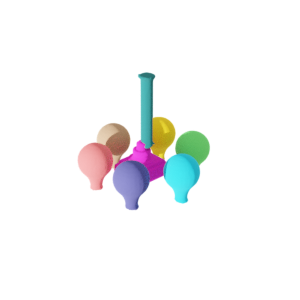} &
        \resultimg{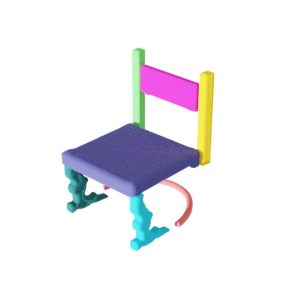} &
        \resultimg{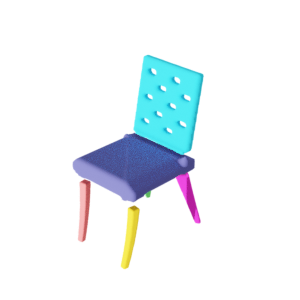} &
        \resultimg{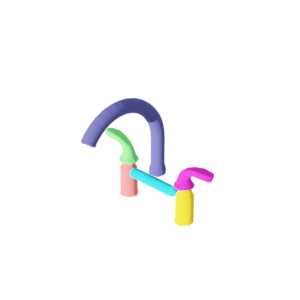} &
        \resultimg{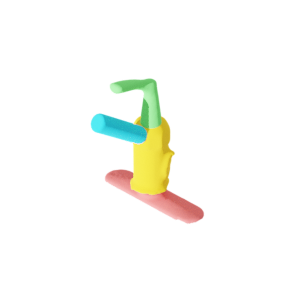} 
        \\
        \raisebox{1.5em}{Targets} & 
        \resultimg{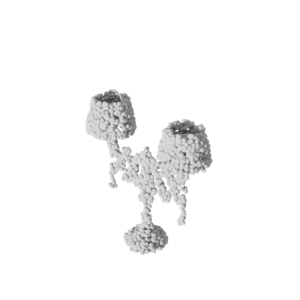} &
        \resultimg{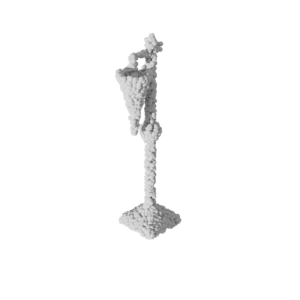} &
        \resultimg{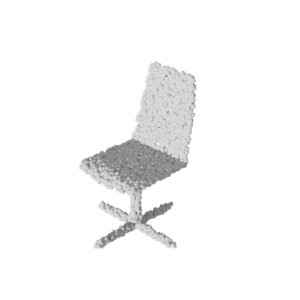} &
        \resultimg{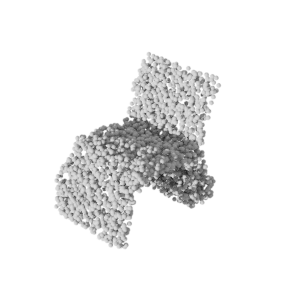} &
        \resultimg{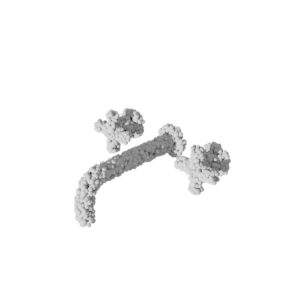} &
        \resultimg{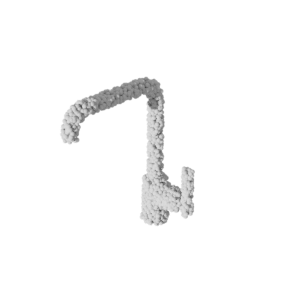} 
        \\
        \raisebox{1.5em}{NP} & 
        \resultimg{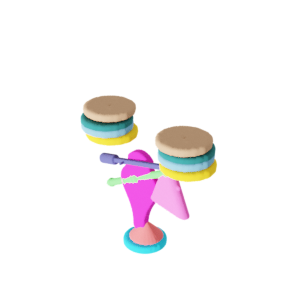} &
        \resultimg{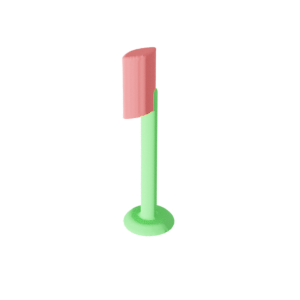} &
        \resultimg{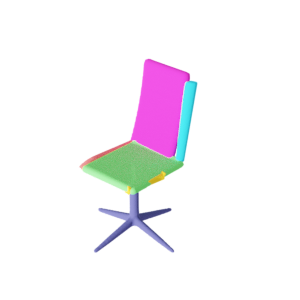} &
        \resultimg{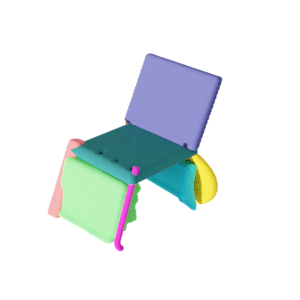} &
        \resultimg{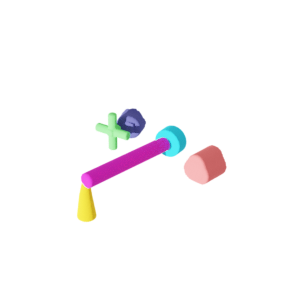} &
        \resultimg{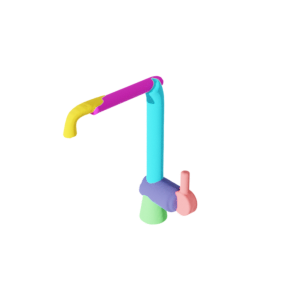} 
        \\
        \raisebox{1.5em}{Ours} & 
        \resultimg{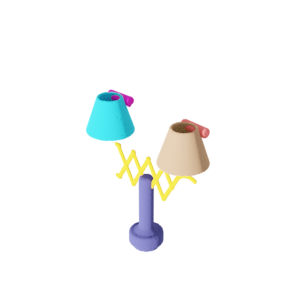} &
        \resultimg{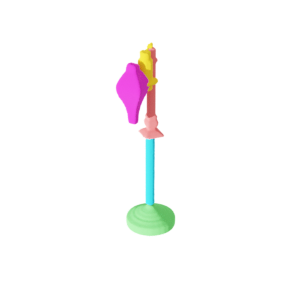} &
        \resultimg{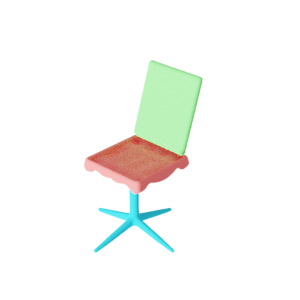} &
        \resultimg{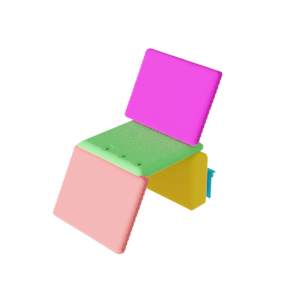} &
        \resultimg{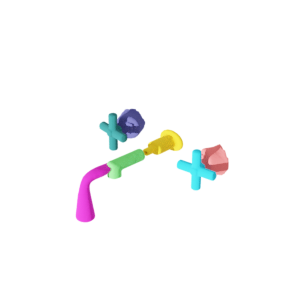} &
        \resultimg{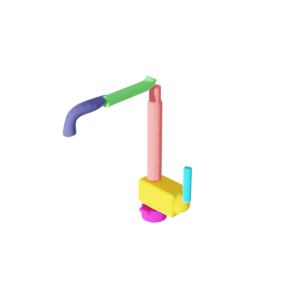} 
        
    \end{tabular}
    \caption{Neural Parts (NP) vs. Ours on reconstructing Lamps, Chairs and Faucets}
    \label{tab:np_vs_ours}
\end{figure}

\begin{figure}[t!]
    \centering
    \small
    \setlength{\tabcolsep}{0pt}
    \renewcommand{\arraystretch}{0}
    \newcommand{\resultimg}[1]{\includegraphics[trim={40pt 40pt 40pt 40pt},clip,width=0.17\linewidth]{#1}}
    \begin{tabular}{rccccc}
        \raisebox{1.5em}{Targets} & 
        \resultimg{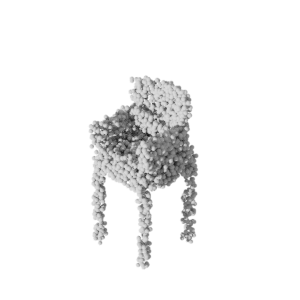} &
        \resultimg{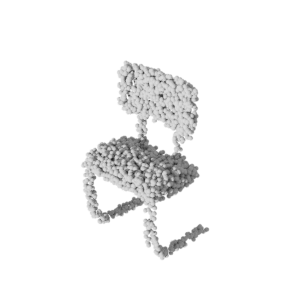} &
        \resultimg{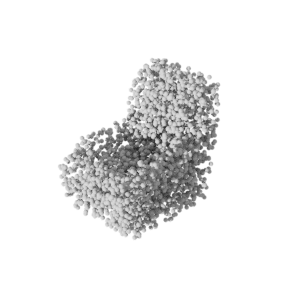} &
        \resultimg{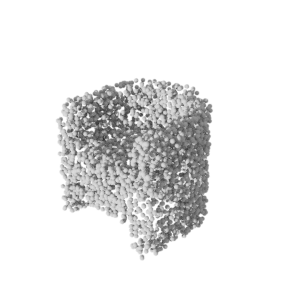} &
        \resultimg{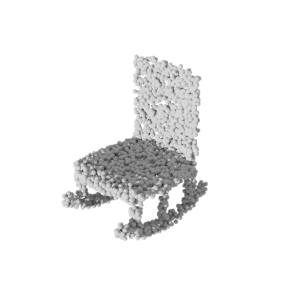}
        \\
        \raisebox{1.5em}{Ours} & 
        \resultimg{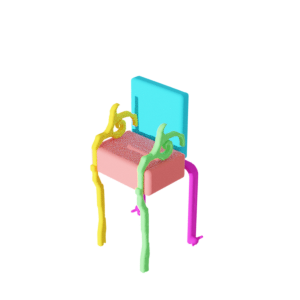} &
        \resultimg{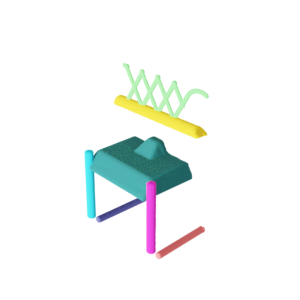} &
        \resultimg{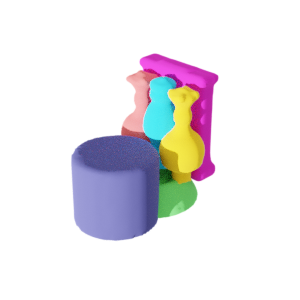} &
        \resultimg{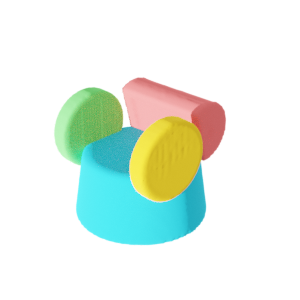} &
        \resultimg{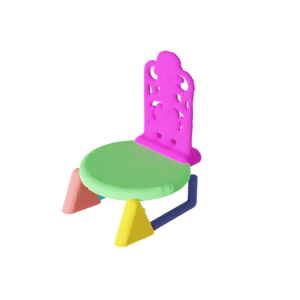}
        \\
        \midrule
        \raisebox{1.5em}{Targets} & 
        \resultimg{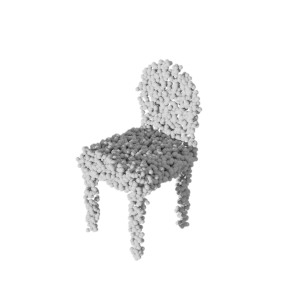} &
        \resultimg{figs/results/2968target.png} &
        \resultimg{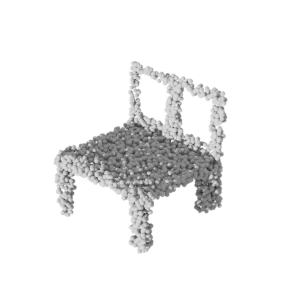} &
        \resultimg{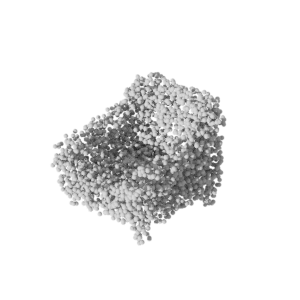} &
        \resultimg{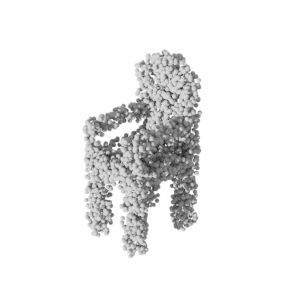} 
        \\
        \raisebox{1.5em}{Ours} & 
        \resultimg{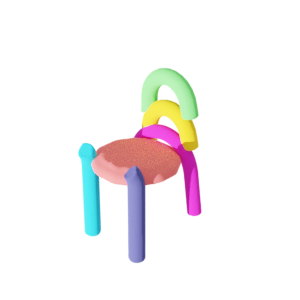} &
        \resultimg{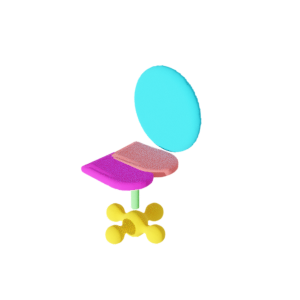} &
        \resultimg{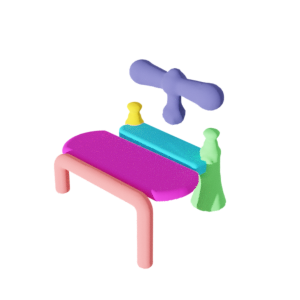} &
        \resultimg{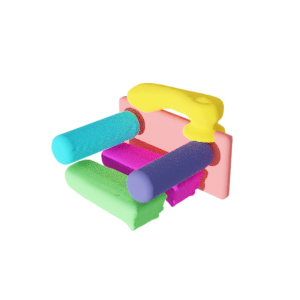} &
        \resultimg{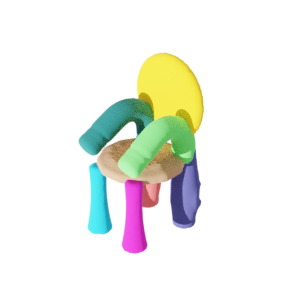} 
        \\
    \end{tabular}
    \caption{Cross Category Reconstruction: reconstructing Chairs from Lamp parts (top) or Faucet parts (bottom)}
    \label{fig:crosscat}
\end{figure}

\paragraph*{Cross-category reconstruction:}
Since our method does not require assembled shapes as training data, it reconstruct shapes out of different sets of parts than those from which they were originally built.
Figure~\ref{fig:crosscat} shows examples of reconstructing chairs out of lamp and faucet parts, respectively.
See the supplemental material more for results.

\paragraph*{Ablation study:}
Table~\ref{tab:phase_ablation} shows the results of ablating different optimization phases for the Faucet category.
Using all phases gives the best performance; Phase II matters more than Phase III.
See the supplement for additional ablations.
% The supplement which includes: (1) Experiments with different number of parts available in the part library. (2) Experiments with different number of input target shapes available. 

\begin{table}[t!]
    \centering
    \scriptsize

    \begin{tabular}{lccc}
        \toprule
        \textbf{Method} & \textbf{SCD$\downarrow$} & \textbf{ VCD$\downarrow$}
        \\
        \midrule
        Phase I & 0.337 & 0.175
        \\
        Phase I + Phase III & 0.328  & 0.171 
        \\
        Phase I + Phase II & 0.282  & 0.144 
        \\
        Phase I + Phase II + Phase III & \textbf{0.255} & \textbf{0.134} 
        \\
        \bottomrule
    \end{tabular}
    \caption{Ablation for phase components in our system. CD values are multiplied by 100.
    }
    \label{tab:phase_ablation}
\end{table}

%% file: 05-conclusion.tex
\section{Conclusion}

We presented an unsupervised algorithm which retrieves and places 3D parts from a given part library to reconstruct 3D target shapes.
Our approach uses a continuous relaxation of this combinatorial problem via a pre-trained part latent space, plus a multi-phase direct optimization scheme to fit part shape and poses while avoiding many of the worst local optimas.
Experimental evaluation shows that our approach outperforms strong baselines for retrieval-based reconstruction and unsupervised part decomposition.
Potential directions for improvement include introducing physical priors into the optimization to make reconstructions more physically plausible and supporting incomplete point clouds as input.  

%% file: supplement_append.tex
\section {Supplement Materials for Unsupervised 3D Shape Reconstruction by Part Retrieval and Assembly}

\subsection{Additional Implementation Details}

In this section we provide additional details for our method and experiments.

\paragraph{Method Details}

Please see Algorithms~\ref{alg:part_optimize}, \ref{alg:part_shift} and \ref{alg:part_borrow} for pseudo-code of Phase I (Part Optimization), Phase II (Part Shift) and Phase III (Part Borrowing), respectively.
For Phase II, we provide additional pseudo-code for helper functions $Swap()$ (Algo.~\ref{alg:swap}), $NNSegment()$ (Algo.~\ref{alg:nnsegment}), $Filter()$ (Algo.~\ref{alg:filter}) and $FathestConnectedComponent()$ (Algo.~\ref{alg:fcc}).
% Additional explanation for helper functions $Swap()$ algo ~\ref{alg:swap}, $NNSegment()$ algo ~\ref{alg:nnsegment}, $Filter()$ algo ~\ref{alg:filter} and $FathestConnectedComponent()$ algo ~\ref{alg:fcc} used in Phase II (Part Shift):
Intuitively, these helper functions perform the following operations:
Firstly, target to optimized parts distance matrix $\textbf{Q}$ is computed, with rows(axis 0) represent points in target, and columns(axis 1) represent parts. Then in $Swap()$, the minimum distance matrix $\textbf{Q}_{min}$ is computed by taking the minimum value of each row of $\textbf{Q}$. The region in the target $\textbf{T}$ that is least covered by any of the parts is extracted as $\textbf{T}^{-}$. Then for each part $\textbf{D}_i \in \mathcal{D}$, we compute the distance matrix $\textbf{Q}^{s}$ without the effect of $\textbf{D}_i$, and set the values of region $\textbf{T}^{-}$ in $\textbf{Q}^{s}$ to zero. The new minimum distance matrix after swap is computed as $\textbf{Q}^{s}_{min}$. The part $\textbf{D}_s$ that leads to the minimum value of $avg(\textbf{Q}^{s}_{min})$ will be replaced by $\textbf{T}^{-}$. Then in $NNSegment()$, the target point cloud $\textbf{T}$ is segmented into a set of segments where $\textbf{S}_i$ consists of all points from $\textbf{T}$ which are closer to $\textbf{D}_i$ than to any other decoded part. Then in $Filter()$, segment $S_i$ is updated with discarding a small amount of points that are already well covered. Finally in $FarthestConnectedComponent()$, the self connectivity matrix $\textbf{W}_{adj}$ is computed by using point to point connection when their distance is below a threshold. This connectivity matrix $\textbf{W}_{adj}$ is used to construct the graph $g$, followed by the connected components $\mathcal{C}$ are extracted from the graph $g$ and the connected component $\textbf{C}_i$ which is farthest from any of the other segments is selected. 

\textbf{Note: We need to correct one description made in the main paper section 3.2.2: The swap operation directly replaces one of the optimized parts $\textbf{D}_i$ with the region that is least covered by any of the parts (call this $\textbf{T}^{-}$ , not $\textbf{S}^{-}$) in the target $\textbf{T}$}.  

\begin{algorithm}[t!]
\begin{algorithmic}[1]
\Input
\Desc{Target shape $\textbf{T}$}
\Desc{Latent codes $\textbf{e}_i$, $i \in k$}
\Desc{Translation vectors $\textbf{t}_i$, $i \in k$}
\Desc{Rotation angles $r_i$, $i \in k$}
\Desc{Pre-trained part decoder $Decode$}
\EndInput
\Output
\Desc{Updated latent codes $\textbf{e}_i$, $i \in k$}
\Desc{Updated translation vectors $\textbf{t}_i$, $i \in k$}
\Desc{Updated rotation angles $r_i$, $i \in k$}
\EndOutput
\Procedure {Optimize}{}
\State $\textbf{D}_i \gets Decode(\textbf{e}_i)$
\State $\textbf{D}_i \gets Pose(\textbf{D}_i, \textbf{t}_i, r_i)$
\State $\mathcal{L} \gets \mathcal{L}_{recon} (\mathcal{D}, \textbf{T}) + \mathcal{L}_{overlap}(\mathcal{D})$
\State $\textbf{e}_i \gets \textbf{e}_i + \nabla{\mathcal{L}}(\textbf{e}_i)$
\State $\textbf{t}_i \gets \textbf{t}_i + \nabla{\mathcal{L}}(\textbf{t}_i)$ 
\State $r_i \gets r_i + \nabla{\mathcal{L}}(r_i)$
\State\Return $\textbf{e}_i, \textbf{t}_i, r_i$
\EndProcedure
\caption{
    Phase I Part Optimization 
    }
\label{alg:part_optimize}
\end{algorithmic}
\end{algorithm}

\begin{algorithm}[t!]
\begin{algorithmic}[1]
\Input
\Desc{Target shape $\textbf{T}$}
\Desc{Latent codes $\textbf{e}_i$, $i \in k$}
\Desc{Translation vectors $\textbf{t}_i$, $i \in k$}
\Desc{Rotation angles $r_i$, $i \in k$}
\Desc{Pre-trained part decoder $Decode$}
\Desc{Pre-trained part encoder $Encode$}
\EndInput
\Output
\Desc{Updated latent codes $\textbf{e}_i$, $i \in k$}
\Desc{Updated translation vectors $\textbf{t}_i$, $i \in k$}
\Desc{Updated rotation angles $r_i$, $i \in k$}
\EndOutput
\\
\Procedure {Shift}{}
\State $\textbf{D}_i \gets Decode(\textbf{e}_i)$
\State $\textbf{D}_i \gets Pose(\textbf{D}_i, \textbf{t}_i, r_i)$
\State $\textbf{Q} \gets cdist(\textbf{T}, \mathcal{D})$
\State $\textbf{D}_i \gets Swap(\textbf{D}_i, \mathcal{D}, \textbf{T}, \textbf{Q})$
\State $\textbf{S}_i \gets NNSegment(\textbf{D}_i, \textbf{T}, \textbf{Q}) $
\State $\textbf{S}_i \gets Filter(\textbf{S}_i,\textbf{T}, \textbf{Q})$
\State $\textbf{C}_i \gets  FarthestConnectedComponent(\textbf{S}_i, \mathcal{S}_{\ne i})$
\State $\textbf{t}_i \gets Center(\textbf{C}_i) $
\State $r_i \gets Rotation(\textbf{C}_i) $ 
\State $\textbf{C}_i \gets Pose(\textbf{C}_i, -\textbf{t}_i, -r_i)$
\State $\textbf{e}_i \gets Encode(\textbf{C}_i)$
\State\Return $\textbf{e}_i, \textbf{t}_i, r_i$ 
\EndProcedure
\\

\caption{
    Phase II Part Shift
    }
\label{alg:part_shift}
\end{algorithmic}
\end{algorithm}

\begin{algorithm}[t!]
\begin{algorithmic}[1]
\\

\Procedure {Swap}{$\textbf{D}_i, \mathcal{D}, \textbf{T}, \textbf{Q}$}
\State $\textbf{Q}_{min} \gets min(\textbf{Q}, axis=1).val$
\State $\textbf{T}^{-} \gets \textbf{T}[ind(topK(\textbf{Q}_{min}))]  $
\State $d_{min} \gets avg(\textbf{Q}_{min})$
\State $\textbf{D}_s \gets None$
\For {$\textbf{D}_i \in \mathcal{D}$}
\State $\textbf{Q}^{s} \gets (\textbf{Q}[:, 0:i-1] \cup \textbf{Q}[:, i+1:])$
\State $\textbf{Q}^{s}_{min} \gets min(\textbf{Q}^{s}, axis=1).val$
\State $\textbf{Q}^{s}_{min}[ind(\textbf{T}^{-})] \gets 0$
\State $d \gets avg(\textbf{Q}^{s}_{min})$
\If {$d < d_{min}$}
\State $d_{min} \gets d$
\State $\textbf{D}_s \gets \textbf{D}_i$ 
\EndIf
\EndFor
\State $\textbf{D}_s \gets \textbf{T}^{-}$ 
\State\Return $\textbf{D}_i$
\EndProcedure

\caption{
    Phase II helper Swap()
    }
\label{alg:swap}
\end{algorithmic}
\end{algorithm}

\begin{algorithm}[t!]
\begin{algorithmic}[1]
\\
\Procedure {NNSegment}{$\textbf{D}_i, \textbf{T}$, $\textbf{Q}$}
\State $\textbf{S}_i \gets \textbf{T}[min(\textbf{Q}, axis=1).ind == i] $
\State\Return $\textbf{S}_i$ 
\EndProcedure
\\

\caption{
    Phase II helper NNSegment()
    }
\label{alg:nnsegment}
\end{algorithmic}
\end{algorithm}

\begin{algorithm}[t!]
\begin{algorithmic}[1]
\\

\Procedure {Filter}{$\textbf{S}_i$, $\textbf{Q}$}
\State $\textbf{S}_i \gets \textbf{S}_i[ind(topK(\textbf{Q}[:, i]))] $
\State\Return $\textbf{S}_i$ 
\EndProcedure
\\

\caption{
    Phase II helper Filter()
    }
\label{alg:filter}
\end{algorithmic}
\end{algorithm}

\begin{algorithm}[t!]
\begin{algorithmic}[1]
\\

\Procedure {FarthestConnectedComponent}{$\textbf{S}_i$, $\mathcal{S}_{\ne i}$}
\State $\textbf{W} \gets cdist(\textbf{S}_i, \textbf{S}_i)$
\State $\textbf{W}_{adj} = \textbf{W}[\textbf{W} < \tau]$
\State $g \gets Graph(\textbf{W}_{adj}) $
\State $\mathcal{C} \gets g.connected\_components$
\State $d_{max} \gets 0$
\For {$\textbf{C} \in \mathcal{C}$}
\State $d \gets avg(cdist(Center(\textbf{C}), \mathcal{S}_{\ne i}))$
\If {$d > d_{max} $}
\State $d_{max} \gets d$
\State $\textbf{C}_i \gets \textbf{C}$
\EndIf
\EndFor
\State\Return $\textbf{C}_i$ 
\EndProcedure
\\

\caption{
    Phase II helper FarthestConnectedComponent()
    }
\label{alg:fcc}
\end{algorithmic}
\end{algorithm}

\begin{algorithm}[t!]
\begin{algorithmic}[1]
\Input
\Desc{Target reconstruction errors: $h_i$, $i \in N$}
\Desc{Target to Target distance matrix $\textbf{M}$}
\EndInput
\Output
\Desc{Updated latent codes $\textbf{e}_i$, $i \in k$}
\Desc{Updated translation vectors $\textbf{t}_i$, $i \in k$}
\Desc{Updated rotation angles $r_i$, $i \in k$}
\EndOutput
\Procedure {Borrow}{}
\State $nb \gets -1$
\For {$ j \in ind(botK(\textbf{M}[i, :]))$}
\If {$ h_i[j] \le \epsilon $}
\State $nb \gets j$
\State $break$
\EndIf
\EndFor
\If {$nb \ge 0$}
\State $\textbf{e}_i \gets \textbf{e}_{nb}$
\State $\textbf{t}_i \gets \textbf{t}_{nb}$
\State $r_i \gets r_{nb}$
\Else
\State $\textbf{e}_i \gets rand()$
\State $\textbf{t}_i \gets rand()$
\State $r_i \gets rand()$
\EndIf
\State\Return $\textbf{e}_i, \textbf{t}_i, r_i$
\EndProcedure
\caption{
    Phase III Part Borrowing 
    }
\label{alg:part_borrow}
\end{algorithmic}
\end{algorithm}

\paragraph{Dataset Preprocessing}
Our method takes volumetric point clouds as input.
In our experiments, we use the meshes provided in PartNet dataset to generate the corresponding point clouds. In the dataset preprocessing, all target shapes are centered to the origin. All parts in the part library are centered to the origin and rotated so the axes of their minimum volume bounding boxes align with the world axes. We then applied a hole filling method \cite{huang2018robust} to make all target shape meshes and part meshes watertight and applied rejection sampling to generate the volumetric point clouds for both target shapes and parts. For detecting symmetry planes for each shape, we iterate over several candidate planes that are perpendicular to xz plane that are centered at the object center. Then the points on one size of the plane are reflected to the other size, if the reflected points are overlap with the points on the other size, the symmetry plane is found. The overlap is determined by a threshold one for category with relative sparse point clouds such as Table and one for categories with relative dense point cloud such as Faucet and Lamp. One additional note for NP, the model that performs best on evaluation set during training is used to evaluate both train and test performance. 

\paragraph{Part VAE Architectures}
Please see Table~\ref{tab:part_encoder} and Table~\ref{tab:part_decoder} for architecture of the Part Encoder and Part Decoder. 

\begin{table}[t!]
    \centering
    \small
    \begin{tabular}{@{}c@{}}
        \toprule
        \textbf{Part Encoder}
        \\
        \midrule
        \textbf{Conv1d} $\left(3, 32, 1 \right)$\\
        \textbf{Batchnorm1d}\\
        \textbf{LeakyRelu}\\
        \textbf{Conv1d} $\left(32, 64, 1 \right)$\\
        \textbf{Batchnorm1d}\\
        \textbf{LeakyRelu}\\
        \textbf{Conv1d} $\left(64, 64, 1 \right)$\\
        \textbf{Batchnorm1d}\\
        \textbf{LeakyRelu}\\
        \textbf{Conv1d} $\left(64, 64, 1 \right)$\\
        \textbf{Batchnorm1d}\\
        \textbf{LeakyRelu}\\
        \textbf{MaxPool}\\
        \textbf{FC}$\left( 64 \times 64 \right)$\\
        \bottomrule
    \end{tabular}
    \vspace{0.1em}
    \caption{
    Detailed architecture of the Part Encoder 
    }
    \label{tab:part_encoder}
    \vspace{0.1mm}
\end{table}

\begin{table}[t!]
    \centering
    \small
    \begin{tabular}{@{}c@{}}
        \toprule
        \textbf{Part Decoder}
        \\
        \midrule
        \textbf{FC}$\left( 64 \times 512 \right)$\\
        \textbf{Batchnorm1d}\\
        \textbf{LeakyRelu}\\
        \textbf{FC}$\left( 512 \times 512 \right)$\\
        \textbf{Batchnorm1d}\\
        \textbf{LeakyRelu}\\
        \textbf{FC}$\left( 512 \times 1024 \right)$\\
        \textbf{Batchnorm1d}\\
        \textbf{LeakyRelu}\\
        \textbf{FC}$\left( 1024 \times 1024 \right)$\\
        \textbf{Batchnorm1d}\\
        \textbf{LeakyRelu}\\
        \textbf{FC}$\left( 1024 \times (512*3) \right)$\\
        \textbf{Reshape}\\
        \bottomrule
    \end{tabular}
    \vspace{0.1em}
    \caption{
    Detailed architecture of the Part Decoder 
    }
    \label{tab:part_decoder}
    \vspace{0.1mm}
\end{table}

\subsection{Additional Ablation Experiments}

We conducted several additional ablation studies on the Faucet category to investigate more about our method.

\begin{figure}[t!]
    \centering
    \includegraphics[width=0.8\linewidth]{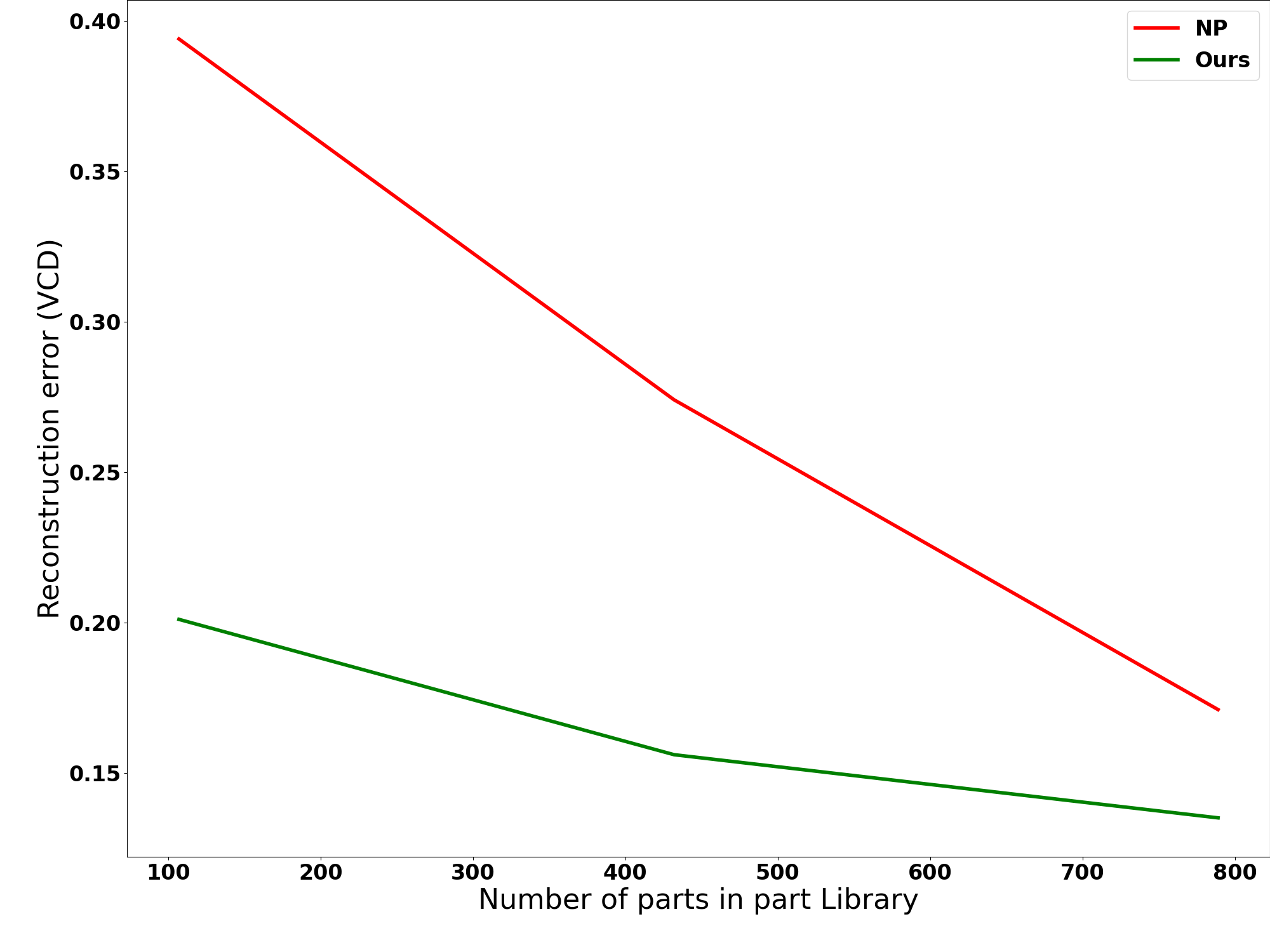}
    \caption{Part library size ablation (training target statistics). 
    }
    \label{fig:lib_size}
\end{figure}

\paragraph{Part library size} We investigate the influence of varying the size of the input part library. We cluster all parts according to the shape similarity, then we randomly choose certain number of clusters combined as the part library for our experiments. Please see Table~\ref{tab:lib_size} and Figure~\ref{fig:lib_size} for the results. As you can see, with fewer parts available in the part library, the performance of both our method and Neural Parts (NP) drops, but our method drops more gracefully (slower) than Neural Parts (NP) does. We think this shows our method can effectively take advantage of its awareness of the available parts.  

\begin{table}[t!]
    \centering
    \setlength{\tabcolsep}{1pt}
    \scriptsize
    \begin{tabular}{lccccc}
        \toprule
        \textbf{Method} & \textbf{Train (SCD) $\downarrow$} & \textbf{Train (VCD) $\downarrow$} & \textbf{Test (SCD) $\downarrow$} & \textbf{Test (VCD) $\downarrow$}
        \\
        \midrule
        NP (107 parts) & 0.668  & 0.394  & 0.623  & 0.353 
        \\
        Ours (107 parts) & \textbf{0.382}  & \textbf{0.201}  & \textbf{0.437}  & \textbf{0.225} 
        \\
        \midrule
        NP (432 parts) & 0.488  & 0.274  & 0.488  & 0.256
        \\
        Ours (432 parts) & \textbf{0.302}  & \textbf{0.156}  & \textbf{0.340}  & \textbf{0.163}
        \\
        \midrule
        NP (789 parts) & 0.326  & 0.171  & 0.370  & 0.174 
        \\
        Ours (789 parts) & \textbf{0.256}  & \textbf{0.135}  & \textbf{0.288}  & \textbf{0.134} 
        \\
        \bottomrule
    \end{tabular}
    \caption{Part library size ablation (Faucet). Note: numbers are multiplied by 100.
    }
    \label{tab:lib_size}
\end{table}

\begin{figure}[t!]
    \centering
    \includegraphics[width=0.8\linewidth]{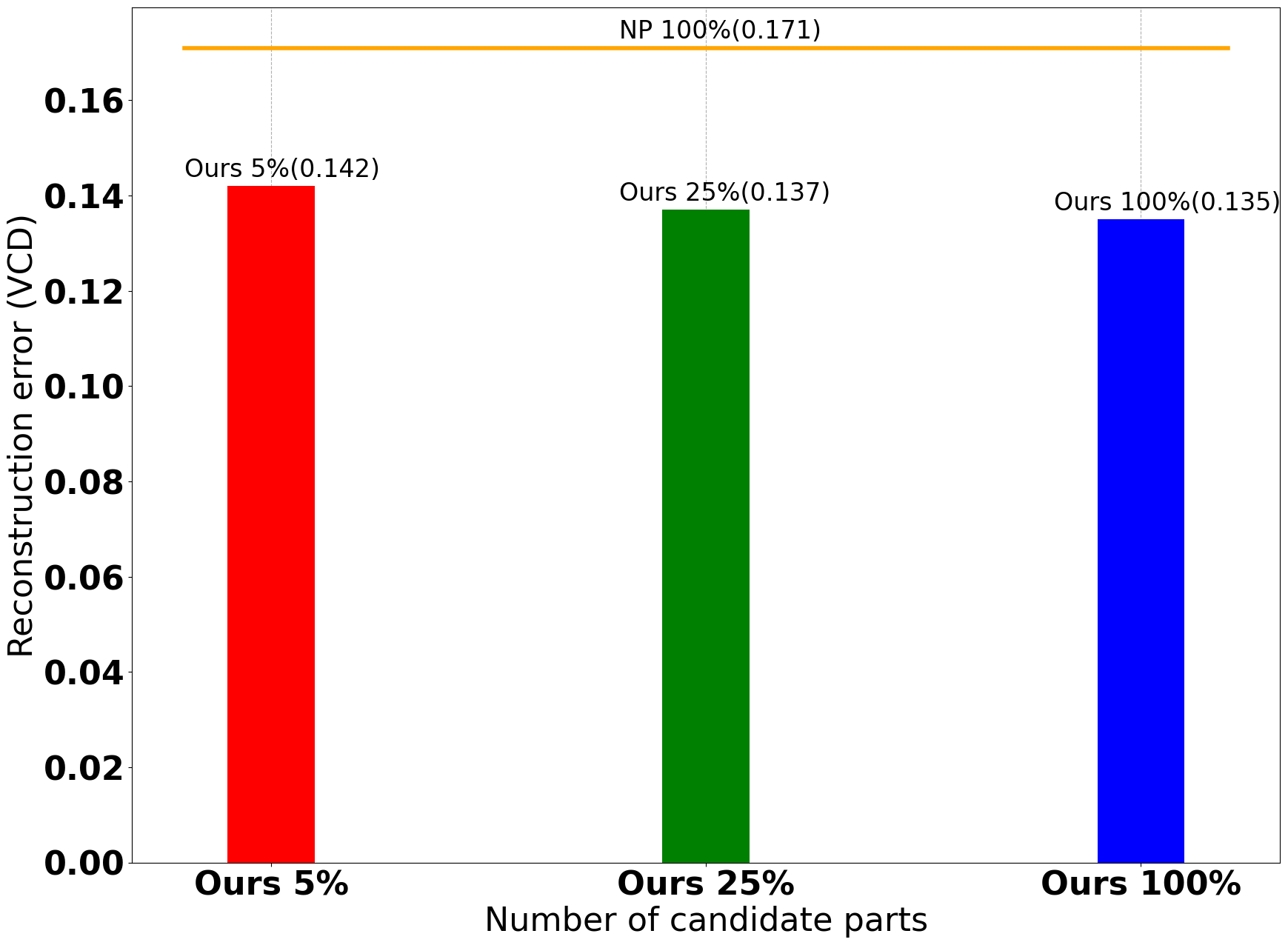}
    \caption{Retrieval part candidate number ablation (training target statistics).
    }
    \label{fig:retrieval_size}
\end{figure}

\paragraph{Retrieval candidate part number} We investigate the influence of varying size of the set of the parts considered as candidates for retrieval. In our main experiments, for all methods, we iterate over the entire part library to find the part that fits best to each segment in the target as the final part retrieval. This is because for Neural Parts (NP) and Brute Force (BF), there is no information to tell us which subsets of all parts to focus as the retrieval candidate parts. However, in our method, we have a part encoding latent space and for each optimized part we have an optimized part latent code. Our method can take advantage of this to just iterate and retrieve a subset of parts that are in the vicinity of the optimized part latent code in the latent space. We conducted the experiments with iterating only the nearest $5\%$, $25\%$ of the entire part library for the final retrieval. Please see Table~\ref{tab:retrieval_size} and Figure~\ref{fig:retrieval_size} for the results. As you can see, even with much smaller amount of part retrieval candidates, our method still outperforms NP.
This proves (1) the effectiveness of our latent space and (2) the potential speed-up advantage of our method when dealing with large input part libraries.

\begin{table}[t!]
    \centering
    \setlength{\tabcolsep}{2pt}
    \scriptsize
    \begin{tabular}{lccccc}
        \toprule
        \textbf{Method} & \textbf{Train (SCD) $\downarrow$} & \textbf{Train (VCD) $\downarrow$} & \textbf{Test (SCD) $\downarrow$} & \textbf{Test (VCD) $\downarrow$}
        \\
        \midrule
        NP (100\% of all parts) & 0.326  & 0.171  & 0.370  & 0.174 
        \\
        Ours (5\% of all parts) & 0.271  & 0.142  & 0.337  & 0.162 
        \\
        Ours (25\% of all parts) & 0.261  & 0.137  & 0.297  & 0.143 
        \\
        Ours (100\% of all parts) & 0.256  & 0.135  & 0.288  & 0.134 
        \\
        \bottomrule
    \end{tabular}
    \caption{Retrieval part candidate number ablation (Faucet). Note: numbers are multiplied by 100.
    }
    \label{tab:retrieval_size}
\end{table}

\begin{figure}[t!]
    \centering
    \includegraphics[width=0.8\linewidth]{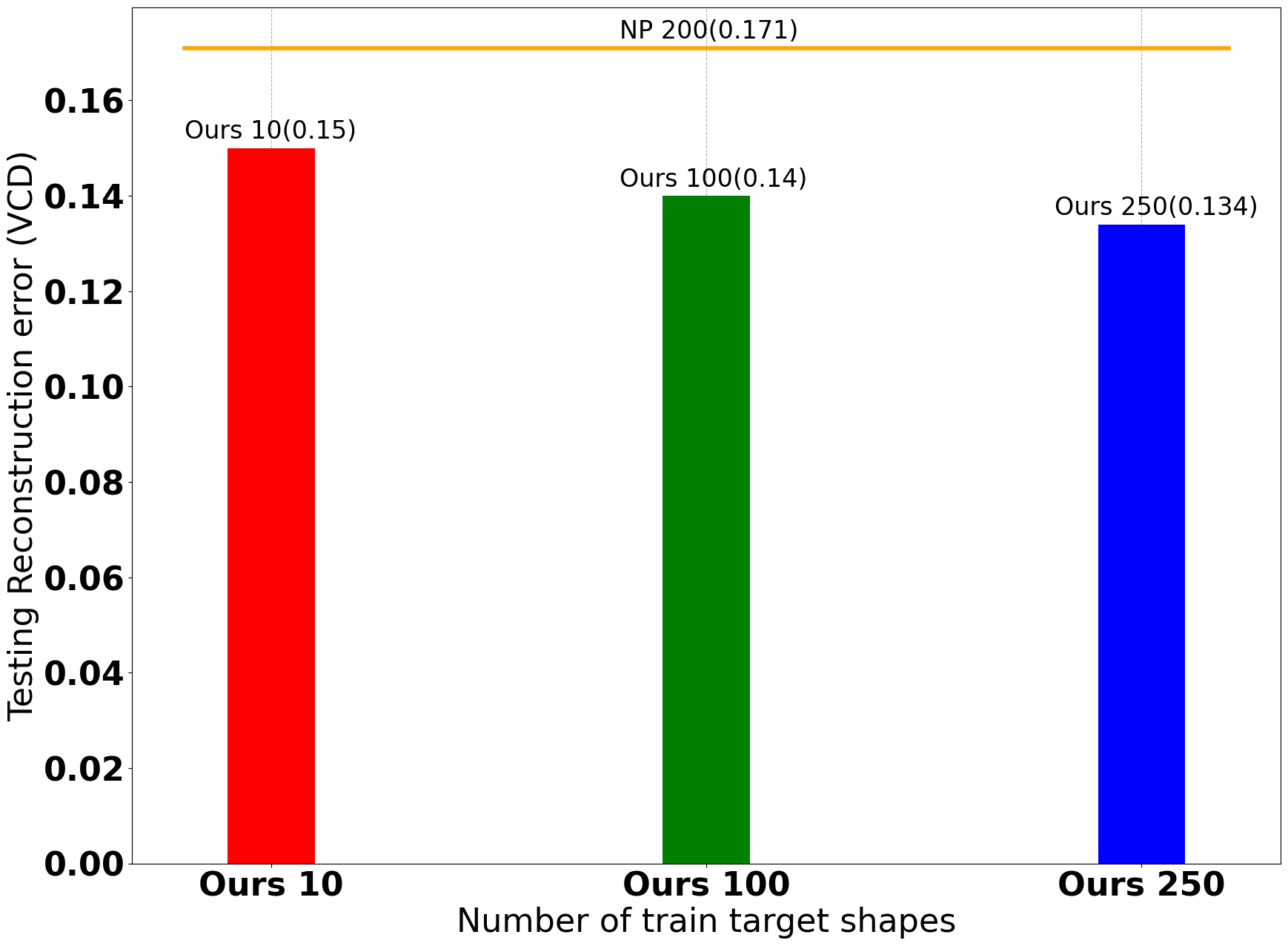}
    \caption{Training target number ablation.
    }
    \label{fig:target_size}
\end{figure}

\paragraph{Training target number} We investigate the influence of varying size of the set of input training target shapes on our amortized inference procedure. Please see Table~\ref{tab:target_size} and Figure~\ref{fig:target_size} for the results. As you can see, the performance of amortized inference does not drop much with fewer training target shapes.  

\begin{table}[t!]
    \centering
    \setlength{\tabcolsep}{2pt}
    \small
    \begin{tabular}{lccccc}
        \toprule
        \textbf{Method} & \textbf{Test (SCD) $\downarrow$} & \textbf{Test (VCD) $\downarrow$}
        \\
        \midrule
        NP (250 train shapes) & 0.370  & 0.174
        \\
        Ours (10 train shapes) & 0.297  & 0.150 
        \\
        Ours (100 train shapes) & 0.292  & 0.140 
        \\
        Ours (250 train shapes) & 0.288  & 0.134 
        \\
        \bottomrule
    \end{tabular}
    \caption{Training target number ablation (Faucet). Note: numbers are multiplied by 100.
    }
    \label{tab:target_size}
\end{table}

\subsection{Failure Cases}

\begin{figure}[t!]
    \centering
    \small
    \setlength{\tabcolsep}{1pt}
    \begin{tabular}{rccc}
        \raisebox{2.5em}{Targets} & 
        \includegraphics[width=0.3\linewidth]{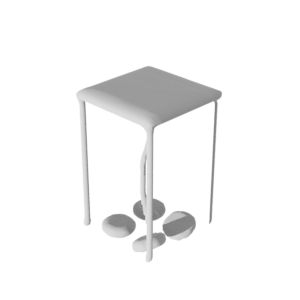} &
        \includegraphics[width=0.3\linewidth]{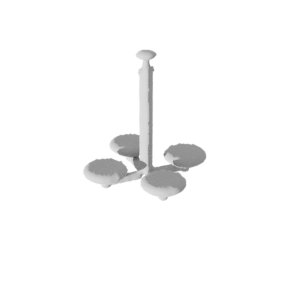} &
        \includegraphics[width=0.3\linewidth]{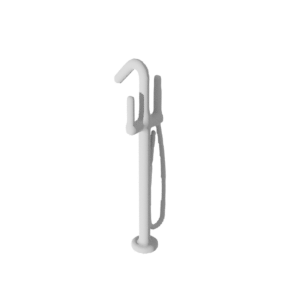} 
        \\
        \raisebox{2.5em}{Ours} & 
        \includegraphics[width=0.3\linewidth]{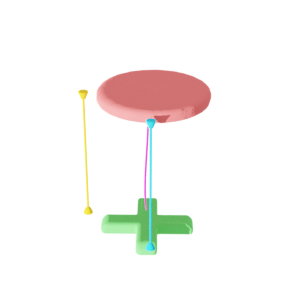} &
        \includegraphics[width=0.3\linewidth]{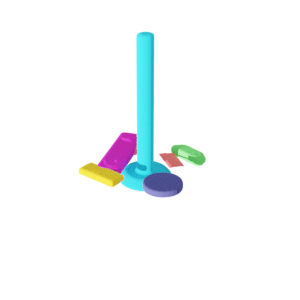} &
        \includegraphics[width=0.3\linewidth]{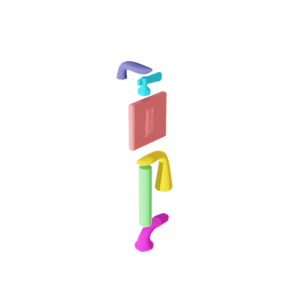} 
    \end{tabular}
    \caption{Failure Cases }
    \label{fig:failure}
\end{figure}

Figure~\ref{fig:failure} shows some failure patterns we identified for our method: (1) Insufficient points sampled on thin regions of the target shape; (2) Sub-optimal hyper-parameters, e.g. the shape is not judged as a symmetrical shape so that the symmetry constraint is not applied, and the connected component selection threshold in Phase II is too large so that disconnected geometry is the target is not successfully represented by different parts; (3) complex cluttered geometry details are missed due to the limitation of the point cloud representation (the small scale cluttered geometry details are represented by several points during the optimization).    

\subsection{Model-based Inference}

We also tried to use neural networks to perform fast inference, but we found that it did not perform as well as the amortized inference used in our method. We designed a RetrievalNet which takes in a target shape point cloud and outputs $k$ part latent codes $\textbf{e}_i$ in the latent space, together with an AssembleNet which takes in a target shape point cloud and a part latent code $\textbf{e}_i$ and outputs a translation vector $\textbf{t}_i$ and a rotation angle $r_i$ for part $i$. These networks are trained to directly regress the latent code, translation vector and rotation angle that we acquired from our direct optimization on the training target shapes. Then inference is performed on testing target shapes.

\begin{figure*}[t!]
    \centering
    \includegraphics[width=0.95\linewidth]{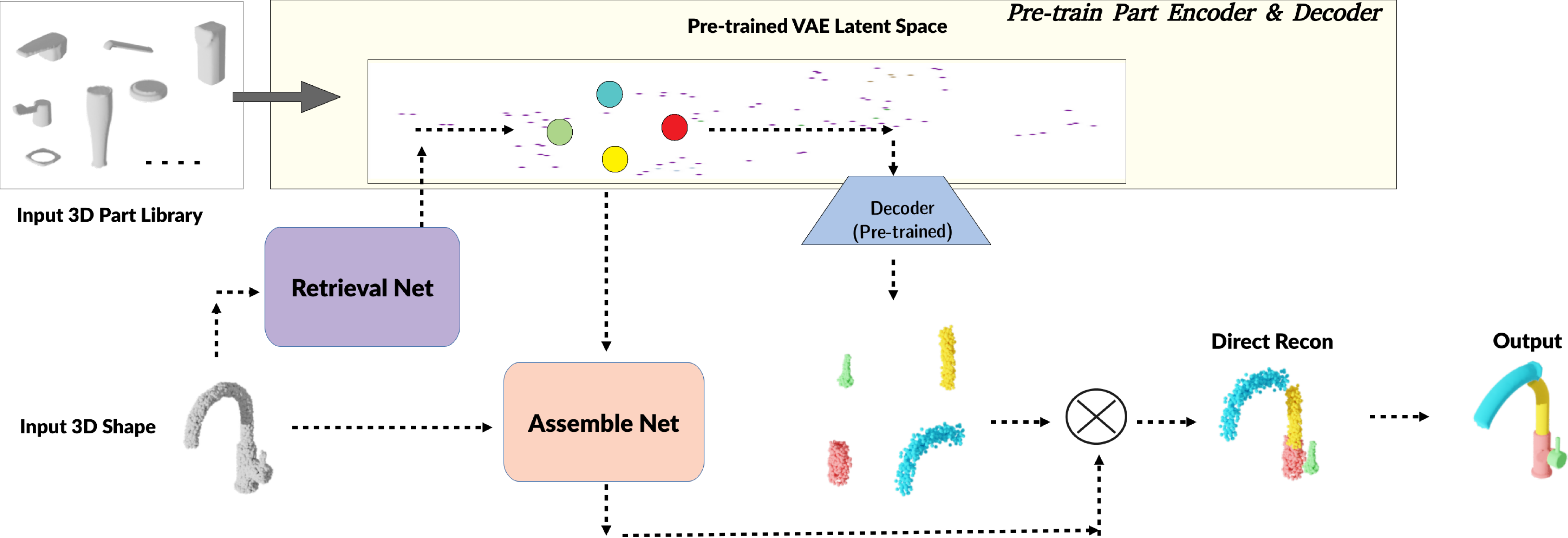}
    \caption{Model based Inference Overview: RetrievalNet takes in a target point cloud $\textbf{T}$ and output k part latent codes $\textbf{e}_i, i \in k$ in the latent space. AssembleNet takes in a target shape point cloud $\textbf{T}$ and a part latent code $\textbf{e}_i$ to output a translation vector $\textbf{t}_i$ and a rotation angle $r_i$ for part i. The decoded parts are translated and rotated according to $\textbf{t}_i$ and $r_i$. The same segmentation and retrieval operations as our method are performed to generate the final output.
    }
    
    \label{fig:model_infer_overview}
\end{figure*}

See Figure~\ref{fig:model_infer_overview} for an overview of this inference mode; see Table~\ref{tab:model_infer} for the reconstruction results on testing shapes from the Faucet category. The model based inference results are not as good as our amortized inference. We think the reason might be: each part output slot of the RetrievalNet plays a fixed role (for example, part $i$ is always made to reconstruct similar region on every different target shape.), so that the direct regression of high dimensional vectors which are independently optimized (can play different roles) can be difficult. Thus, it cannot properly handle large structural difference across target shapes. 

\begin{table}[t!]
    \centering
    \setlength{\tabcolsep}{2pt}
    \small
    \begin{tabular}{lccccc}
        \toprule
        \textbf{Method} & \textbf{Test (SCD) $\downarrow$} & \textbf{Test (VCD) $\downarrow$}
        \\
        \midrule
        NP & 0.370  & 0.174 
        \\
        Model Infer & 0.370  & 0.185 
        \\
        Ours & 0.288 & 0.134 
        \\
        \bottomrule
    \end{tabular}
    \caption{Comparison results NP vs. Model Infer vs. Ours (Faucet).  Note: numbers are multiplied by 100.
    }
    \label{tab:model_infer}
\end{table}

\subsection{Additional experiment results}

\begin{figure*}[t!]
    \centering
    \small
    \setlength{\tabcolsep}{1pt}
    \begin{tabular}{rcccccc}
        \raisebox{2.5em}{Targets} & 
        \includegraphics[width=0.15\linewidth]{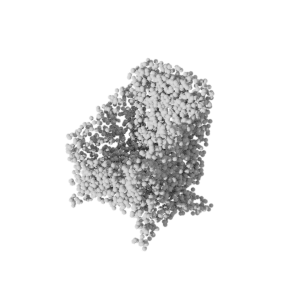} &
        \includegraphics[width=0.15\linewidth]{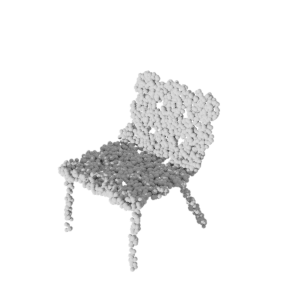} &
        \includegraphics[width=0.15\linewidth]{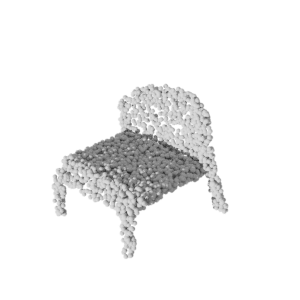} &
        \includegraphics[width=0.15\linewidth]{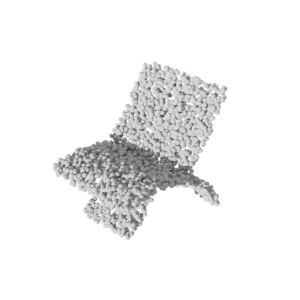} &
        \includegraphics[width=0.15\linewidth]{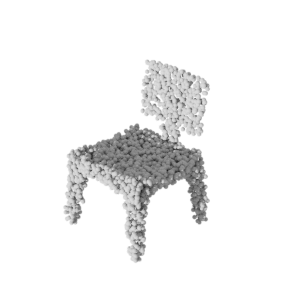} &
        \includegraphics[width=0.15\linewidth]{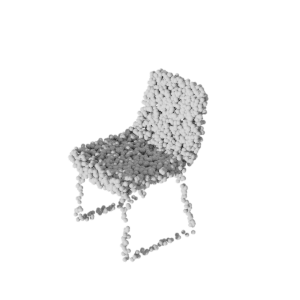} 
        \\
        \raisebox{2.5em}{JRD} & 
        \includegraphics[width=0.15\linewidth]{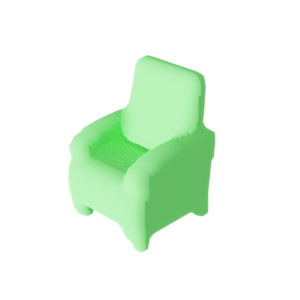} &
        \includegraphics[width=0.15\linewidth]{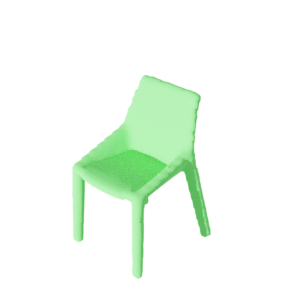} &
        \includegraphics[width=0.15\linewidth]{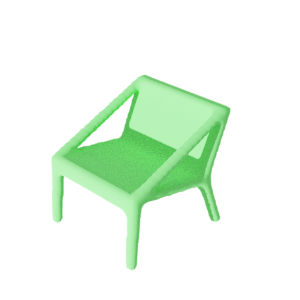} &
        \includegraphics[width=0.15\linewidth]{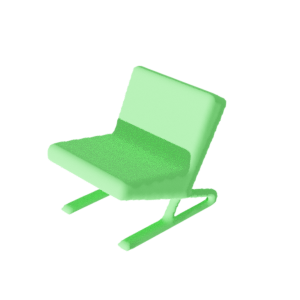} &
        \includegraphics[width=0.15\linewidth]{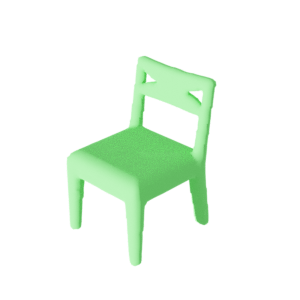} &
        \includegraphics[width=0.15\linewidth]{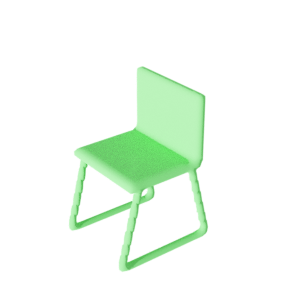} 
        \\
        \raisebox{2.5em}{Ours} & 
        \includegraphics[width=0.15\linewidth]{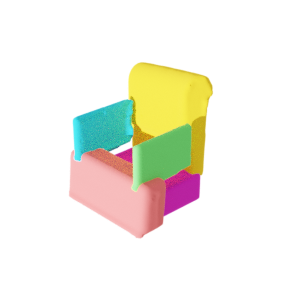} &
        \includegraphics[width=0.15\linewidth]{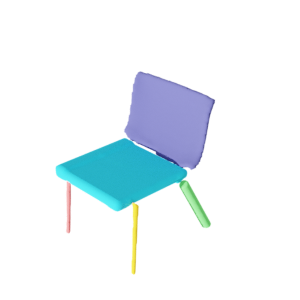} &
        \includegraphics[width=0.15\linewidth]{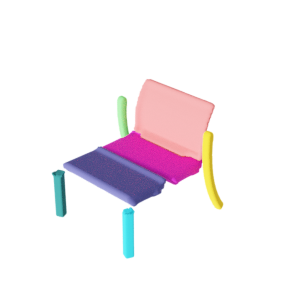} &
        \includegraphics[width=0.15\linewidth]{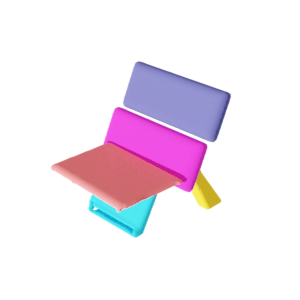} &
        \includegraphics[width=0.15\linewidth]{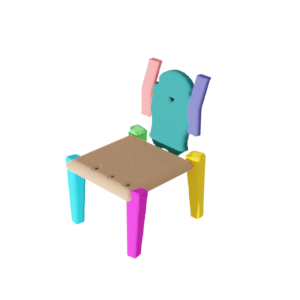} &
        \includegraphics[width=0.15\linewidth]{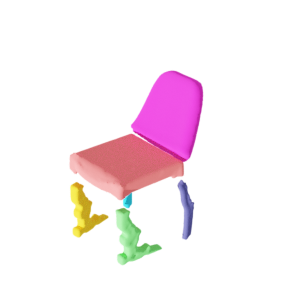} 
    \end{tabular}
    \caption{JRD vs. Ours on testing Chair shapes (1)  }
    \label{tab:jrd_vs_our_1}
\end{figure*}

\begin{figure*}[t!]
    \centering
    \small
    \setlength{\tabcolsep}{1pt}
    \begin{tabular}{rcccccc}
            \raisebox{2.5em}{Targets} & 
        \includegraphics[width=0.15\linewidth]{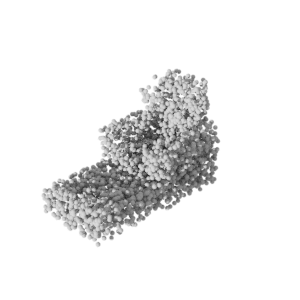} &
        \includegraphics[width=0.15\linewidth]{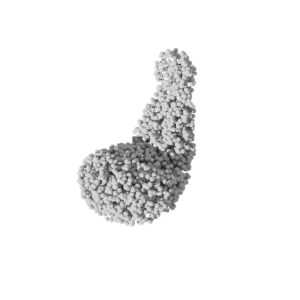} &
        \includegraphics[width=0.15\linewidth]{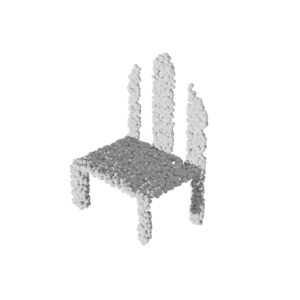} &
        \includegraphics[width=0.15\linewidth]{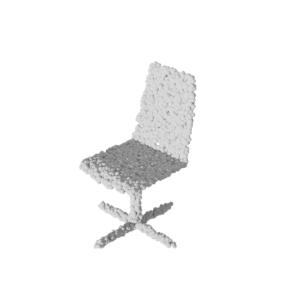} &
        \includegraphics[width=0.15\linewidth]{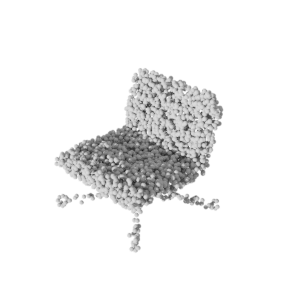} &
        \includegraphics[width=0.15\linewidth]{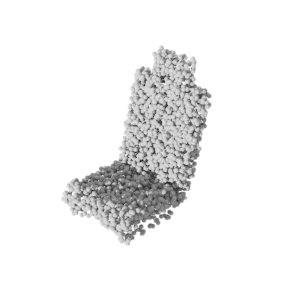} 
        \\
        \raisebox{2.5em}{JRD} & 
        \includegraphics[width=0.15\linewidth]{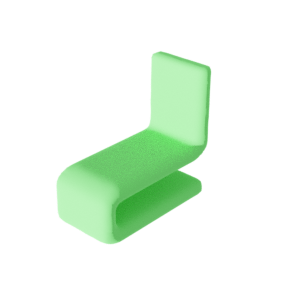} &
        \includegraphics[width=0.15\linewidth]{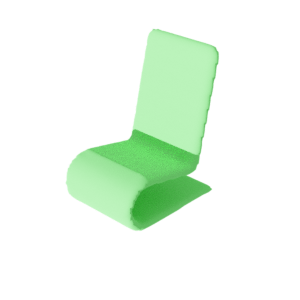} &
        \includegraphics[width=0.15\linewidth]{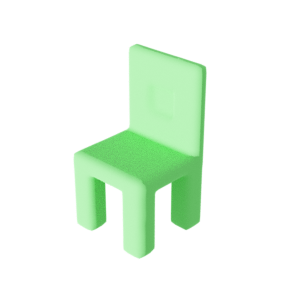} &
        \includegraphics[width=0.15\linewidth]{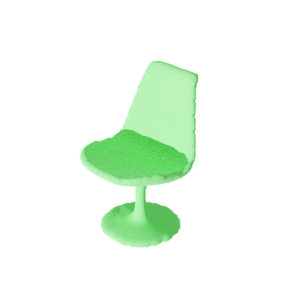} &
        \includegraphics[width=0.15\linewidth]{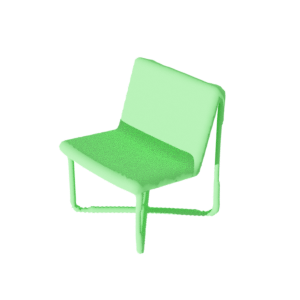} &
        \includegraphics[width=0.15\linewidth]{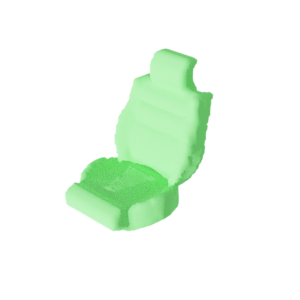} 
        \\
        \raisebox{2.5em}{Ours} & 
        \includegraphics[width=0.15\linewidth]{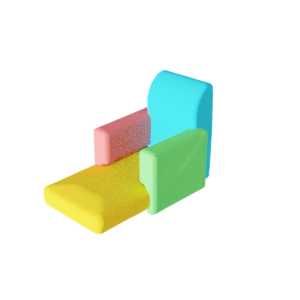} &
        \includegraphics[width=0.15\linewidth]{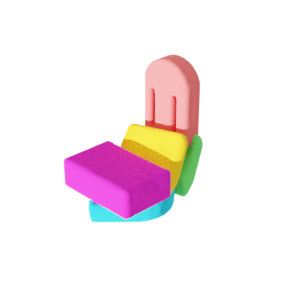} &
        \includegraphics[width=0.15\linewidth]{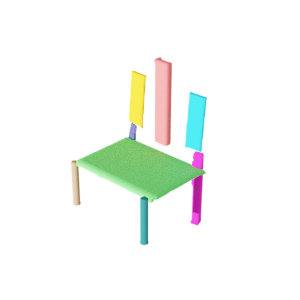} &
        \includegraphics[width=0.15\linewidth]{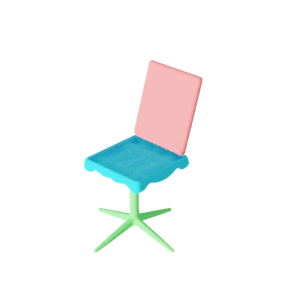} &
        \includegraphics[width=0.15\linewidth]{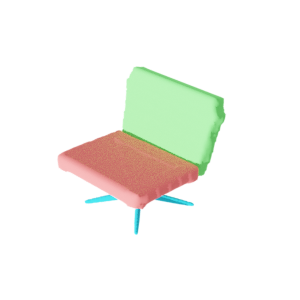} &
        \includegraphics[width=0.15\linewidth]{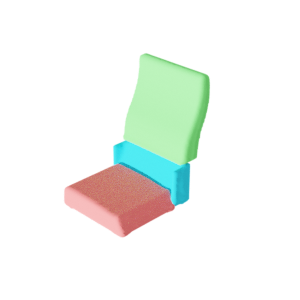} 
    \end{tabular}
    \caption{JRD vs. Ours on testing Chair shapes (2)  }
    \label{tab:jrd_vs_our_2}
\end{figure*}

\begin{figure*}[t!]
    \centering
    \small
    \setlength{\tabcolsep}{1pt}
    \begin{tabular}{rcccccccc}
        \raisebox{2.5em}{Targets} & 
        \includegraphics[width=0.15\linewidth]{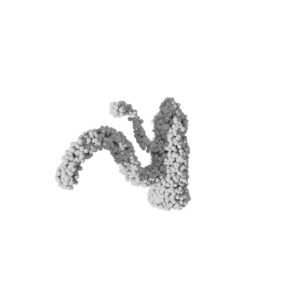} &
        \includegraphics[width=0.15\linewidth]{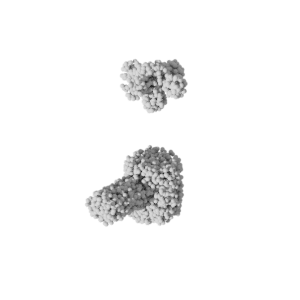} &
        \includegraphics[width=0.15\linewidth]{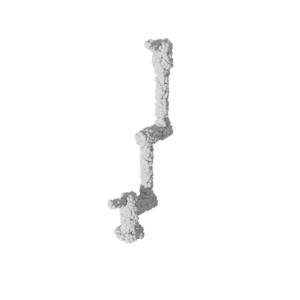} &
        \includegraphics[width=0.15\linewidth]{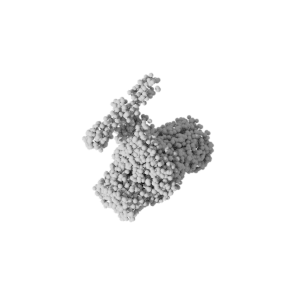} &
        \includegraphics[width=0.15\linewidth]{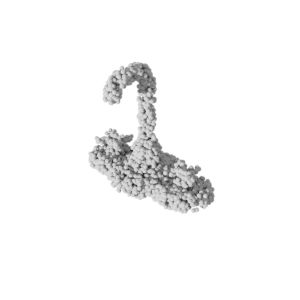} &
        \includegraphics[width=0.15\linewidth]{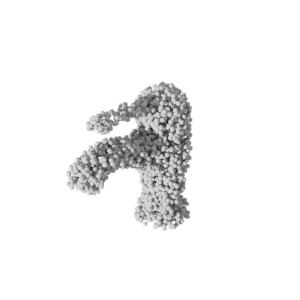} & 
        \\
        \raisebox{2.5em}{NP} & 
        \includegraphics[width=0.15\linewidth]{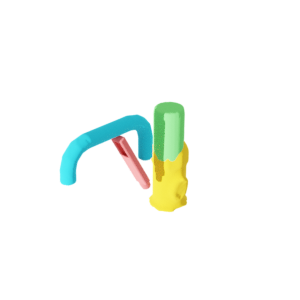} &
        \includegraphics[width=0.15\linewidth]{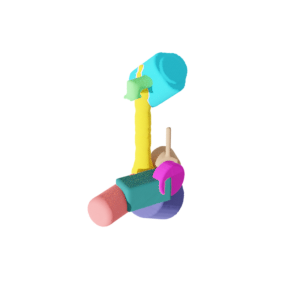} &
        \includegraphics[width=0.15\linewidth]{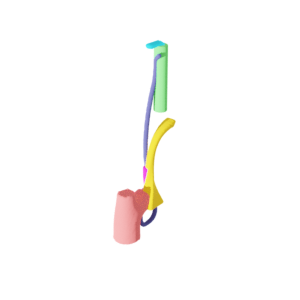} &
        \includegraphics[width=0.15\linewidth]{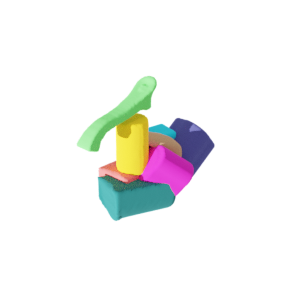} &
        \includegraphics[width=0.15\linewidth]{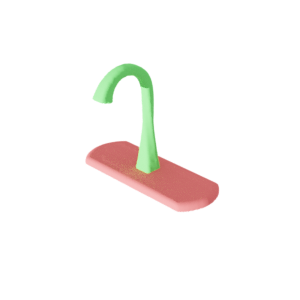} &
        \includegraphics[width=0.15\linewidth]{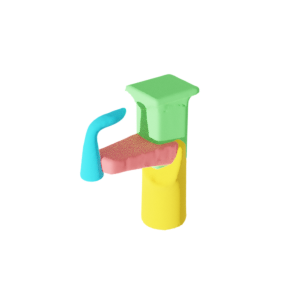} &
        \\
        \raisebox{2.5em}{Ours} & 
        \includegraphics[width=0.15\linewidth]{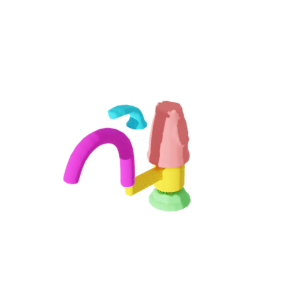} &
        \includegraphics[width=0.15\linewidth]{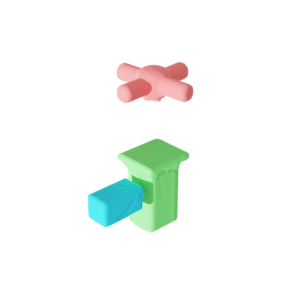} &
        \includegraphics[width=0.15\linewidth]{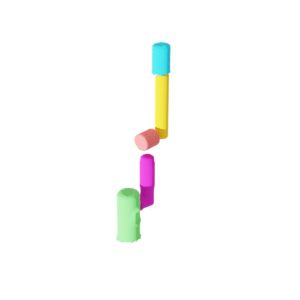} &
        \includegraphics[width=0.15\linewidth]{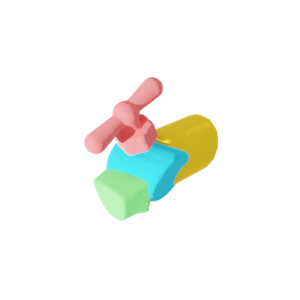} &
        \includegraphics[width=0.15\linewidth]{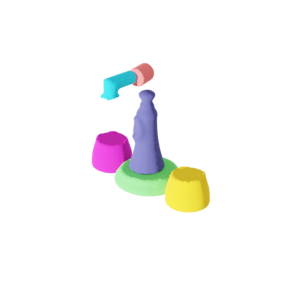} &
        \includegraphics[width=0.15\linewidth]{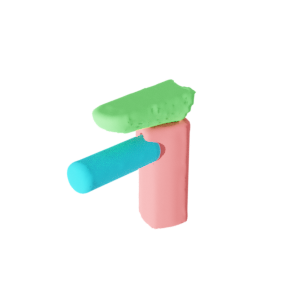}&

    \end{tabular}
    \caption{NP vs. Ours on Faucet category (1)}
    \label{tab:np_vs_our_faucet_1}
\end{figure*}

\begin{figure*}[t!]
    \centering
    \small
    \setlength{\tabcolsep}{1pt}
    \begin{tabular}{rcccccccc}
        \raisebox{2.5em}{Targets} & 
        \includegraphics[width=0.15\linewidth]{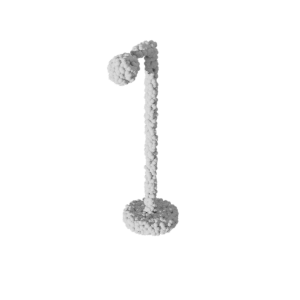} &
        \includegraphics[width=0.15\linewidth]{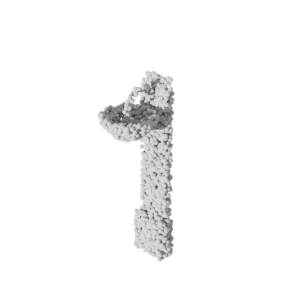} &
        \includegraphics[width=0.15\linewidth]{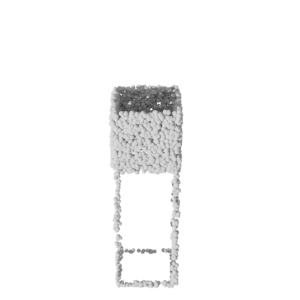} &
        \includegraphics[width=0.15\linewidth]{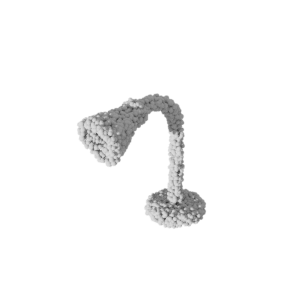} &
        \includegraphics[width=0.15\linewidth]{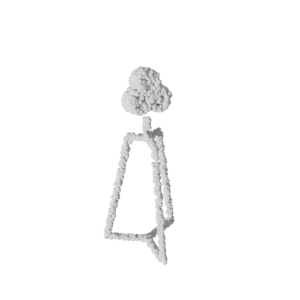} &
        \includegraphics[width=0.15\linewidth]{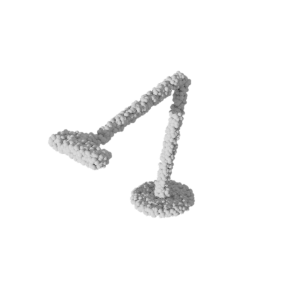} 
        \\
        \raisebox{2.5em}{NP} & 
        \includegraphics[width=0.15\linewidth]{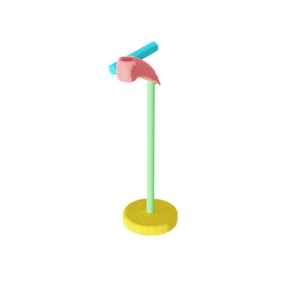} &
        \includegraphics[width=0.15\linewidth]{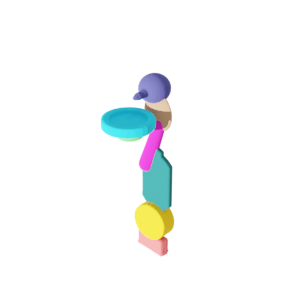} &
        \includegraphics[width=0.15\linewidth]{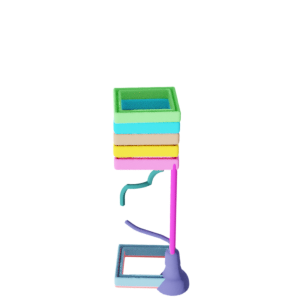} &
        \includegraphics[width=0.15\linewidth]{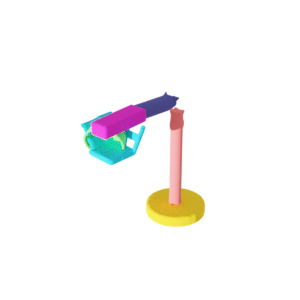} &
        \includegraphics[width=0.15\linewidth]{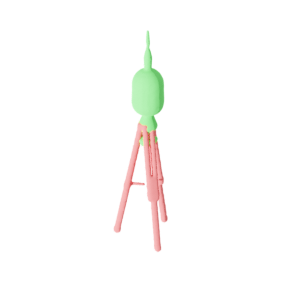} &
        \includegraphics[width=0.15\linewidth]{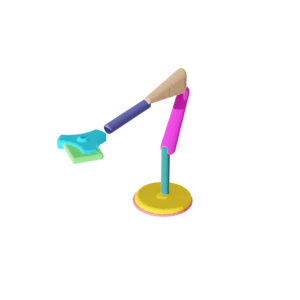} 
        \\
        \raisebox{2.5em}{Ours} & 
        \includegraphics[width=0.15\linewidth]{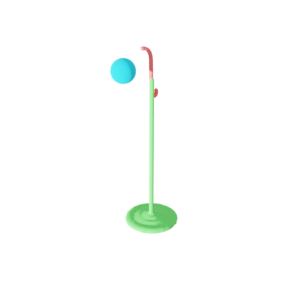} &
        \includegraphics[width=0.15\linewidth]{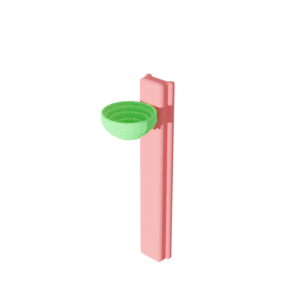} &
        \includegraphics[width=0.15\linewidth]{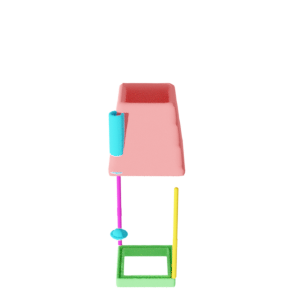} &
        \includegraphics[width=0.15\linewidth]{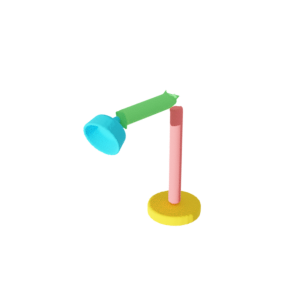} &
        \includegraphics[width=0.15\linewidth]{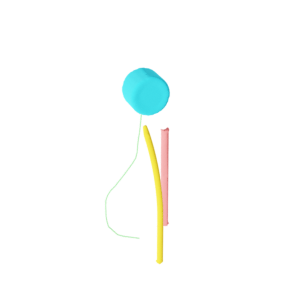} &
        \includegraphics[width=0.15\linewidth]{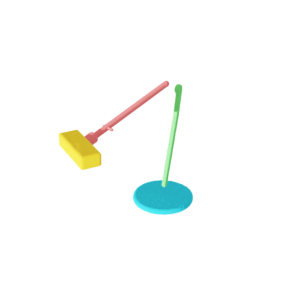} 
    \end{tabular}
    \caption{NP vs. Ours on Lamp category (5)}
    \label{tab:np_vs_our_lamp_5}
\end{figure*}

\begin{figure*}[t!]
    \centering
    \small
    \setlength{\tabcolsep}{1pt}
    \begin{tabular}{rcccccccc}
            \raisebox{2.5em}{Targets} & 
        \includegraphics[width=0.15\linewidth]{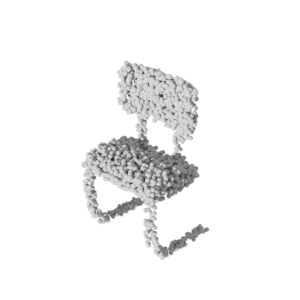} &
        \includegraphics[width=0.15\linewidth]{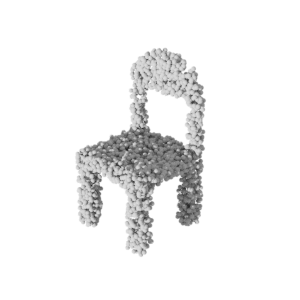} &
        \includegraphics[width=0.15\linewidth]{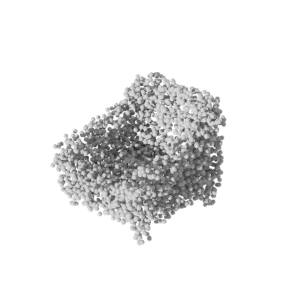} &
        \includegraphics[width=0.15\linewidth]{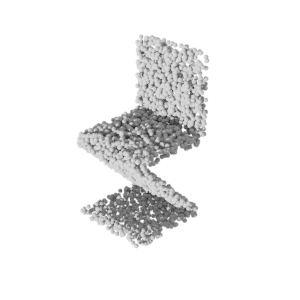} &
        \includegraphics[width=0.15\linewidth]{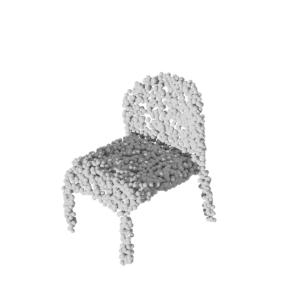} &
        \includegraphics[width=0.15\linewidth]{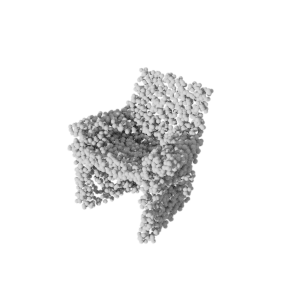}  
        \\
        \raisebox{2.5em}{Ours} & 
        \includegraphics[width=0.15\linewidth]{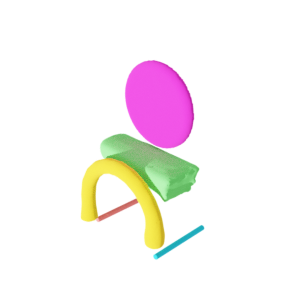} &
        \includegraphics[width=0.15\linewidth]{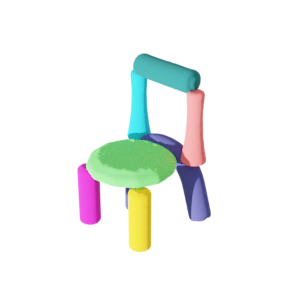} &
        \includegraphics[width=0.15\linewidth]{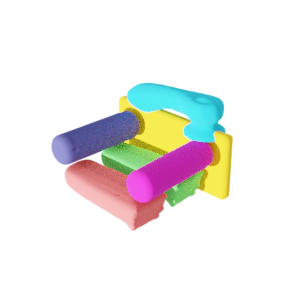} &
        \includegraphics[width=0.15\linewidth]{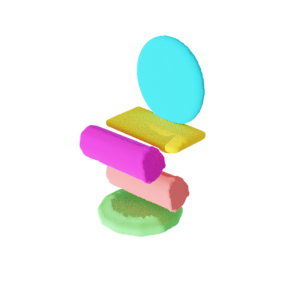} &
        \includegraphics[width=0.15\linewidth]{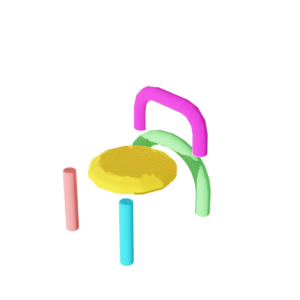} &
        \includegraphics[width=0.15\linewidth]{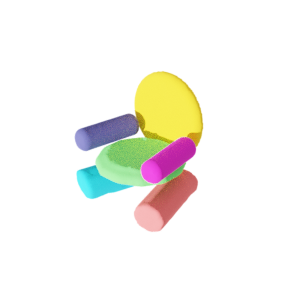}
    \end{tabular}
    \caption{Faucet Parts to Chairs}
    \label{tab:f2c}
\end{figure*}

\begin{figure*}[t!]
    \centering
    \small
    \setlength{\tabcolsep}{1pt}
    \begin{tabular}{rcccccccc}
            \raisebox{2.5em}{Targets} & 
        \includegraphics[width=0.15\linewidth]{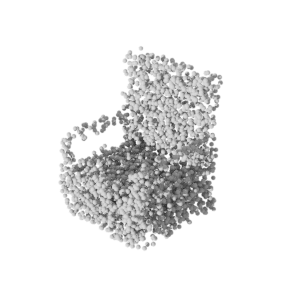} &
        \includegraphics[width=0.15\linewidth]{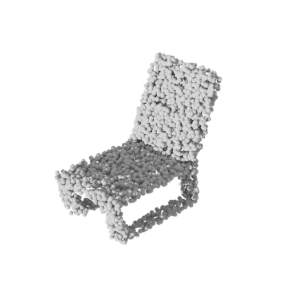} &
        \includegraphics[width=0.15\linewidth]{figs/results_supp/2862targetpc.png} &
        \includegraphics[width=0.15\linewidth]{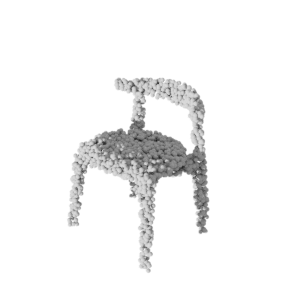} &
        \includegraphics[width=0.15\linewidth]{figs/results_supp/2955targetpc.png} &
        \includegraphics[width=0.15\linewidth]{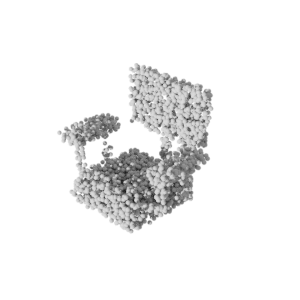}  
        \\
        \raisebox{2.5em}{Ours} & 
        \includegraphics[width=0.15\linewidth]{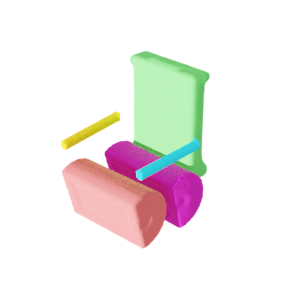} &
        \includegraphics[width=0.15\linewidth]{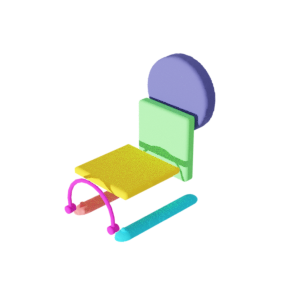} &
        \includegraphics[width=0.15\linewidth]{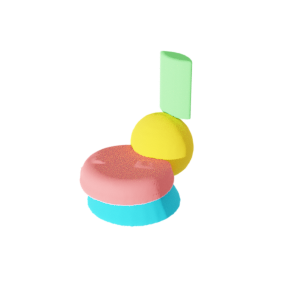} &
        \includegraphics[width=0.15\linewidth]{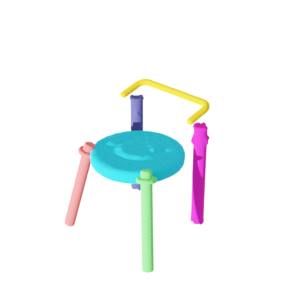} &
        \includegraphics[width=0.15\linewidth]{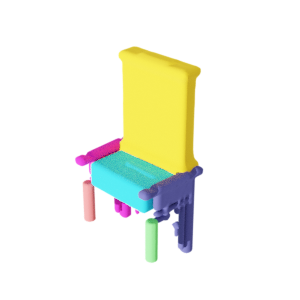} &
        \includegraphics[width=0.15\linewidth]{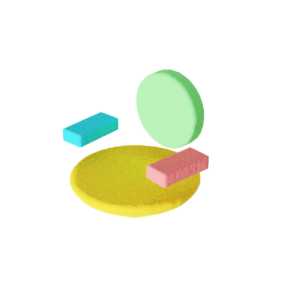}
    \end{tabular}
    \caption{Lamp Parts to Chairs}
    \label{tab:l2c}
\end{figure*}

Please see Figure~\ref{tab:jrd_vs_our_1} and Figures~\ref{tab:jrd_vs_our_2} for additional JRD vs. Ours qualitative comparison results. Please see Figure~\ref{tab:np_vs_our_faucet_1}--
% fig~\ref{tab:np_vs_our_faucet_2}, fig~\ref{tab:np_vs_our_faucet_3}, fig~\ref{tab:np_vs_our_faucet_4}, fig~\ref{tab:np_vs_our_faucet_5}, fig~\ref{tab:np_vs_our_chair_1}, fig~\ref{tab:np_vs_our_chair_2}, fig~\ref{tab:np_vs_our_chair_3}, fig~\ref{tab:np_vs_our_chair_4}, fig~\ref{tab:np_vs_our_chair_5}, fig~\ref{tab:np_vs_our_lamp_1}, fig~\ref{tab:np_vs_our_lamp_2}, fig~\ref{tab:np_vs_our_lamp_3}, fig~\ref{tab:np_vs_our_lamp_4} and
\ref{tab:np_vs_our_lamp_5} for additional NP vs. Ours qualitative comparison results.  Please see Figures~\ref{tab:f2c} and \ref{tab:l2c} for additional qualitative cross-category reconstruction results.

\begin{figure*}[t!]
    \centering
    \small
    \setlength{\tabcolsep}{1pt}
    \begin{tabular}{rcccccccc}
        \raisebox{2.5em}{Targets} & 
        \includegraphics[width=0.15\linewidth]{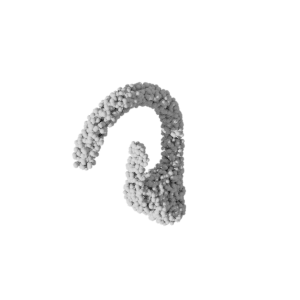} &
        \includegraphics[width=0.15\linewidth]{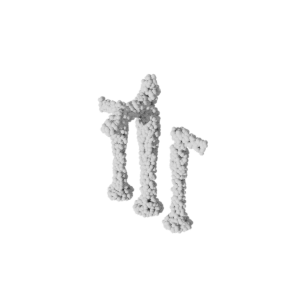} &
        \includegraphics[width=0.15\linewidth]{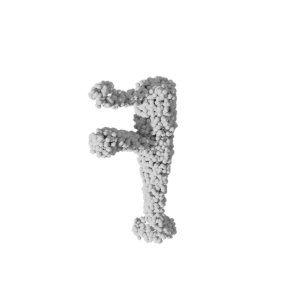} &
        \includegraphics[width=0.15\linewidth]{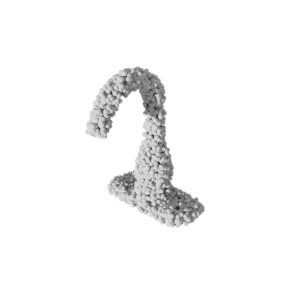} &
        \includegraphics[width=0.15\linewidth]{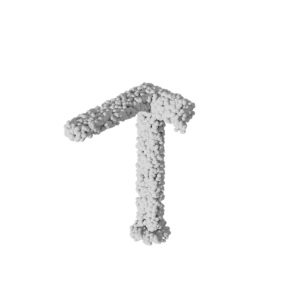} &
        \includegraphics[width=0.15\linewidth]{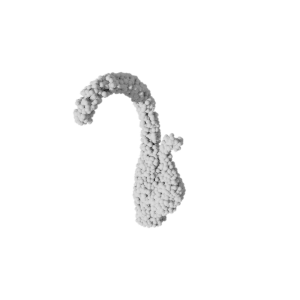} & 
        \\
        \raisebox{2.5em}{NP} & 
        \includegraphics[width=0.15\linewidth]{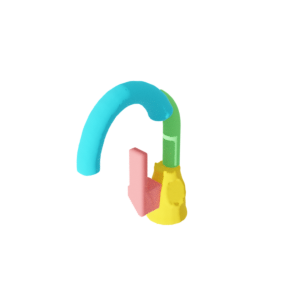} &
        \includegraphics[width=0.15\linewidth]{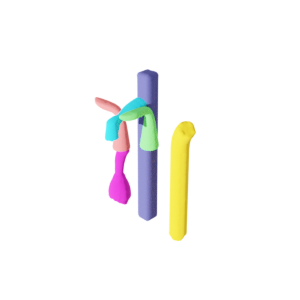} &
        \includegraphics[width=0.15\linewidth]{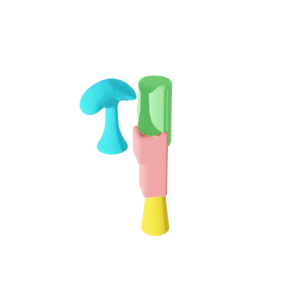} &
        \includegraphics[width=0.15\linewidth]{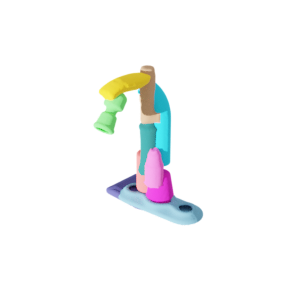} &
        \includegraphics[width=0.15\linewidth]{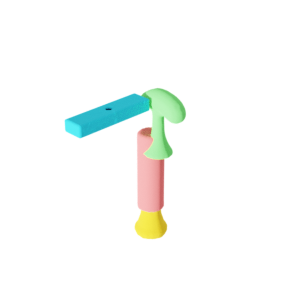} &
        \includegraphics[width=0.15\linewidth]{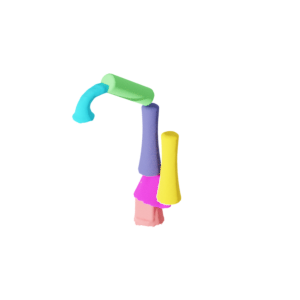} &
        \\
        \raisebox{2.5em}{Ours} & 
        \includegraphics[width=0.15\linewidth]{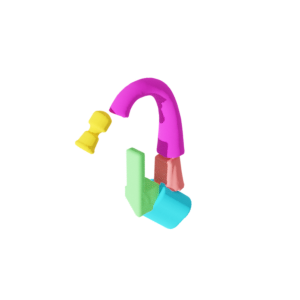} &
        \includegraphics[width=0.15\linewidth]{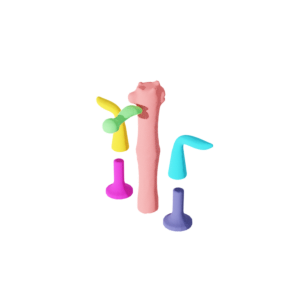} &
        \includegraphics[width=0.15\linewidth]{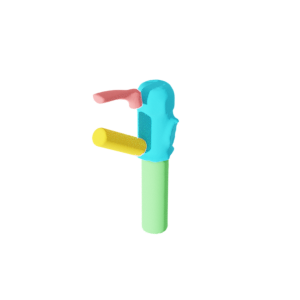} &
        \includegraphics[width=0.15\linewidth]{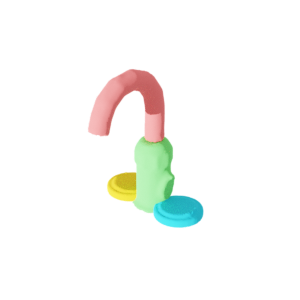} &
        \includegraphics[width=0.15\linewidth]{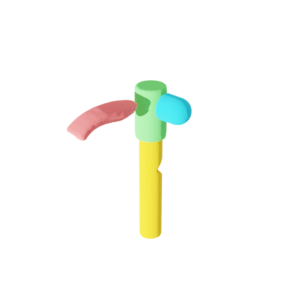} &
        \includegraphics[width=0.15\linewidth]{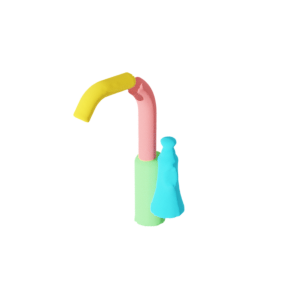}&

    \end{tabular}
    \caption{NP vs. Ours on Faucet category (2)}
    \label{tab:np_vs_our_faucet_2}
\end{figure*}

\begin{figure*}[t!]
    \centering
    \small
    \setlength{\tabcolsep}{1pt}
    \begin{tabular}{rcccccccc}
            \raisebox{2.5em}{Targets} & 
        \includegraphics[width=0.15\linewidth]{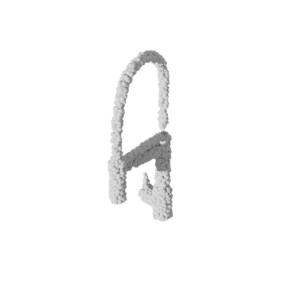} &
        \includegraphics[width=0.15\linewidth]{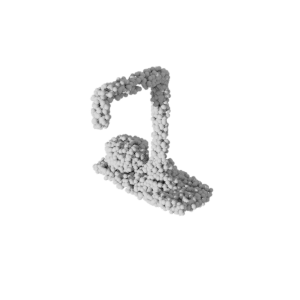} &
        \includegraphics[width=0.15\linewidth]{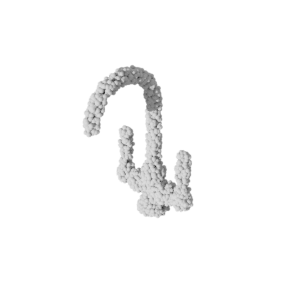} &
        \includegraphics[width=0.15\linewidth]{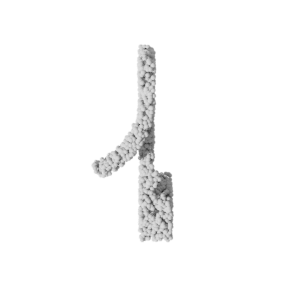} &
        \includegraphics[width=0.15\linewidth]{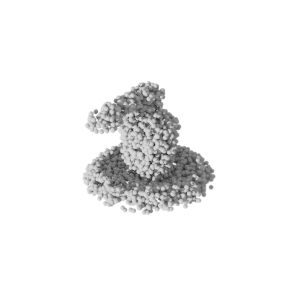} &
        \includegraphics[width=0.15\linewidth]{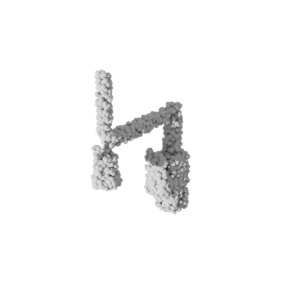} & 

        \\
        \raisebox{2.5em}{NP} & 
        \includegraphics[width=0.15\linewidth]{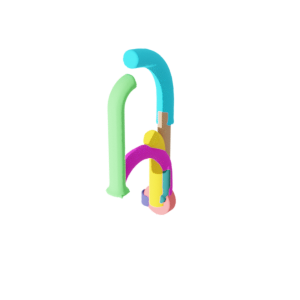} &
        \includegraphics[width=0.15\linewidth]{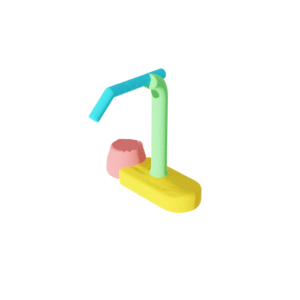} &
        \includegraphics[width=0.15\linewidth]{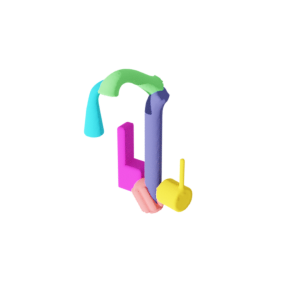} &
        \includegraphics[width=0.15\linewidth]{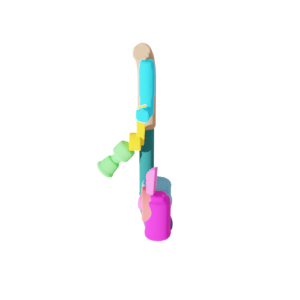} &
        \includegraphics[width=0.15\linewidth]{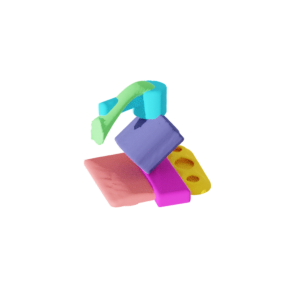} &
        \includegraphics[width=0.15\linewidth]{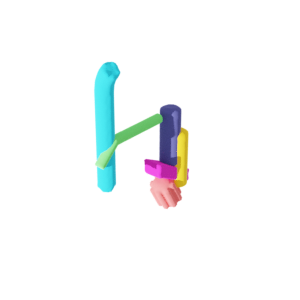} &

        \\
        \raisebox{2.5em}{Ours} & 
        \includegraphics[width=0.15\linewidth]{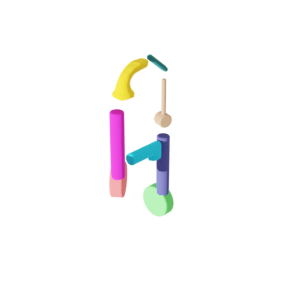} &
        \includegraphics[width=0.15\linewidth]{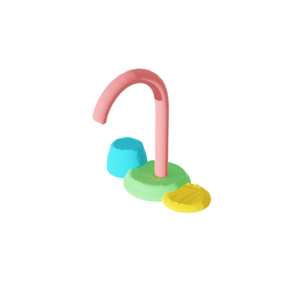} &
        \includegraphics[width=0.15\linewidth]{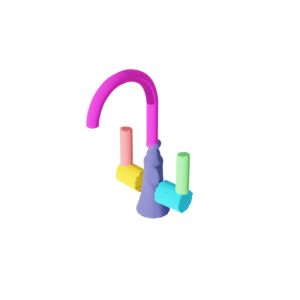} &
        \includegraphics[width=0.15\linewidth]{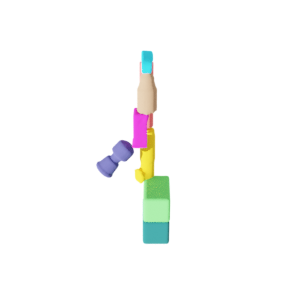} &
        \includegraphics[width=0.15\linewidth]{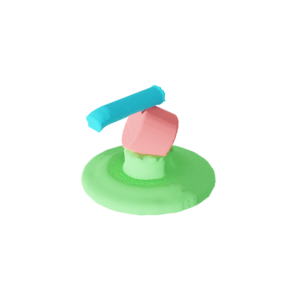} &
        \includegraphics[width=0.15\linewidth]{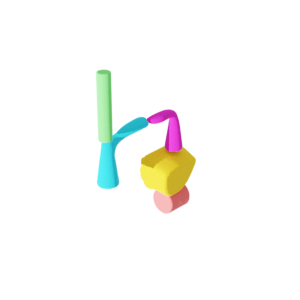}&

    \end{tabular}
    \caption{NP vs. Ours on Faucet category (3)}
    \label{tab:np_vs_our_faucet_3}
\end{figure*}

\begin{figure*}[t!]
    \centering
    \small
    \setlength{\tabcolsep}{1pt}
    \begin{tabular}{rcccccccc}
            \raisebox{2.5em}{Targets} & 
        \includegraphics[width=0.15\linewidth]{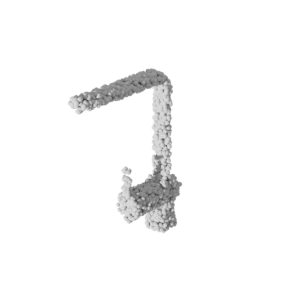} &
        \includegraphics[width=0.15\linewidth]{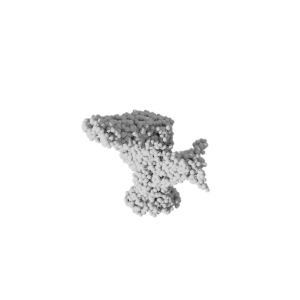} &
        \includegraphics[width=0.15\linewidth]{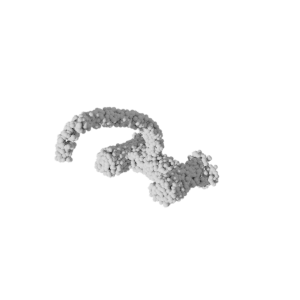} &
        \includegraphics[width=0.15\linewidth]{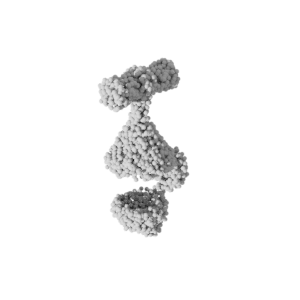} &
        \includegraphics[width=0.15\linewidth]{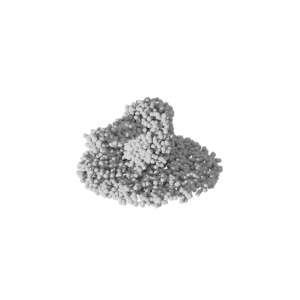} &
        \includegraphics[width=0.15\linewidth]{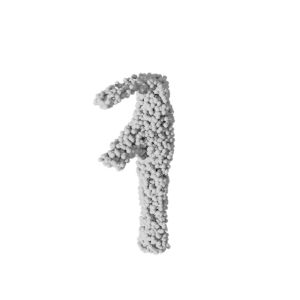} & 
        \\
        \raisebox{2.5em}{NP} & 
        \includegraphics[width=0.15\linewidth]{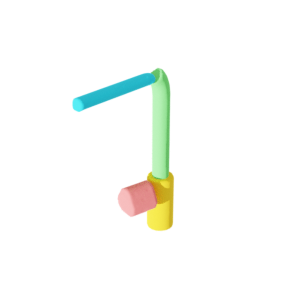} &
        \includegraphics[width=0.15\linewidth]{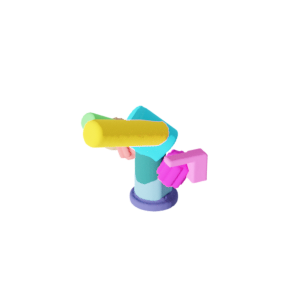} &
        \includegraphics[width=0.15\linewidth]{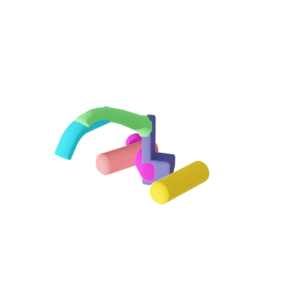} &
        \includegraphics[width=0.15\linewidth]{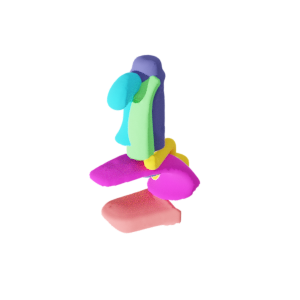} &
        \includegraphics[width=0.15\linewidth]{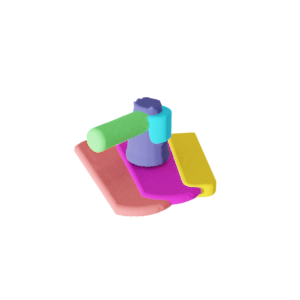} &
        \includegraphics[width=0.15\linewidth]{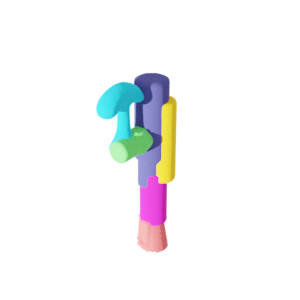} &

        \\
        \raisebox{2.5em}{Ours} & 
        \includegraphics[width=0.15\linewidth]{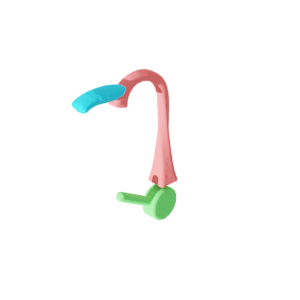} &
        \includegraphics[width=0.15\linewidth]{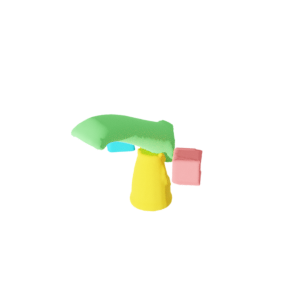} &
        \includegraphics[width=0.15\linewidth]{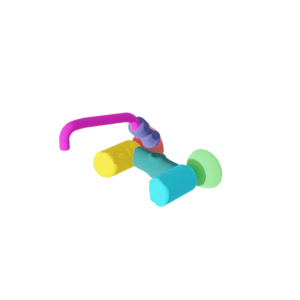} &
        \includegraphics[width=0.15\linewidth]{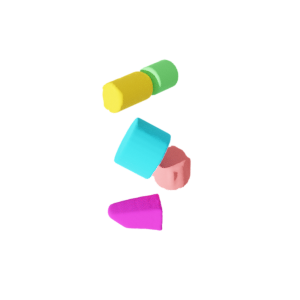} &
        \includegraphics[width=0.15\linewidth]{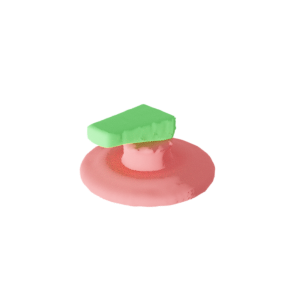} &
        \includegraphics[width=0.15\linewidth]{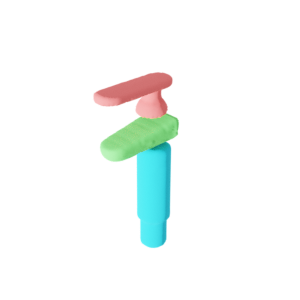}&
    \end{tabular}
    \caption{NP vs. Ours on Faucet category (4)}
    \label{tab:np_vs_our_faucet_4}
\end{figure*}

\begin{figure*}[t!]
    \centering
    \small
    \setlength{\tabcolsep}{1pt}
    \begin{tabular}{rcccccccc}
            \raisebox{2.5em}{Targets} & 
        \includegraphics[width=0.15\linewidth]{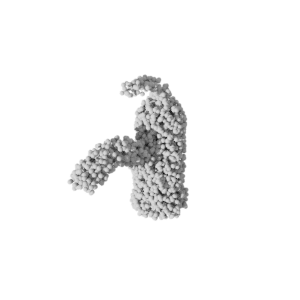} &
        \includegraphics[width=0.15\linewidth]{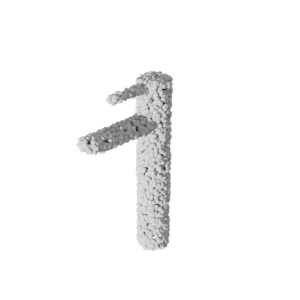} &
        \includegraphics[width=0.15\linewidth]{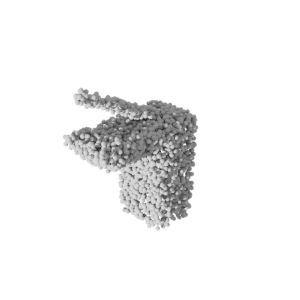} &
        \includegraphics[width=0.15\linewidth]{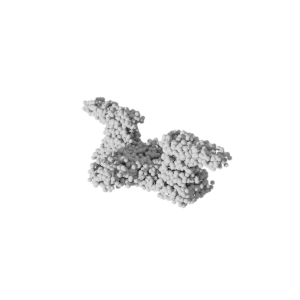} &
        \includegraphics[width=0.15\linewidth]{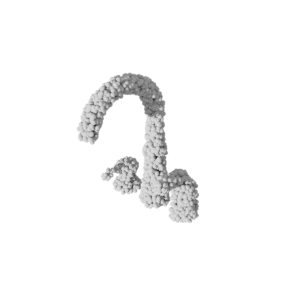} &
        \includegraphics[width=0.15\linewidth]{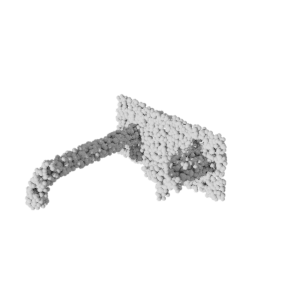} 
        \\
        \raisebox{2.5em}{NP} & 
        \includegraphics[width=0.15\linewidth]{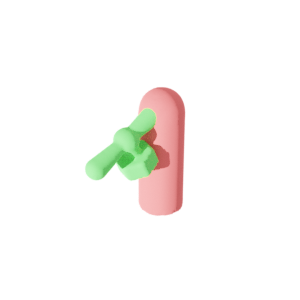} &
        \includegraphics[width=0.15\linewidth]{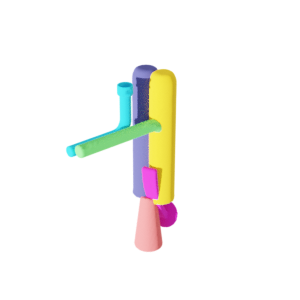} &
        \includegraphics[width=0.15\linewidth]{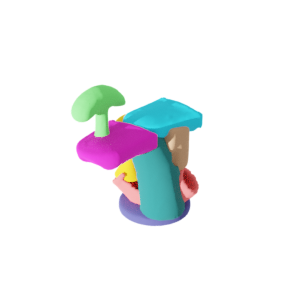} &
        \includegraphics[width=0.15\linewidth]{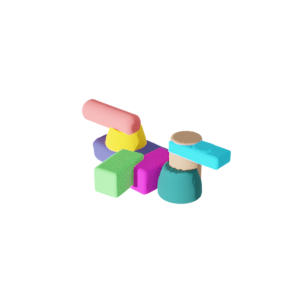} &
        \includegraphics[width=0.15\linewidth]{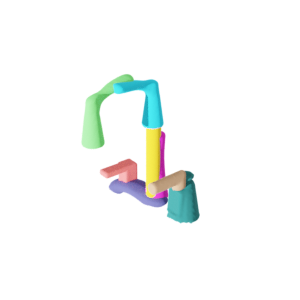} &
        \includegraphics[width=0.15\linewidth]{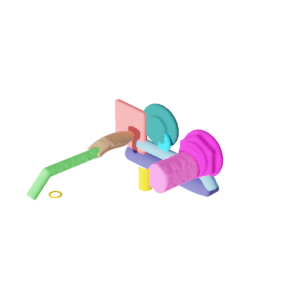} 
        \\
        \raisebox{2.5em}{Ours} & 
        \includegraphics[width=0.15\linewidth]{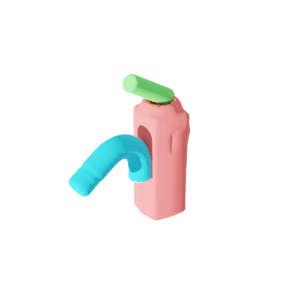} &
        \includegraphics[width=0.15\linewidth]{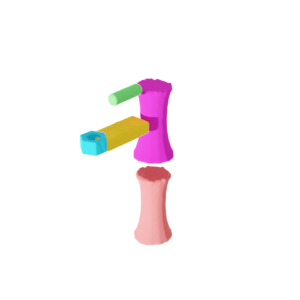} & 
        \includegraphics[width=0.15\linewidth]{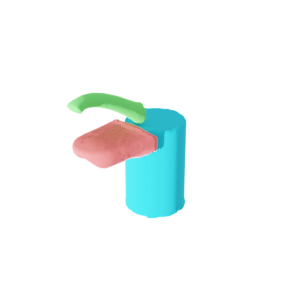} &
        \includegraphics[width=0.15\linewidth]{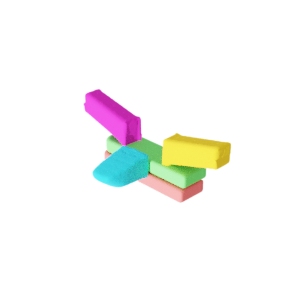} &
        \includegraphics[width=0.15\linewidth]{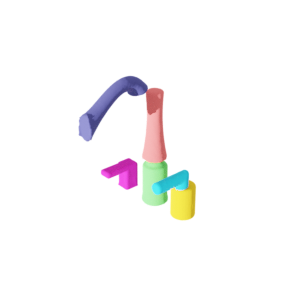} &
        \includegraphics[width=0.15\linewidth]{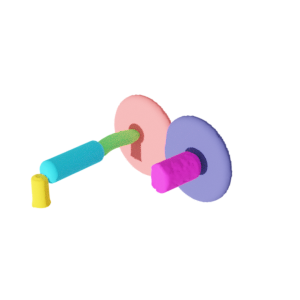}  
    \end{tabular}
    \caption{NP vs. Ours on Faucet category (5)}
    \label{tab:np_vs_our_faucet_5}
\end{figure*}

\begin{figure*}[t!]
    \centering
    \small
    \setlength{\tabcolsep}{1pt}
    \begin{tabular}{rcccccccc}
            \raisebox{2.5em}{Targets} & 
        \includegraphics[width=0.15\linewidth]{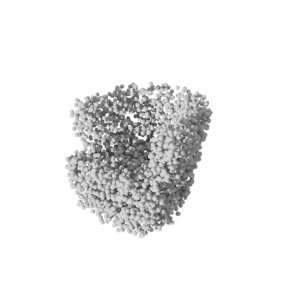} &
        \includegraphics[width=0.15\linewidth]{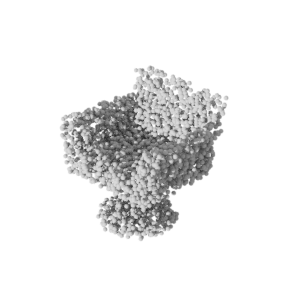} &
        \includegraphics[width=0.15\linewidth]{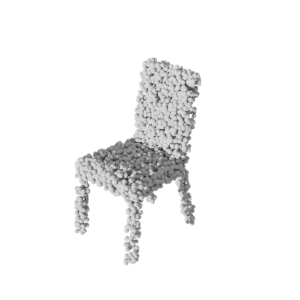} &
        \includegraphics[width=0.15\linewidth]{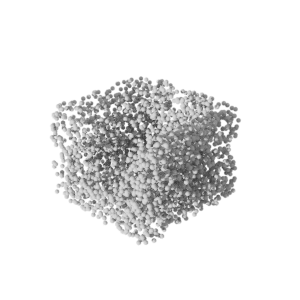} &
        \includegraphics[width=0.15\linewidth]{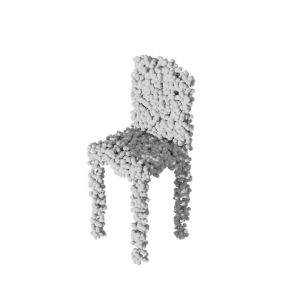} &
        \includegraphics[width=0.15\linewidth]{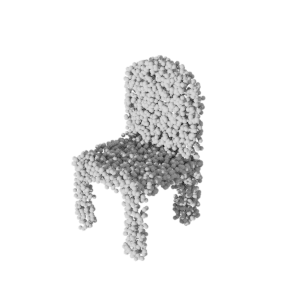} 
        \\
        \raisebox{2.5em}{NP} & 
        \includegraphics[width=0.15\linewidth]{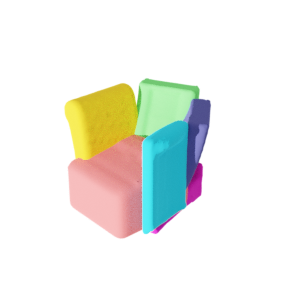} &
        \includegraphics[width=0.15\linewidth]{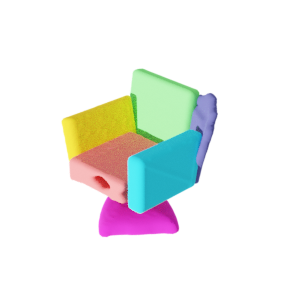} &
        \includegraphics[width=0.15\linewidth]{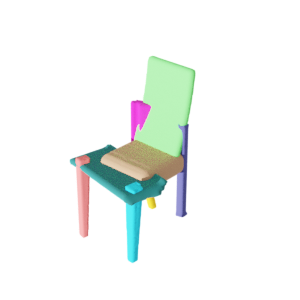} &
        \includegraphics[width=0.15\linewidth]{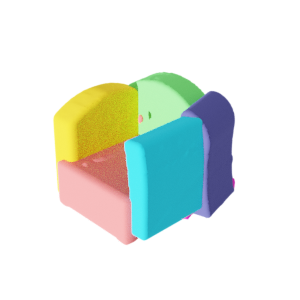} &
        \includegraphics[width=0.15\linewidth]{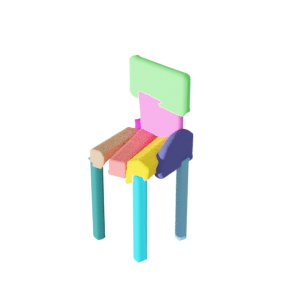} &
        \includegraphics[width=0.15\linewidth]{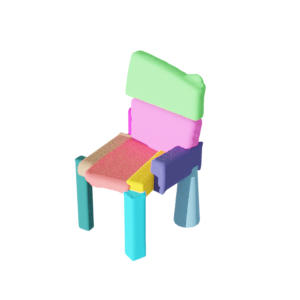} 
        \\
        \raisebox{2.5em}{Ours} & 
        \includegraphics[width=0.15\linewidth]{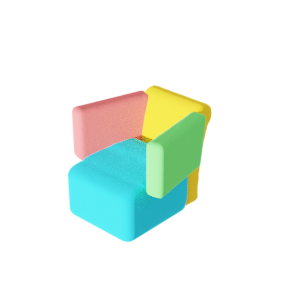} &
        \includegraphics[width=0.15\linewidth]{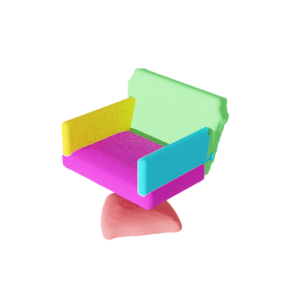} &
        \includegraphics[width=0.15\linewidth]{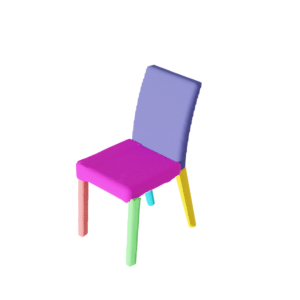} &
        \includegraphics[width=0.15\linewidth]{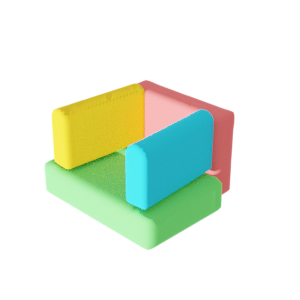} &
        \includegraphics[width=0.15\linewidth]{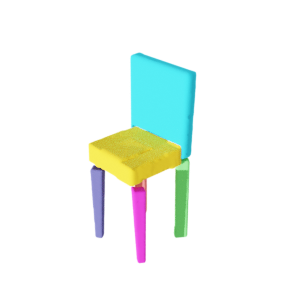} &
        \includegraphics[width=0.15\linewidth]{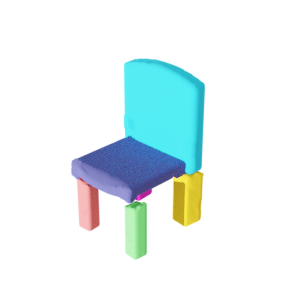}

    \end{tabular}
    \caption{NP vs. Ours on Chair category (1)}
    \label{tab:np_vs_our_chair_1}
\end{figure*}

\begin{figure*}[t!]
    \centering
    \small
    \setlength{\tabcolsep}{1pt}
    \begin{tabular}{rcccccccc}
            \raisebox{2.5em}{Targets} & 
        \includegraphics[width=0.15\linewidth]{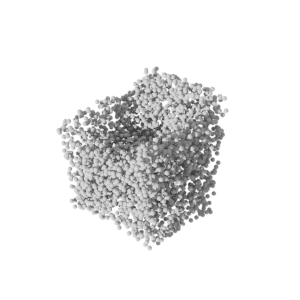} &
        \includegraphics[width=0.15\linewidth]{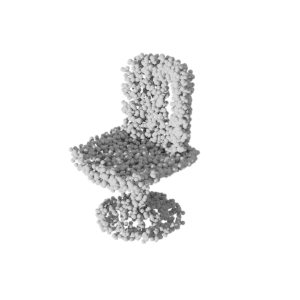} &
        \includegraphics[width=0.15\linewidth]{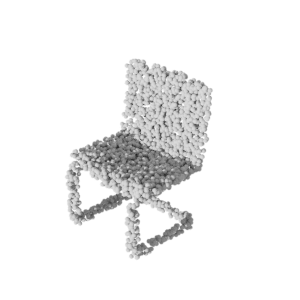} &
        \includegraphics[width=0.15\linewidth]{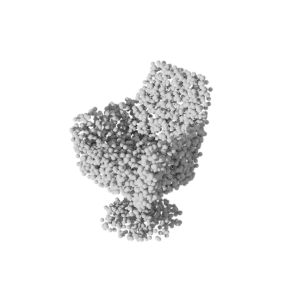} &
        \includegraphics[width=0.15\linewidth]{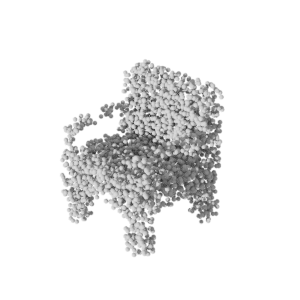} &
        \includegraphics[width=0.15\linewidth]{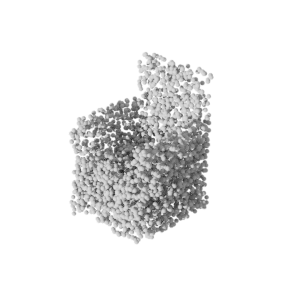} 
        \\
        \raisebox{2.5em}{NP} & 
        \includegraphics[width=0.15\linewidth]{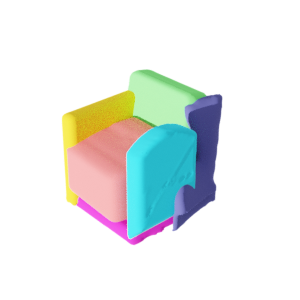} &
        \includegraphics[width=0.15\linewidth]{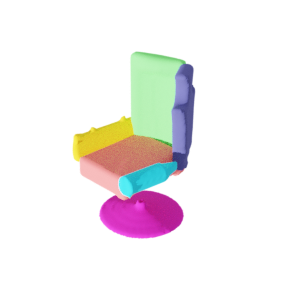} &
        \includegraphics[width=0.15\linewidth]{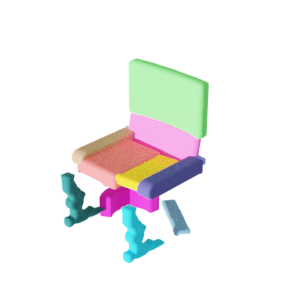} &
        \includegraphics[width=0.15\linewidth]{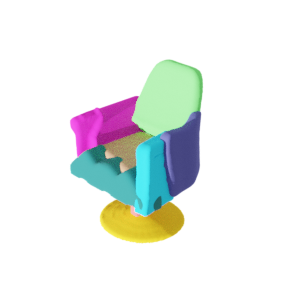} &
        \includegraphics[width=0.15\linewidth]{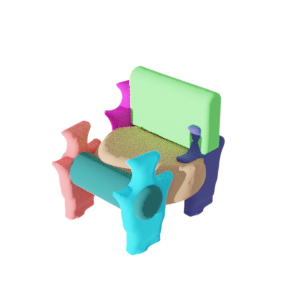} &
        \includegraphics[width=0.15\linewidth]{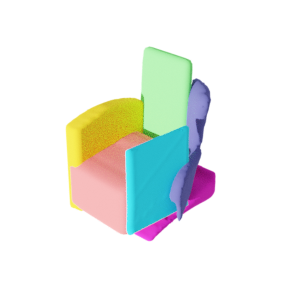} 
        \\
        \raisebox{2.5em}{Ours} & 
        \includegraphics[width=0.15\linewidth]{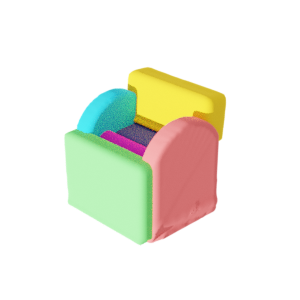} &
        \includegraphics[width=0.15\linewidth]{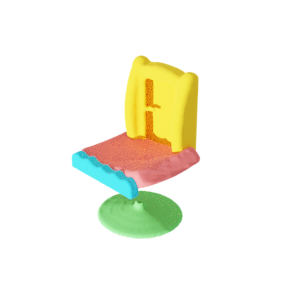} &
        \includegraphics[width=0.15\linewidth]{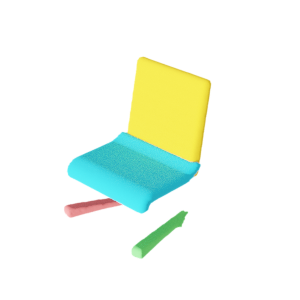} &
        \includegraphics[width=0.15\linewidth]{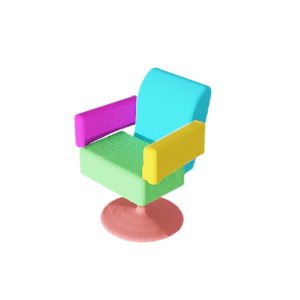} &
        \includegraphics[width=0.15\linewidth]{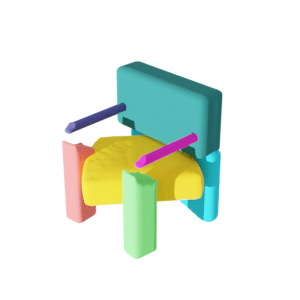} &
        \includegraphics[width=0.15\linewidth]{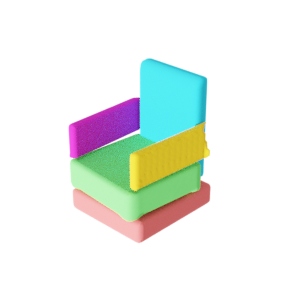}
    \end{tabular}
    \caption{NP vs. Ours on Chair category (2)}
    \label{tab:np_vs_our_chair_2}
\end{figure*}

\begin{figure*}[t!]
    \centering
    \small
    \setlength{\tabcolsep}{1pt}
    \begin{tabular}{rcccccccc}
            \raisebox{2.5em}{Targets} & 
        \includegraphics[width=0.15\linewidth]{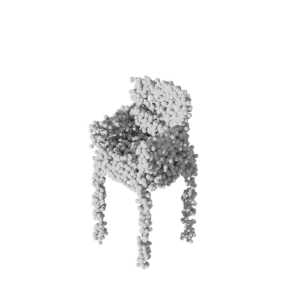} &
        \includegraphics[width=0.15\linewidth]{figs/results_supp/2544targetpc.png} &
        \includegraphics[width=0.15\linewidth]{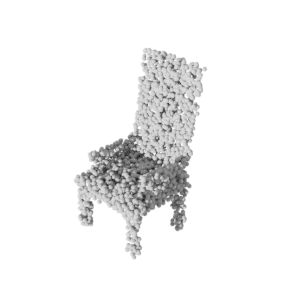} &
        \includegraphics[width=0.15\linewidth]{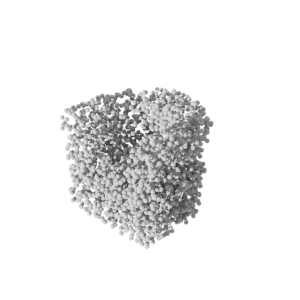} &
        \includegraphics[width=0.15\linewidth]{figs/results_supp/2862targetpc.png} &
        \includegraphics[width=0.15\linewidth]{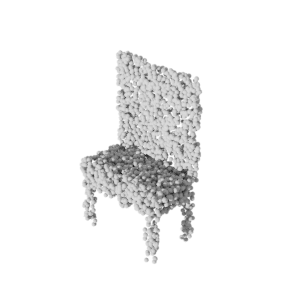} 
        \\
        \raisebox{2.5em}{NP} & 
        \includegraphics[width=0.15\linewidth]{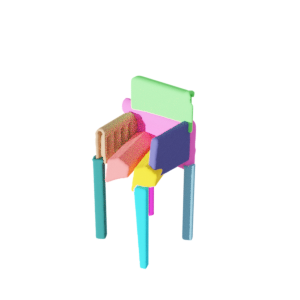} &
        \includegraphics[width=0.15\linewidth]{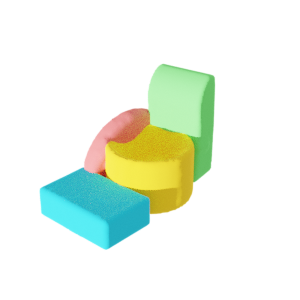} &
        \includegraphics[width=0.15\linewidth]{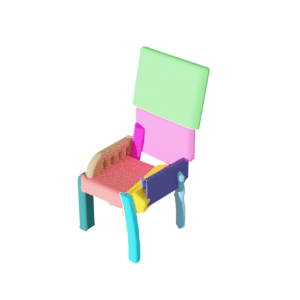} &
        \includegraphics[width=0.15\linewidth]{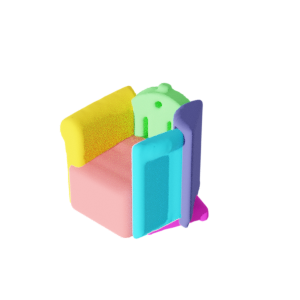} &
        \includegraphics[width=0.15\linewidth]{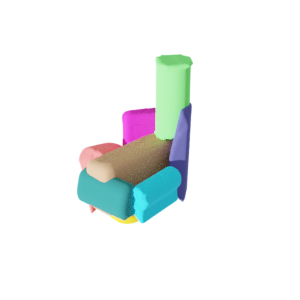} &
        \includegraphics[width=0.15\linewidth]{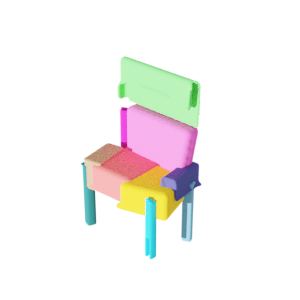} 
        \\
        \raisebox{2.5em}{Ours} & 
        \includegraphics[width=0.15\linewidth]{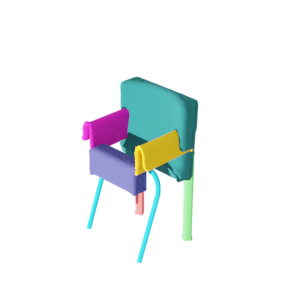} &
        \includegraphics[width=0.15\linewidth]{figs/results_supp/2544recon_ourmesh.png} &
        \includegraphics[width=0.15\linewidth]{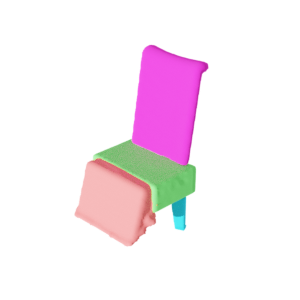} &
        \includegraphics[width=0.15\linewidth]{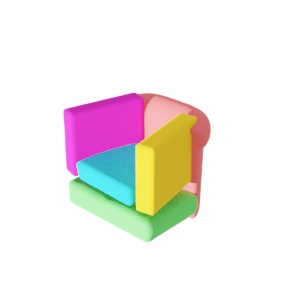} &
        \includegraphics[width=0.15\linewidth]{figs/results_supp/2862recon_ourmesh.png} &
        \includegraphics[width=0.15\linewidth]{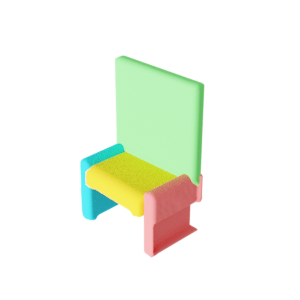}
    \end{tabular}
    \caption{NP vs. Ours on Chair category (3)}
    \label{tab:np_vs_our_chair_3}
\end{figure*}

\begin{figure*}[t!]
    \centering
    \small
    \setlength{\tabcolsep}{1pt}
    \begin{tabular}{rcccccccc}
            \raisebox{2.5em}{Targets} & 
        \includegraphics[width=0.15\linewidth]{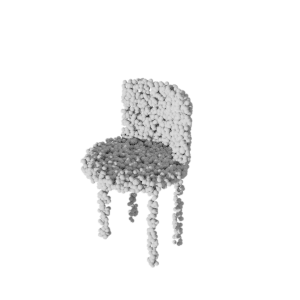} &
        \includegraphics[width=0.15\linewidth]{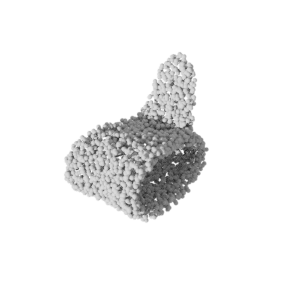} &
        \includegraphics[width=0.15\linewidth]{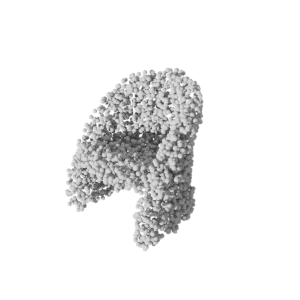} &
        \includegraphics[width=0.15\linewidth]{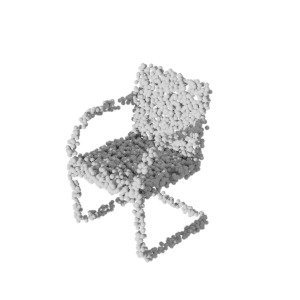} &
        \includegraphics[width=0.15\linewidth]{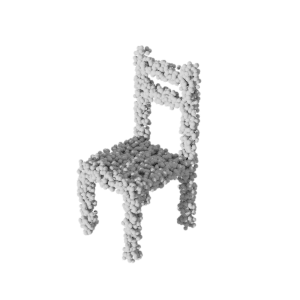} &
        \includegraphics[width=0.15\linewidth]{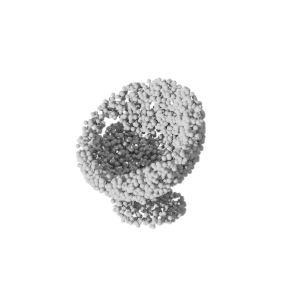}  
        \\
        \raisebox{2.5em}{NP} & 
        \includegraphics[width=0.15\linewidth]{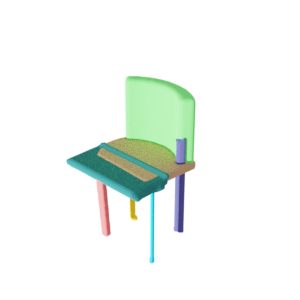} &
        \includegraphics[width=0.15\linewidth]{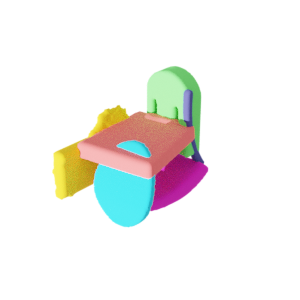} &
        \includegraphics[width=0.15\linewidth]{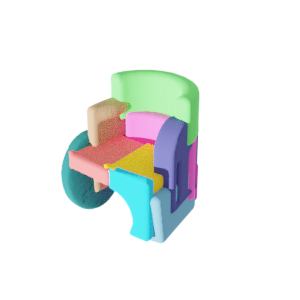} &
        \includegraphics[width=0.15\linewidth]{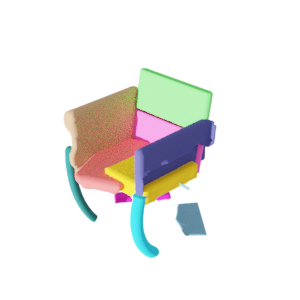} &
        \includegraphics[width=0.15\linewidth]{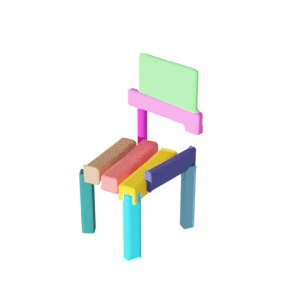} &
        \includegraphics[width=0.15\linewidth]{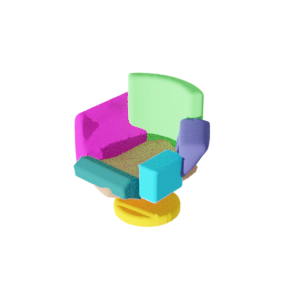} 

        \\
        \raisebox{2.5em}{Ours} & 
        \includegraphics[width=0.15\linewidth]{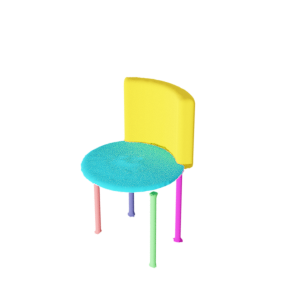} &
        \includegraphics[width=0.15\linewidth]{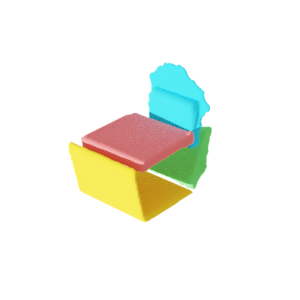} &
        \includegraphics[width=0.15\linewidth]{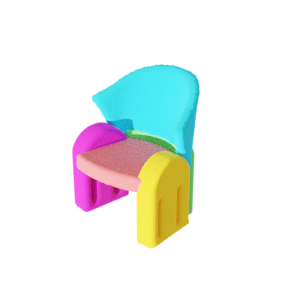} &
        \includegraphics[width=0.15\linewidth]{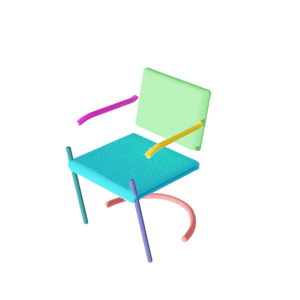} &
        \includegraphics[width=0.15\linewidth]{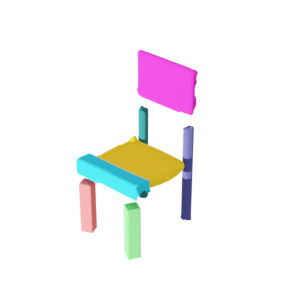} &
        \includegraphics[width=0.15\linewidth]{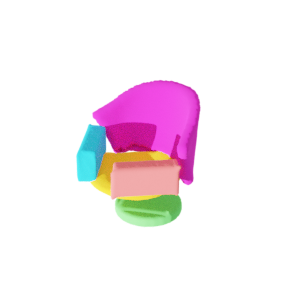}
    \end{tabular}
    \caption{NP vs. Ours on Chair category (4)}
    \label{tab:np_vs_our_chair_4}
\end{figure*}

\begin{figure*}[t!]
    \centering
    \small
    \setlength{\tabcolsep}{1pt}
    \begin{tabular}{rcccccccc}
        \raisebox{2.5em}{Targets} & 
        \includegraphics[width=0.15\linewidth]{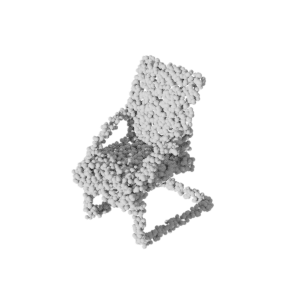} &
        \includegraphics[width=0.15\linewidth]{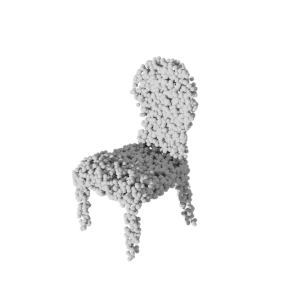} &
        \includegraphics[width=0.15\linewidth]{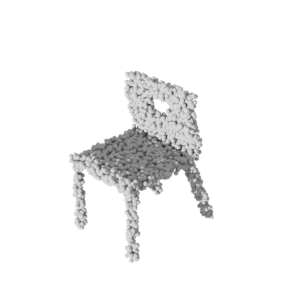} &
        \includegraphics[width=0.15\linewidth]{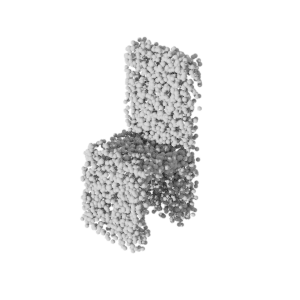} &
        \includegraphics[width=0.15\linewidth]{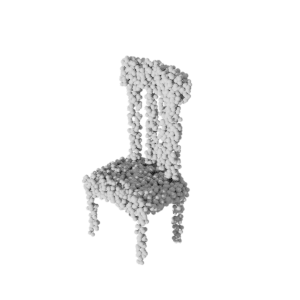} &
        \includegraphics[width=0.15\linewidth]{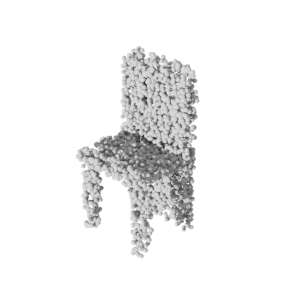} 
        \\
        \raisebox{2.5em}{NP} & 
        \includegraphics[width=0.15\linewidth]{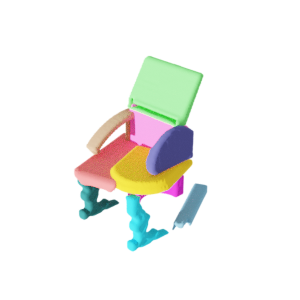} &
        \includegraphics[width=0.15\linewidth]{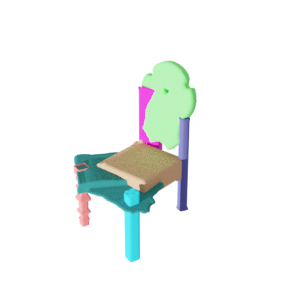} &
        \includegraphics[width=0.15\linewidth]{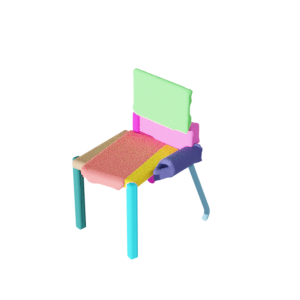} &
        \includegraphics[width=0.15\linewidth]{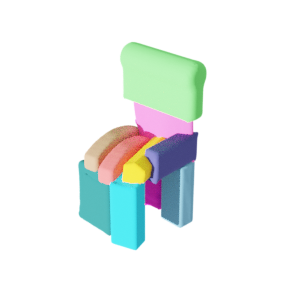} &
        \includegraphics[width=0.15\linewidth]{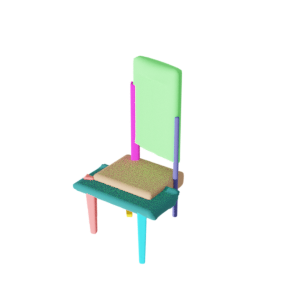} &
        \includegraphics[width=0.15\linewidth]{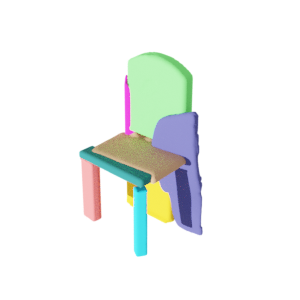} 
        \\
        \raisebox{2.5em}{Ours} & 
        \includegraphics[width=0.15\linewidth]{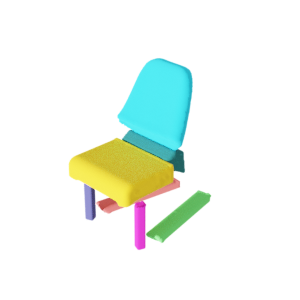} &
        \includegraphics[width=0.15\linewidth]{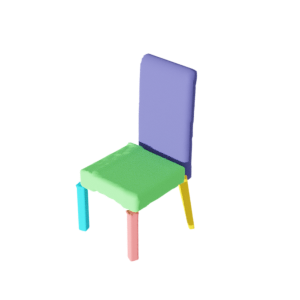} &
        \includegraphics[width=0.15\linewidth]{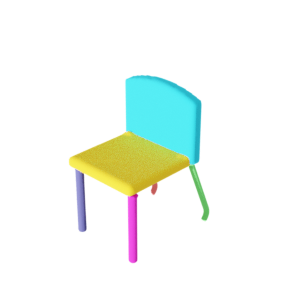} &
        \includegraphics[width=0.15\linewidth]{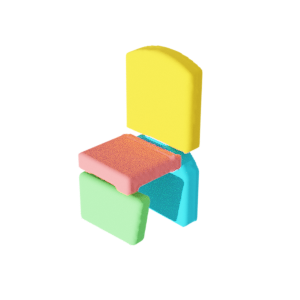} &
        \includegraphics[width=0.15\linewidth]{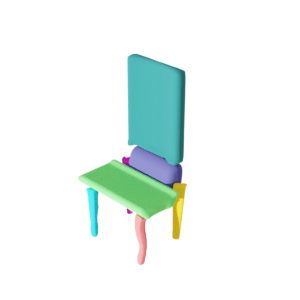} &
        \includegraphics[width=0.15\linewidth]{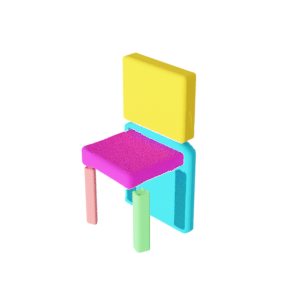}  
    \end{tabular}
    \caption{NP vs. Ours on Chair category (5)}
    \label{tab:np_vs_our_chair_5}
\end{figure*}

\begin{figure*}[t!]
    \centering
    \small
    \setlength{\tabcolsep}{1pt}
    \begin{tabular}{rcccccccc}
        \raisebox{2.5em}{Targets} & 
        \includegraphics[width=0.15\linewidth]{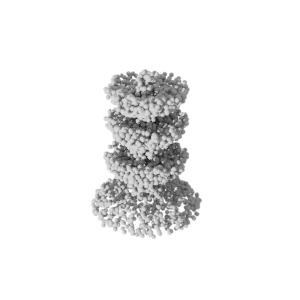} &
        \includegraphics[width=0.15\linewidth]{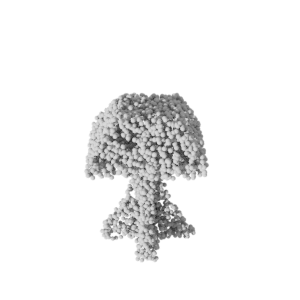} &
        \includegraphics[width=0.15\linewidth]{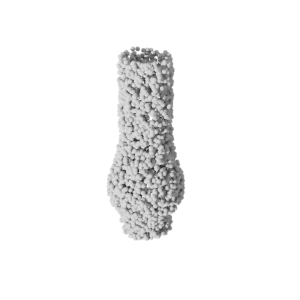} &
        \includegraphics[width=0.15\linewidth]{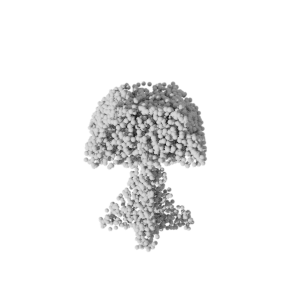} &
        \includegraphics[width=0.15\linewidth]{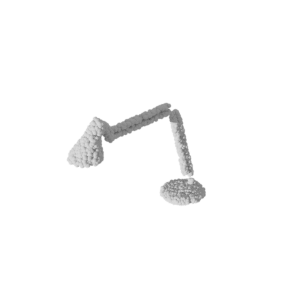} &
        \includegraphics[width=0.15\linewidth]{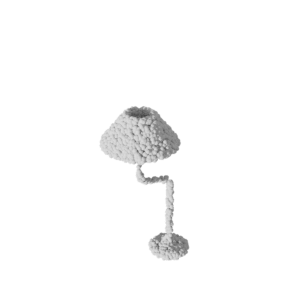} & 
        \\
        \raisebox{2.5em}{NP} & 
        \includegraphics[width=0.15\linewidth]{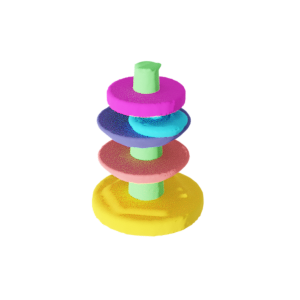} &
        \includegraphics[width=0.15\linewidth]{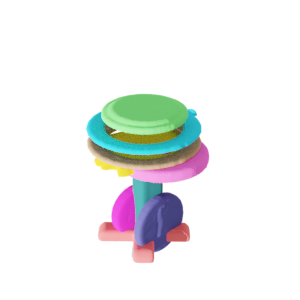} &
        \includegraphics[width=0.15\linewidth]{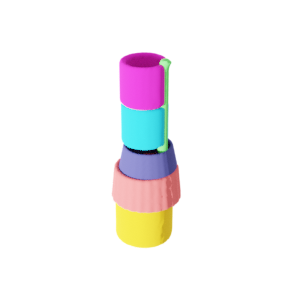} &
        \includegraphics[width=0.15\linewidth]{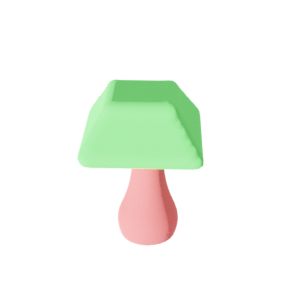} &
        \includegraphics[width=0.15\linewidth]{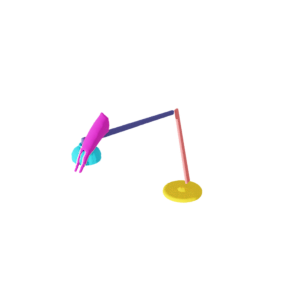} &
        \includegraphics[width=0.15\linewidth]{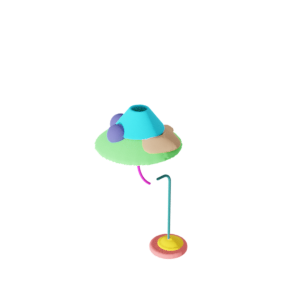} 
        \\
        \raisebox{2.5em}{Ours} & 
        \includegraphics[width=0.15\linewidth]{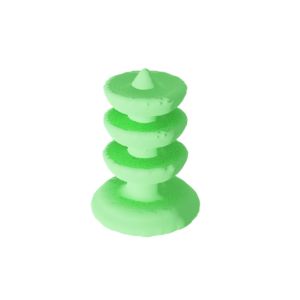} &
        \includegraphics[width=0.15\linewidth]{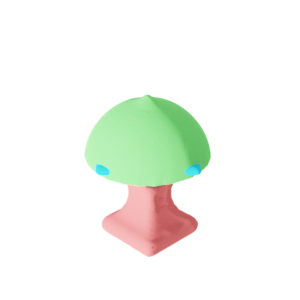} &
        \includegraphics[width=0.15\linewidth]{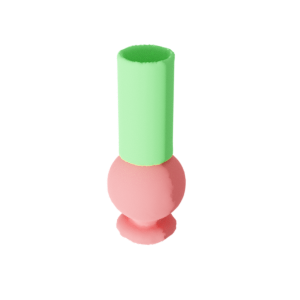} &
        \includegraphics[width=0.15\linewidth]{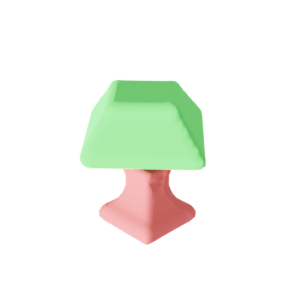} &
        \includegraphics[width=0.15\linewidth]{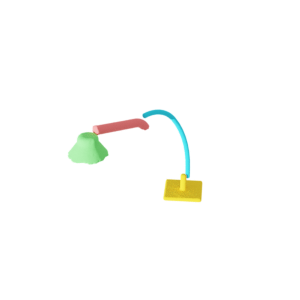} &
        \includegraphics[width=0.15\linewidth]{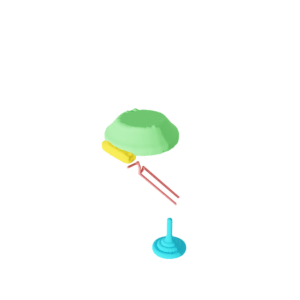}
    \end{tabular}
    \caption{NP vs. Ours on Lamp category (1)}
    \label{tab:np_vs_our_lamp_1}
\end{figure*}

\begin{figure*}[t!]
    \centering
    \small
    \setlength{\tabcolsep}{1pt}
    \begin{tabular}{rcccccccc}
            \raisebox{2.5em}{Targets} & 
        \includegraphics[width=0.15\linewidth]{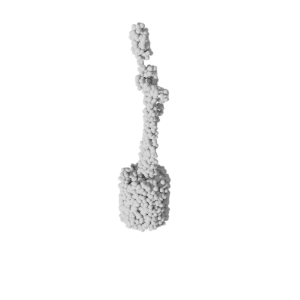} &
        \includegraphics[width=0.15\linewidth]{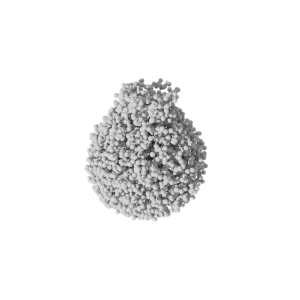} &
        \includegraphics[width=0.15\linewidth]{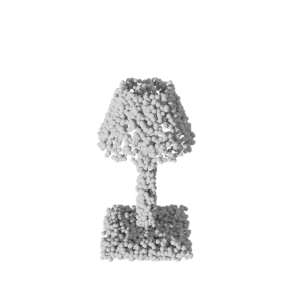} &
        \includegraphics[width=0.15\linewidth]{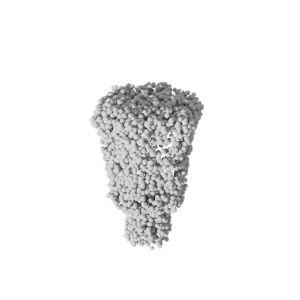} &
        \includegraphics[width=0.15\linewidth]{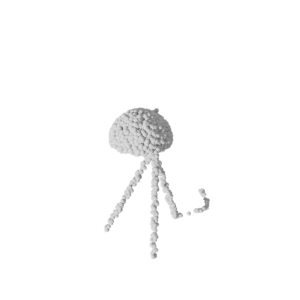} &
        \includegraphics[width=0.15\linewidth]{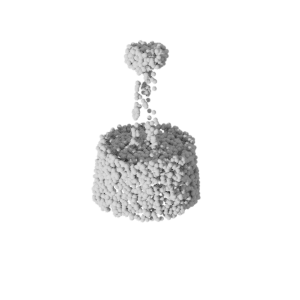} 
        \\
        \raisebox{2.5em}{NP} & 
        \includegraphics[width=0.15\linewidth]{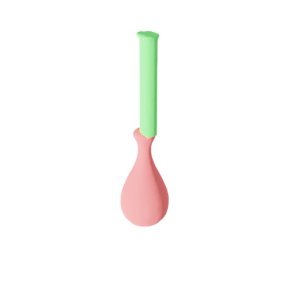} &
        \includegraphics[width=0.15\linewidth]{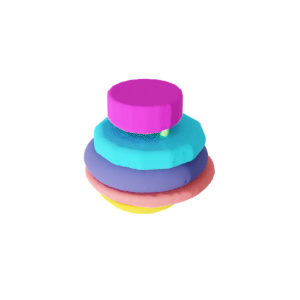} &
        \includegraphics[width=0.15\linewidth]{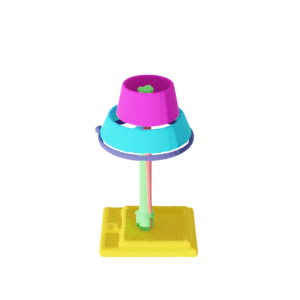} &
        \includegraphics[width=0.15\linewidth]{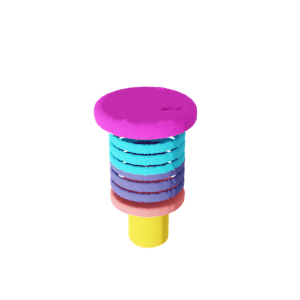} &
        \includegraphics[width=0.15\linewidth]{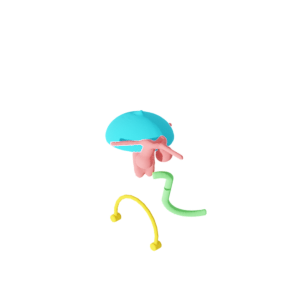} &
        \includegraphics[width=0.15\linewidth]{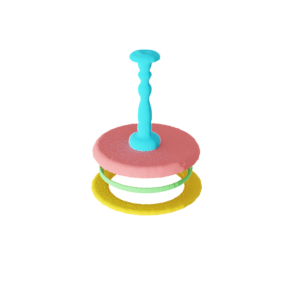} 
        \\
        \raisebox{2.5em}{Ours} & 
        \includegraphics[width=0.15\linewidth]{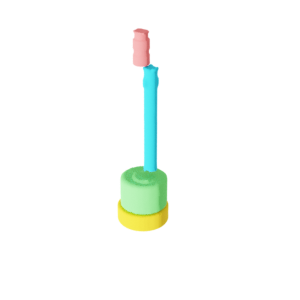} &
        \includegraphics[width=0.15\linewidth]{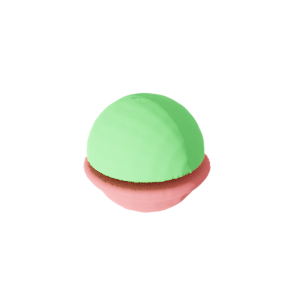} &
        \includegraphics[width=0.15\linewidth]{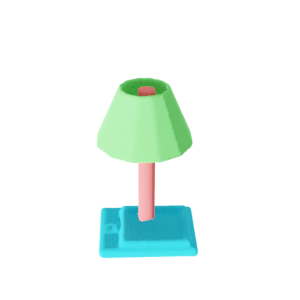} &
        \includegraphics[width=0.15\linewidth]{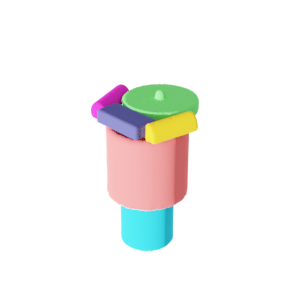} &
        \includegraphics[width=0.15\linewidth]{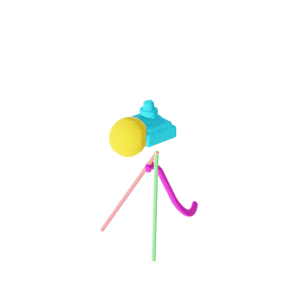} &
        \includegraphics[width=0.15\linewidth]{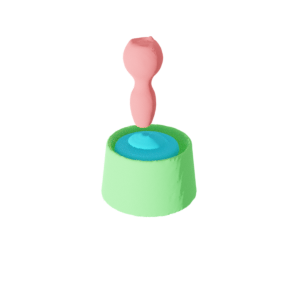}
    \end{tabular}
    \caption{NP vs. Ours on Lamp category (2)}
    \label{tab:np_vs_our_lamp_2}
\end{figure*}

\begin{figure*}[t!]
    \centering
    \small
    \setlength{\tabcolsep}{1pt}
    \begin{tabular}{rcccccccc}
            \raisebox{2.5em}{Targets} & 
        \includegraphics[width=0.15\linewidth]{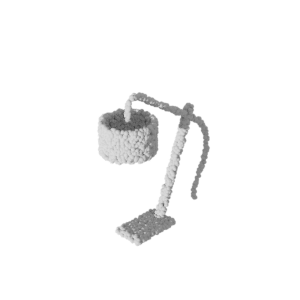} &
        \includegraphics[width=0.15\linewidth]{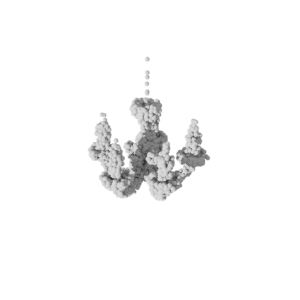} &
        \includegraphics[width=0.15\linewidth]{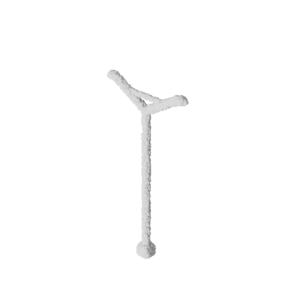} &
        \includegraphics[width=0.15\linewidth]{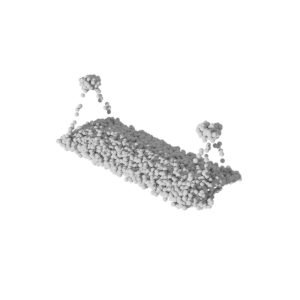} &
        \includegraphics[width=0.15\linewidth]{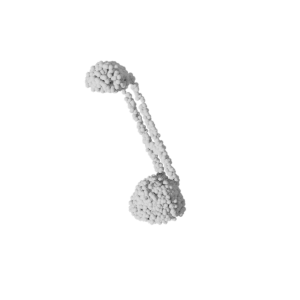} &
        \includegraphics[width=0.15\linewidth]{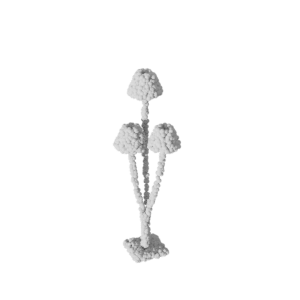} 
        \\
        \raisebox{2.5em}{NP} & 
        \includegraphics[width=0.15\linewidth]{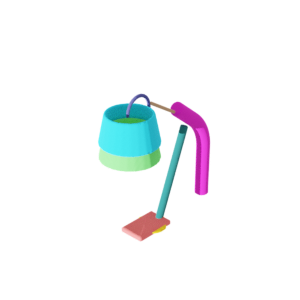} &
        \includegraphics[width=0.15\linewidth]{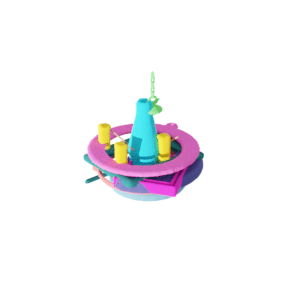} &
        \includegraphics[width=0.15\linewidth]{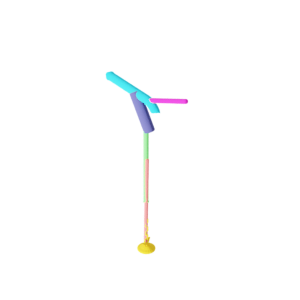} &
        \includegraphics[width=0.15\linewidth]{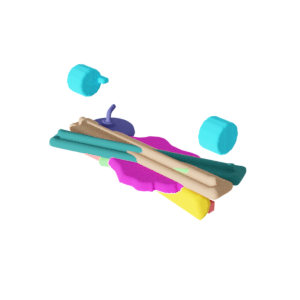} &
        \includegraphics[width=0.15\linewidth]{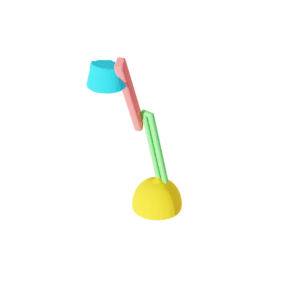} &
        \includegraphics[width=0.15\linewidth]{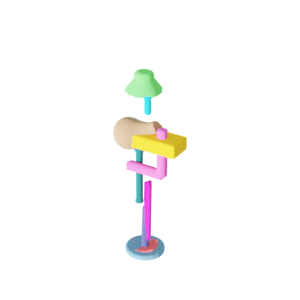} 
        \\
        \raisebox{2.5em}{Ours} & 
        \includegraphics[width=0.15\linewidth]{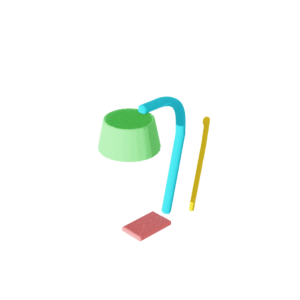} &
        \includegraphics[width=0.15\linewidth]{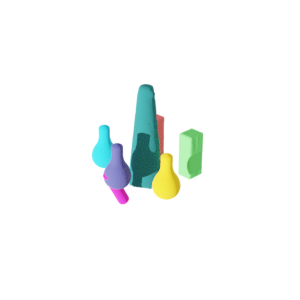} &
        \includegraphics[width=0.15\linewidth]{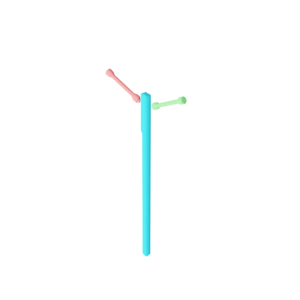} &
        \includegraphics[width=0.15\linewidth]{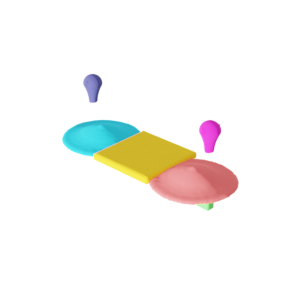} &
        \includegraphics[width=0.15\linewidth]{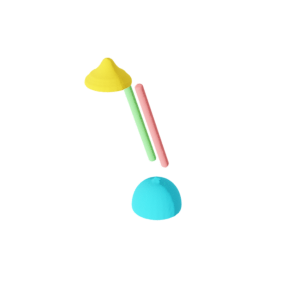} &
        \includegraphics[width=0.15\linewidth]{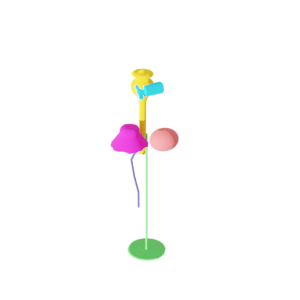}
    \end{tabular}
    \caption{NP vs. Ours on Lamp category (3)}
    \label{tab:np_vs_our_lamp_3}
\end{figure*}

\begin{figure*}[t!]
    \centering
    \small
    \setlength{\tabcolsep}{1pt}
    \begin{tabular}{rcccccccc}
            \raisebox{2.5em}{Targets} & 
        \includegraphics[width=0.15\linewidth]{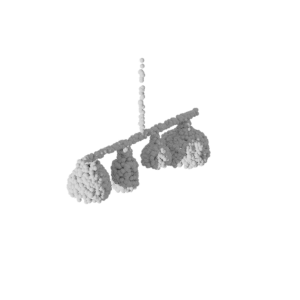} &
        \includegraphics[width=0.15\linewidth]{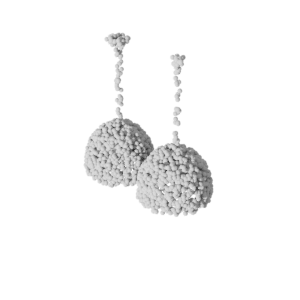} &
        \includegraphics[width=0.15\linewidth]{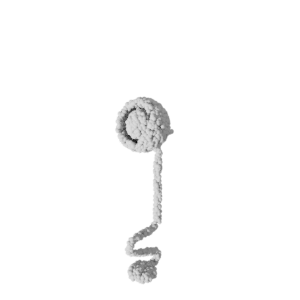} &
        \includegraphics[width=0.15\linewidth]{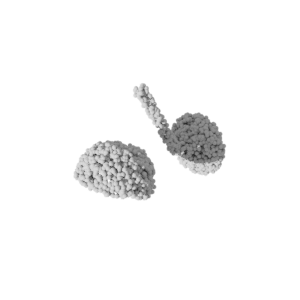} &
        \includegraphics[width=0.15\linewidth]{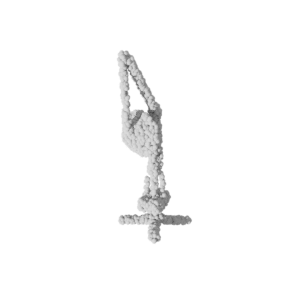} &
        \includegraphics[width=0.15\linewidth]{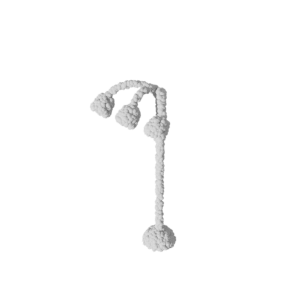} 
        \\
        \raisebox{2.5em}{NP} & 
        \includegraphics[width=0.15\linewidth]{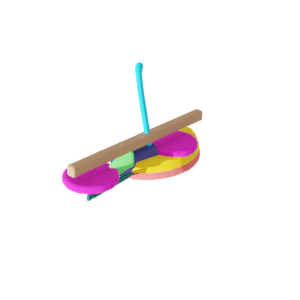} &
        \includegraphics[width=0.15\linewidth]{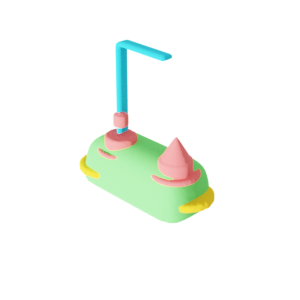} &
        \includegraphics[width=0.15\linewidth]{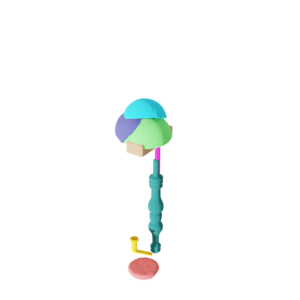} &
        \includegraphics[width=0.15\linewidth]{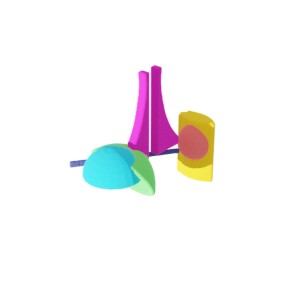} &
        \includegraphics[width=0.15\linewidth]{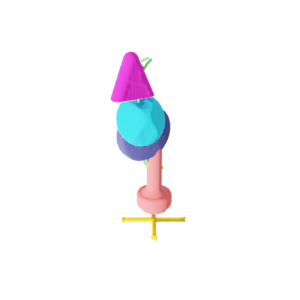} &
        \includegraphics[width=0.15\linewidth]{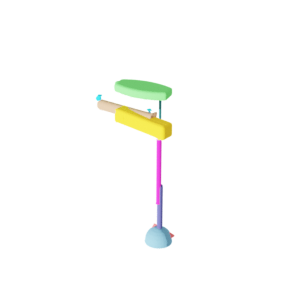} 

        \\
        \raisebox{2.5em}{Ours} & 
        \includegraphics[width=0.15\linewidth]{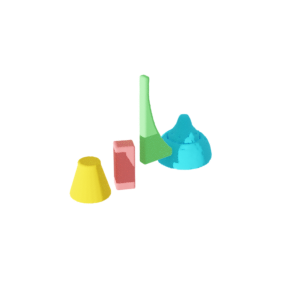} &
        \includegraphics[width=0.15\linewidth]{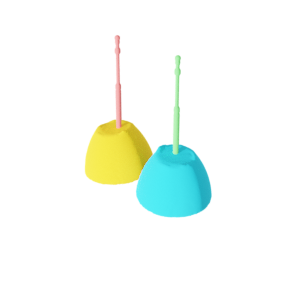} &
        \includegraphics[width=0.15\linewidth]{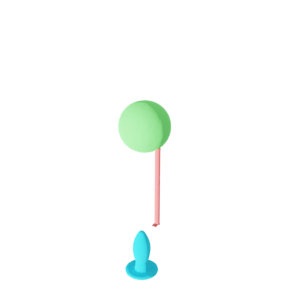} &
        \includegraphics[width=0.15\linewidth]{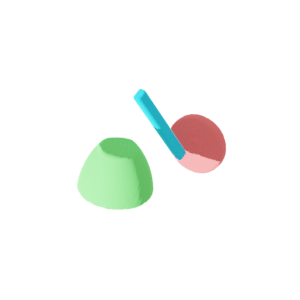} &
        \includegraphics[width=0.15\linewidth]{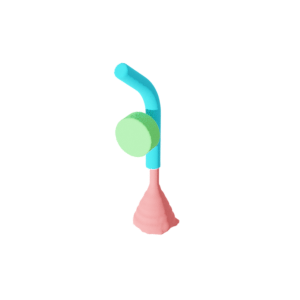} &
        \includegraphics[width=0.15\linewidth]{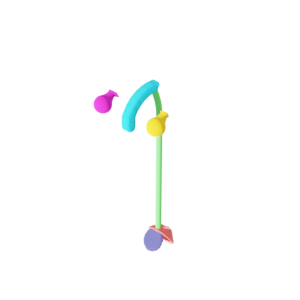}
    \end{tabular}
    \caption{NP vs. Ours on Lamp category (4)}
    \label{tab:np_vs_our_lamp_4}
\end{figure*}

%{\small
%\bibliographystyle{ieee_fullname}
%\bibliography{egbib}
%}